\documentclass[thesis]{thesis-umich}
 \pdfoutput=1

\usepackage{blindtext} 
\usepackage[export]{adjustbox}
\usepackage{tabularx}
\usepackage{float}
\usepackage{wrapfig}
\usepackage{lscape}

\title{Non-Parametric Neural Style Transfer}

\author{Nicholas Isaac Kolkin}

\department{Computer Science}

\email{nick.kolkin@ttic.edu}

\orcid{0000-0003-1233-1969}

\frontispiece{\includegraphics[width=4in]{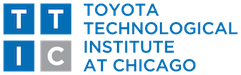}}

\frontpagestyle{7} 

\acknowledgments[7]{Thank you to my mentors. First and foremost thank you Greg Shakhnarovich, your patience and wisdom has guided me through this process, and made it the most educational period of my life. Thank you Jason Salavon, Eli Shechtman, Sylvain Paris, and Ayan Chakrabarti for teaching me about the many facets of image synthesis. Thank you Matt Kusner, Steven Tyree, Jake Gardner and Kilian Weinberger for introducing me to machine learning, and helping me discover my love of it early on. 

\vspace{0.5cm}
Thank you Cooper Nederhood, Sunnie Kim, Igor Vasiljevic, and Reza Mostajabi; working closely with you in the past few years have been some of the highlights grad school. Thank you Gabriel Hope for being my first research buddy, I will never forget our nights hacking at WU-SVM. Thank you to the rest of the PALS group for the many fun and interesting discussions about computer vision. Thank you Adam Bohlander for all of the help you've given me personally and generally everything you've done to make computation at TTI painless. Thank you to the rest of the students, faculty, and staff at TTI for creating a fun, welcoming, and nerdy place to learn and do research. 

\vspace{0.5cm}
Thank you Salonica breakfast club, for the laughs, hot takes, super smash bros, running, and everything else. 

\vspace{0.5cm}
Thank you Emma, you made 2020 the best year of grad school, which is quite a feat.

\vspace{0.5cm}
Thank you Mom, Dad, Zach, Grandparents, and the rest of my family. Words can't express how grateful I am for your love and support. }

\committee{ %
Professor Gregory Shakhnarovich, Chair \\
Professor Matthew Turk \\
Professor Michael Maire}

\chair{Professor First Name}

\showlistoftables

\abstract{
It seems easy to imagine a photograph of the Eiffel Tower painted in the style of Vincent van Gogh's 'The Starry Night', but upon introspection it is difficult to precisely define what this would entail. What visual elements must an image contain to represent the 'content' of the Eiffel Tower? What visual elements of 'The Starry Night' are caused by van Gogh's 'style' rather than his decision to depict a village under the night sky? Precisely defining 'content' and 'style' is a central challenge of designing algorithms for artistic style transfer, algorithms which can recreate photographs using an artwork's style. My efforts defining these terms, and designing style transfer algorithms themselves, are the focus of this thesis. I will begin by proposing novel definitions of style and content based on optimal transport and self-similarity, and demonstrating how a style transfer algorithm based on these definitions generates outputs with improved visual quality. Then I will describe how the traditional texture-based definition of style can be expanded to include elements of geometry and proportion by jointly optimizing a keypoint-guided deformation field alongside the stylized output's pixels. Finally I will describe a framework inspired by both modern neural style transfer algorithms and traditional patch-based synthesis approaches which is fast, general, and offers state-of-the-art visual quality.

}

\DeclareMathOperator*{\argmin}{argmin}

\begin{document}

\chapter{Introduction}
\begin{figure}[H]
    \centering
    \includegraphics[width=\linewidth]{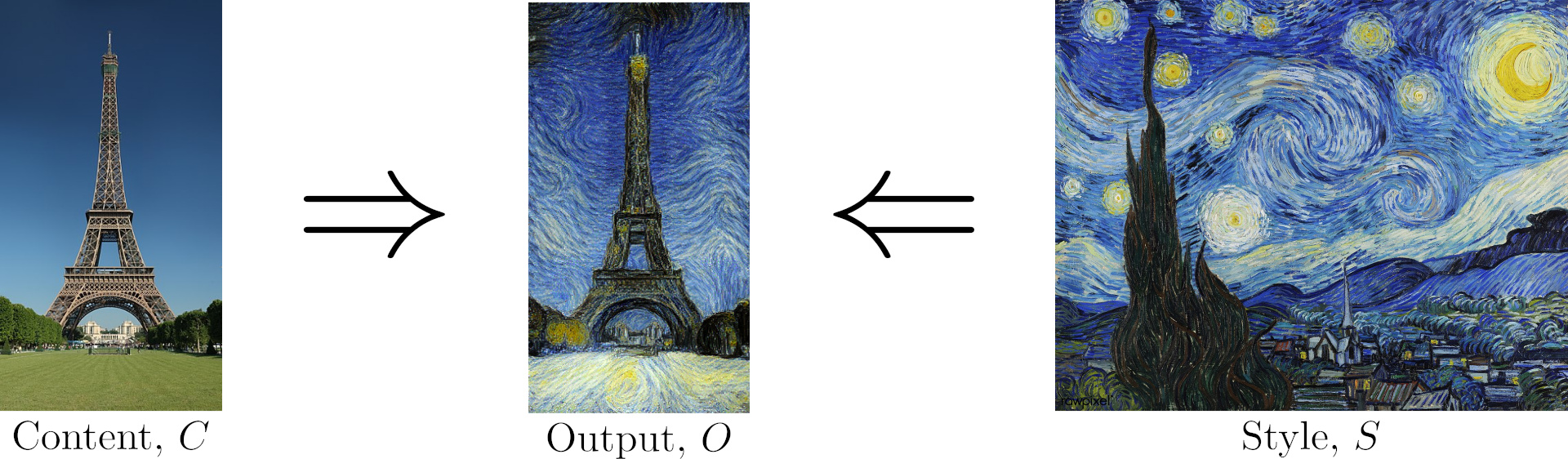}
    \caption{Style transfer algorithms combine the content of image $C$ with the style of artwork $S$ to create the novel output image $O$.}
    \label{fig:fig1}
\end{figure}
\label{chpt:introduction}
Naively it seems easy to imagine a photograph of the Eiffel Tower painted in the style of Vincent van Gogh's 'The Starry Night', but upon introspection it is difficult to precisely define what this entails. What visual elements must an image maintain to still represent the 'content' of the Eiffel Tower? What visual elements of 'The Starry Night' are caused by van Gogh's 'style' rather than his decision to depict a village under the night sky? How can we represent style and content, extract these representations from images, then mix and match the results to synthesize novel stylized images? These are the fundamental questions of Artistic Style Transfer and this thesis will explore my progress answering them over the past several years.

\section{What is Style Transfer?}
Style transfer algorithms receive two inputs, a content image $C$ and a style image $S$, then produce an output image $O$. This is a form of data-driven non-photorealistic rendering where $C$ defines the scene to be rendered and $S$ defines the rendering procedure. Our immediate motivation is to improve style transfer's utility as an artistic tool. Better style transfer algorithms can endow artists with the ability to automate tedious tasks or create visual effects which are difficult to replicate manually. For example:
\begin{itemize}
    \item The novice who wishes to represent the contents of a personal photograph in the style of a favorite artist.
    \item The artist who wishes to propagate their work carefully illustrating one frame of an animation to rough sketches of other frames.
    \item The graphics designer who wishes to stylize an object using a photo-realistic texture. A task which is difficult to achieve manually, even for experts. 
\end{itemize}

However, style transfer has value beyond this. Representational artwork (i.e. artwork depicting scenes with recognizable semantics) offers a unique window into the human visual system. It is a byproduct of our internal representations of content which artists craft to be both recognizable and visually stimulating. Artwork frequently presents visual stimuli which we are unlikely to encounter in the natural world, and which test the boundaries of our perceptual system. Style transfer can be used to leverage these exciting properties of art in two ways. One is that improving style transfer motivates developing an explicit understanding of 'content' and it's invariances. In the future these insights could be used to develop better automatic recognition algorithms. The second, more direct, path is using style transfer as a black box to augment training data, producing many stylizations of labeled content images to aid generalization \cite{zheng2019stada, jackson2019style}. Both paths are served by developing algorithms that can more closely match the abilities of human artists. 

Progress replicating these capabilities has been uneven, and before discussing my own work specifically, it will be useful to describe what 'style' and 'content' mean within the style transfer community, and how this differs from the definitions used by artists.

We can begin with an operational definition. In other words, what does it mean for a style transfer algorithm today to 'succeed'?  

\begin{figure}
    \centering
    \includegraphics[width=\linewidth]{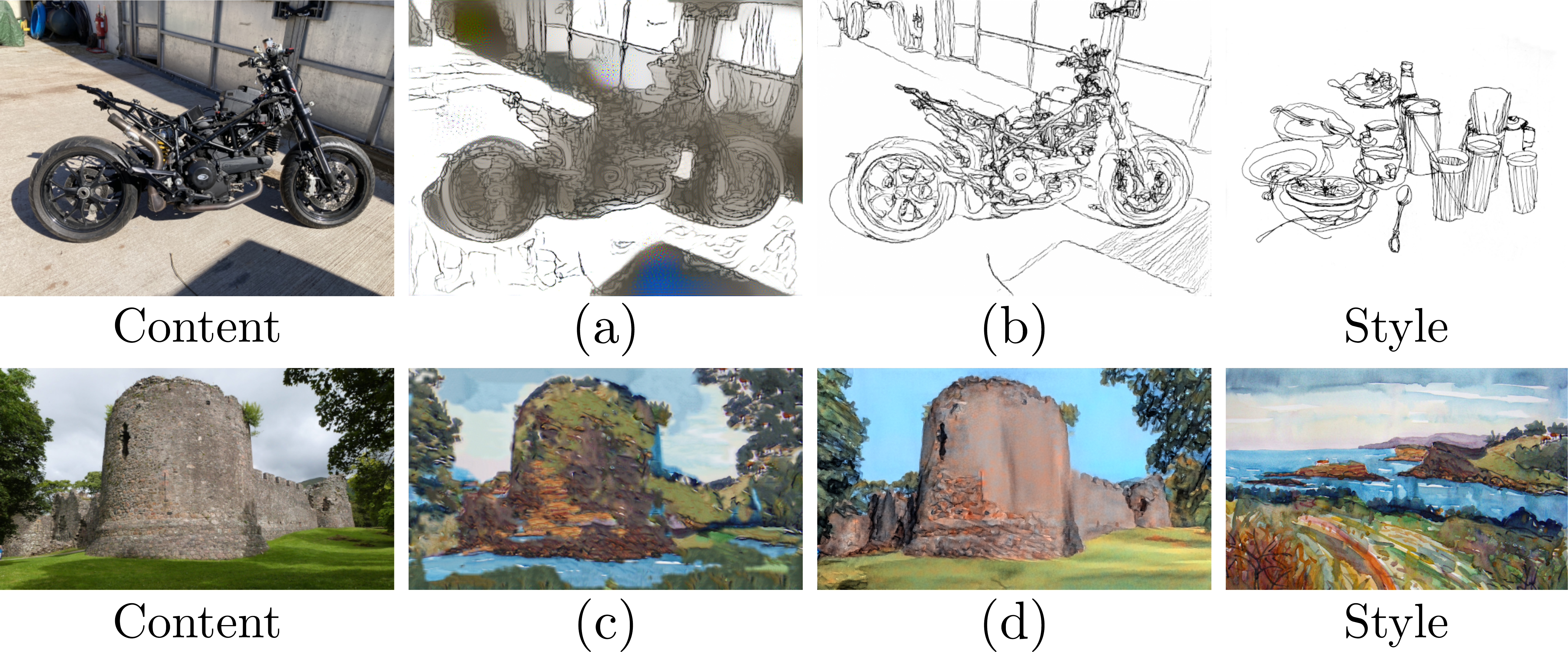}
    \caption{Examples of style transfer successes and failures. In (a) the target style is poorly captured by the output because it is not composed of the same sparse pen lines as the style. While the lines that compose (b) are not as sharply defined as those in the style, but it is still a much closer match than (a). Image (c) captures the patch distribution of the style well, but does not maintain the salient edges and self-similarity patterns of the original content (regions with the same texture in the content take on multiple textures in the output), making it hard to perceive the same semantics in the resulting output. Image (d) preserves content well because it shares the same edges and self-similarity patterns as the content. It approximates the patch distribution of the style reasonably, but fails to capture the style's discrete brush strokes. By today's standards (b) and (d) are fairly successful examples style transfer, despite their flaws.}
    \label{fig:succ_fail}
\end{figure}

An algorithm 'succeeds' in matching a target style when the distribution of patches in $O$ and $S$ is similar. In other words if I zoom in on a small section of $O$, it is difficult to tell that I haven't actually zoomed in on a small section of $S$. If this patch is a single pixel, then succeeding in this means matching the color distribution of the style. If this patch is the entire image, then succeeding in this would mean ultimate success in style transfer, where I believe that the stylized output and original style are sections of the same larger piece of art. 

When $O$ 'successfully' maintains content it should preserve perceptually important edges from $C$ (i.e. humans asked to label the boundaries between perceptually distinct regions, as in the labeling of the BSDS dataset \cite{MartinFTM01}, would label roughly the same contours in the content and the stylized output) and regions with similar textures in $C$ should take on similar textures in $O$ (although this last quality isn't always desirable, for example when $S$ is a cubist style). While the explicit definitions of style and content used by various style transfer algorithms can dramatically differ from this interpretation, in my experience outputs which humans perceive as `good', `compelling', or `successful' tend to share the above characteristics.

Of course, despite some attempt at formalism, I have based these definitions on several poorly defined terms. Parameterizing a distribution of patches, measuring the distance between patches, measuring the distance between distributions of patches, and defining and extracting perceptually important edges or homogeneous regions of texture are all open problems. Therefore I think it will also useful to give some additional qualitative criteria I use to judge style transfer outputs. 

The least stringent (although still commonly failed) threshold is whether $O$ contains visually jarring artifacts that are not present in the style or content. A more ambitious threshold is whether $O$ appears to have been created with the same artistic tools as $S$. For example, if $S$ is a pencil drawing, does $O$ appear to be drawn by a physical pencil, or does it contain mistakes that give it away as being generated by a computer? The most lofty threshold is whether $O$ can be imagined to be a continuation of $S$. In other words can we imagine $O$ and $S$ to be two sections of a single larger and unseen piece of art? This last threshold is rarely crossed, because the definition of 'style' used by the style transfer community only encapsulates very limited subset of the definition used by artists.

An artist asked to define style would likely say that it consists of both global attributes such as composition and choice of subject matter; and local attributes such as technique and choice of media. Almost all style transfer algorithms today focus on the latter, local, aspects of style. While the best outputs may reproduce the textures of the target style while keeping the content perfectly recognizable, they will not alter the perspective the content is rendered from, or change the layout and composition of the scene to better match the artist's style. Only a handful of methods alter the proportions and geometry of objects (one of which is a part of this thesis), and efforts in this direction are still nascent. This means that some styles are better suited to today's algorithms than others. The distinctive brushstrokes of many Impressionist and Post-Impressionist paintings can be modelled fairly well, while the careful composition of Renaissance paintings cannot. 

As a result most style transfer today can be seen as a generalization of texture transfer. Unlike traditional texture transfer, where the target texture is fairly homogeneous, a target style might contain many different textures. However, in Chapter \ref{chpt:dst} we will take a step beyond this paradigm, and propose a method by which geometry and proportion can be stylized as well.
\newpage
\begin{figure}[H]
    \centering
    \includegraphics[width=\linewidth]{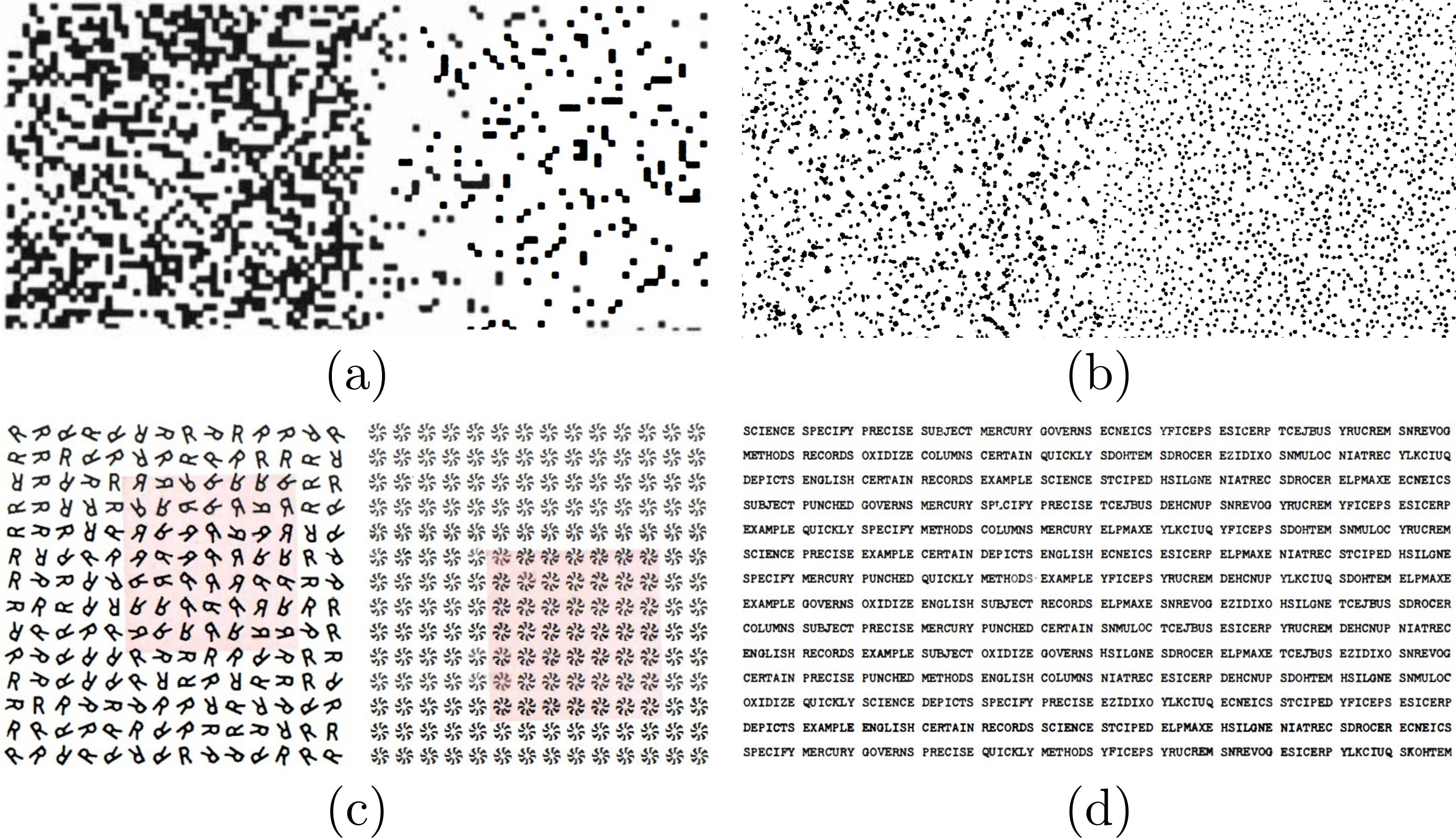}
    \caption{Examples of stimuli used by Julesz to study pre-attentive texture discrimination. Image (a) displays two fields that differ in their first order statistics (average brightness), and can be easily be disambiguated by the pre-attentive visual system. Image (b) displays two fields that differ in second order statistics (minimum spacing between dots), like (a) these are easily disambiguated by the pre-attentive visual system. Images (c) and (d) display texture fields that only differ in higher order statistics and \textbf{cannot} be disambiguated without careful inspection. Image (c) showcases that altering a texture by mirror-flipping is not easily detected. Image (d) showcases that without careful inspection it is impossible to discriminate between jumbles of letters and proper words. Images (a) and (d) were taken from \cite{julesz1965texture}, and images (b) and (c) were taken from \cite{julesz1975experiments} (the contrast of (b) was increased for clarity).}
    \label{fig:julesz_intro}
\end{figure}

\section{A Brief History of Style Transfer}
\subsection{Texture Synthesis}

Computer scientists' definition of texture and style can be traced to the research of Bela Julesz on texture discrimination in humans \cite{julesz1965texture, julesz1975experiments, julesz1981theory}. Julesz used a computer to generate a grid of 'micropatterns' (i.e. image patches) using a random Markov process (i.e. micropatterns were either sampled independently or depended only on their immediately neighbors). Different distributions of micropatterns were used for different sections of the grid, for example the right and left halves of the image might use different distributions of micropatterns, or the central portion of the image might use a different distribution than the boundary. Julesz found that human perception was extremely sensitive to the first and second order statistics of micropatterns (e.g. average brightness, variance of dot deviation from uniform grid, rotations other than 180$^\circ$), but careful scrutiny was required to differentiate between textures of micropatterns that differed only in higher order statistics (e.g. mirror images, words vs random letter sequences, one vs two small white holes in black circles). Later work by Julesz discovered that some higher level statistics (whether small dots were inside or outside a hollow black ring) were easily detected, however the core idea remained that texture was a local phenomena which could be captured by summary statistics of micropatterns (i.e. image patches). Quoting \cite{efros2001image}, this lead to a high-level framework for texture synthesis: "(1) picking the right set of statistics to match, (2) finding an algorithm that matches them."

\begin{wrapfigure}{r}{0.48\linewidth}
    \centering
    \vspace{-0.5cm}
    \includegraphics[width=\linewidth]{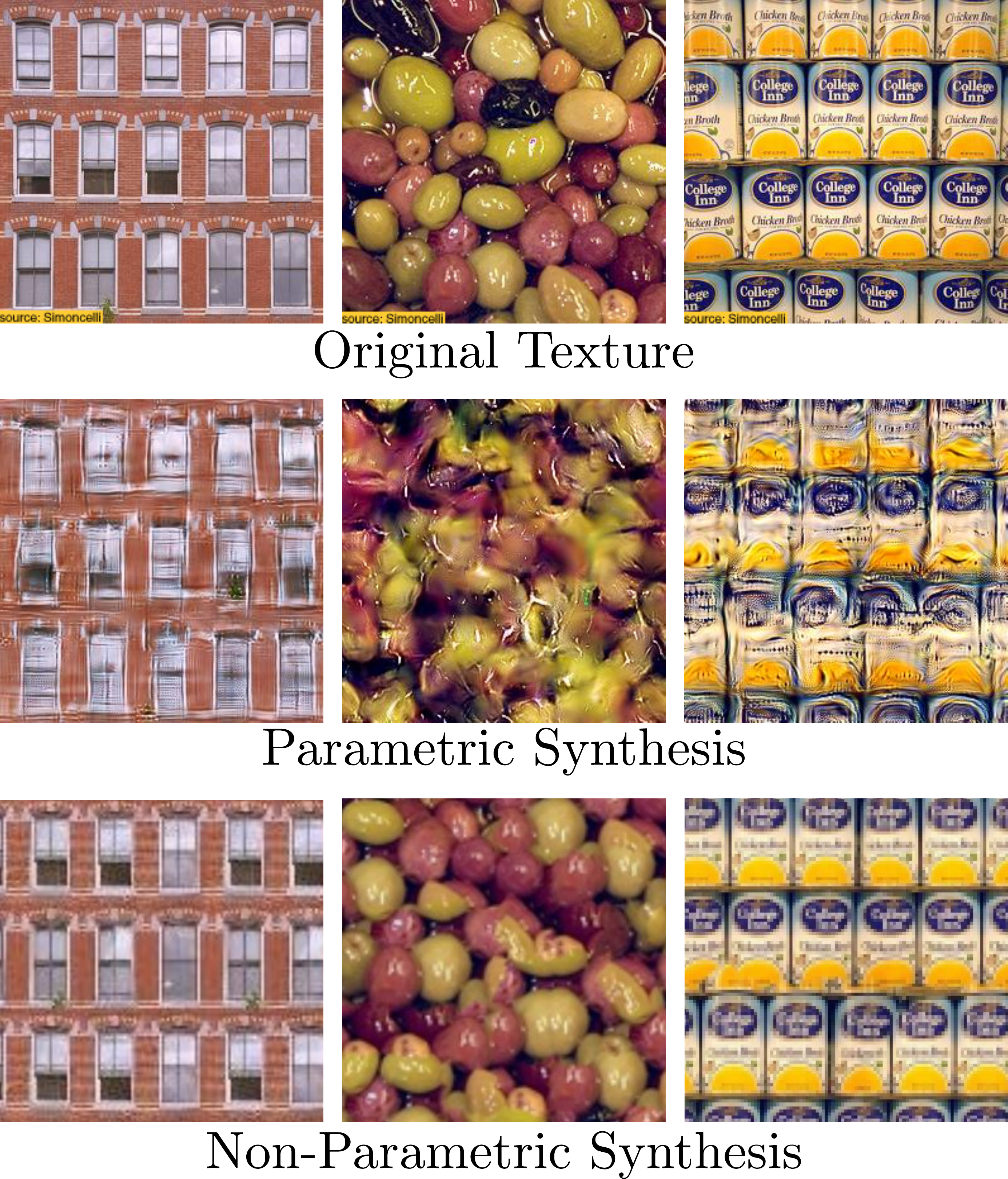}
    \caption{Examples of texture exemplars, outputs produced by the parametric approach of \cite{Portilla00}, and outputs produces by the non-parametric patch synthesis approach of \cite{efros2001image}. The lower resolution of the non-parametric results is not an inherent limitation of the method.}
    \vspace{-1cm}
    \label{fig:dst_face}
\end{wrapfigure}

In texture synthesis an algorithm receives a small image containing the desired texture, then should be able to synthesize new images perceived to consist of the same texture, but with arbitrary height and width. Some work in texture synthesis hewed very closely to the original findings of Julesz, applying hand-crafted feature extractors to image patches, then generating new images that matched summary statistics derived from these features \cite{Zhu98frame, Portilla00}. Other works, noting that textures in the real world can exhibit far more complex structure than the artificial examples used by Julesz, synthesized textures by stitching together image patches or pixels sampled from example texture \cite{wei2000fast,Ashikhmin01,Liang01, efros2001image}.

\subsection{Early Style Transfer}
\begin{figure}[H]
    \centering
    \includegraphics[width=\linewidth]{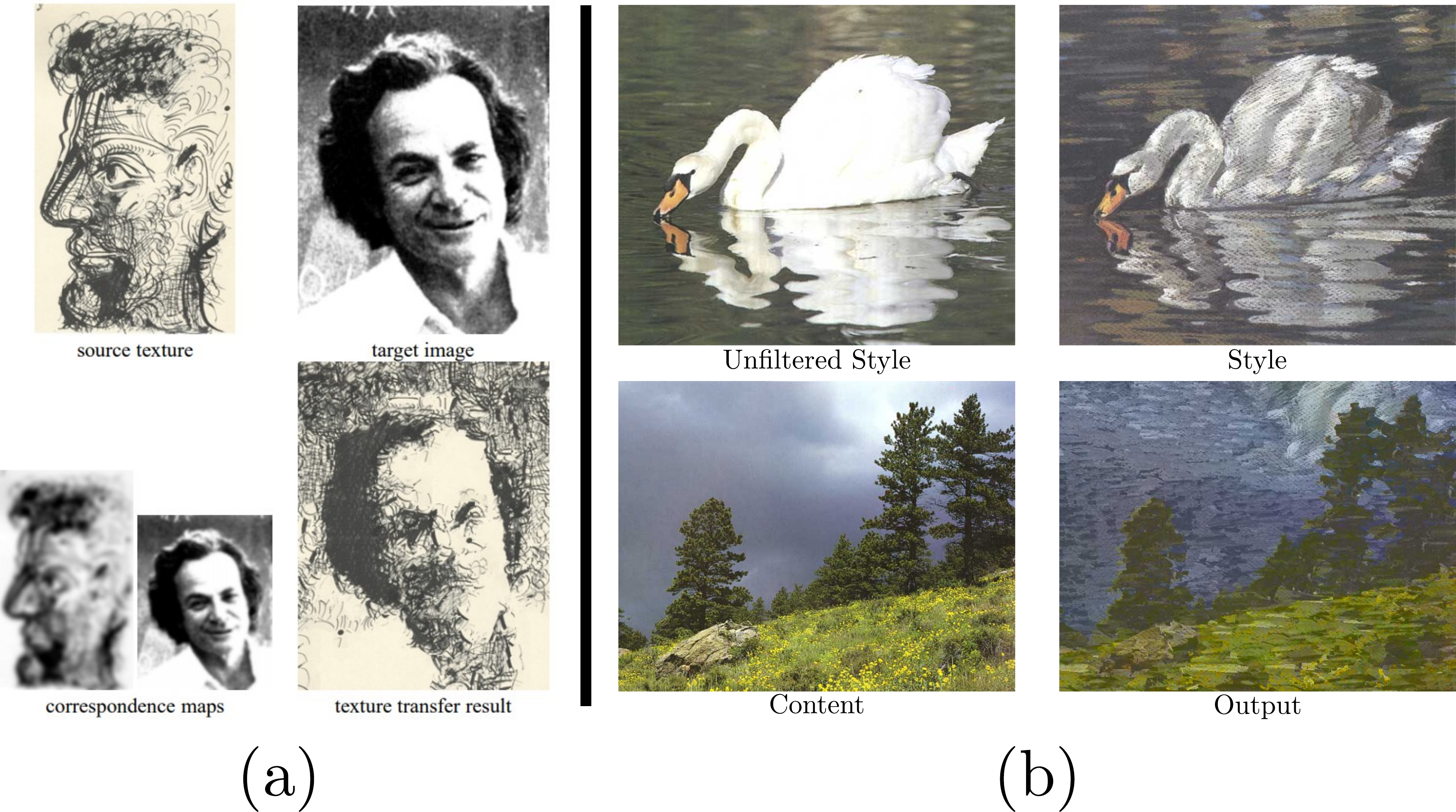}
    \caption{Examples of early style transfer algorithms, which were all non-parametric. These differed from earlier texture synthesis algorithms mainly in that it was no longer assumed that the exemplar contained a single homogeneous texture. For this reason additional supervision/guidance was incorporated into the texture synthesis framework. (a) displays the output of the algorithm proposed by Efros and Freeman \cite{efros2001image}, which used a blurred luminance map as a source of guidance. (b) displays the output of the algorithm proposed by Hertzmann et al. \cite{hertzmann2001image}, which required an 'unfiltered' (i.e. 'unstylized') version of the style image to supervise the mapping between the photo-realistic and artistic domains.}
    \label{fig:og_st_intro}
\end{figure}
Early efforts in style transfer first emerged from the patch synthesis school, appearing simultaneously in \cite{hertzmann2001image} as a novel task and in \cite{efros2001image} as an extension of texture synthesis. In \cite{efros2001image} patches from different regions of style are guided to regions of the content with similar luminosity. In \cite{hertzmann2001image} guidance is based on an 'unfiltered' version of the style which can be more easily matched to the photographic content image (e.g. a photograph of some apples on a table is the unfiltered style, and a watercolor painting of the same apples from the same viewpoint is the filtered style). Extending texture synthesis to heterogeneous style images using additional guidance is a powerful technique, and this approach is still influential today where it has led to state of the art algorithms in data-driven 3d rendering \cite{fivser2016stylit, Fiser17}.

Patch based methods have many benefits, they can easily capture fine details of the style, will never fail to stylize the output by maintaining photographic high frequencies. However, approaches based on patch synthesis face several challenges. (1) The style may not contain the necessary patches to properly represent the content, (2) convincingly blending neighboring patches can fail and lead to blurry regions or jarring 'seams' between different textures, and (3) correctly matching style patches to the correct location either requires additional effort by users or depends on the quality of the automatically generated guidance. 

\subsection{Neural Style Transfer}
\begin{figure}[H]
    \centering
    \includegraphics[width=\linewidth]{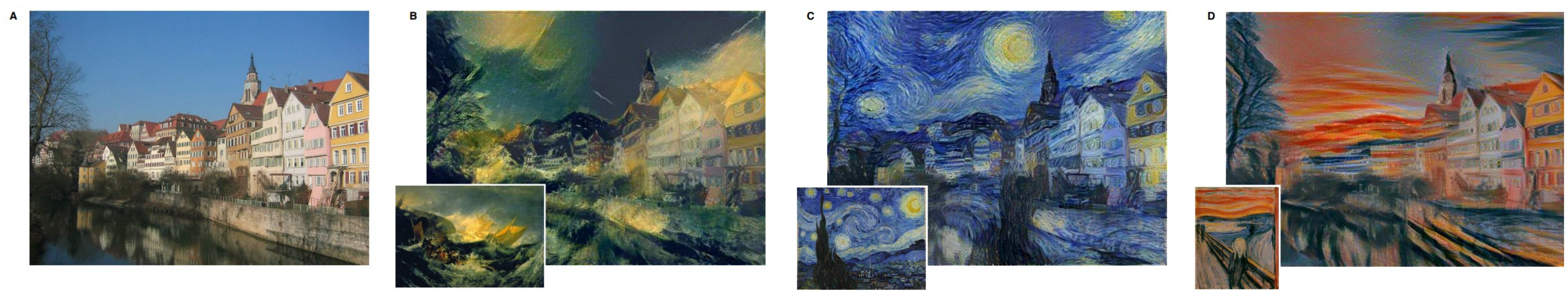}
    \caption{Examples outputs taken from the seminal neural style transfer paper of Gatys et al \cite{Gatys16}. This algorithm marked a departure from previous style transfer algorithms in that it was parametric, matching the statistics of a fixed set of features, and taking motivation from the much earlier work of Portilla and Simoncelli \cite{Portilla00}. Unlike this earlier work the features used by Gatys et al. were not hand-crafted, but were instead learned from scratch by a neural network pre-trained for image classification \cite{simonyan2014very}}
    \label{fig:nst_intro}
\end{figure}
Hoping to address these challenges, a large portion of the style transfer community has turned to developing algorithms that operate in the feature space of a neural network, rather than using the raw pixels of patches or hand-crafted features extracted from same. Not only is it easier to blend features in the hidden state of a neural network rather than in pixel space \cite{li2016combining}, but the higher quality features facilitate better matching between regions of the content and style. The most commonly used neural representations in style transfer are derived from networks pretrained for image classification on photographs. In order to successfully classify the semantic contents of images, these networks learn to produce a hierarchy of features ranging from edges in the early layers, to semantic part detectors in later layers (for example dog faces or human legs). 

The highly influential work of Gatys et al. \cite{gatys2016image} demonstrated that, despite being learned from photographs, these features also capture the texture or 'style' of non-photorealistic artwork. In the spirit of Julesz the sufficient statistics of features taken from the early layers can be used to compute a 'style similarity' loss, and an output optimized to minimize this loss takes on many attributes of the desired style. Beyond this insight, other details of the style transfer algorithm proposed by \cite{gatys2016image} have also had a huge impact. First, in addition to a 'style similarity' term the objective function also contains a 'content similarity' term based on deep layers of the network (which were assumed to capture semantics but not style) which is jointly minimized with the style loss, allowing trading off between content preservation and stylization. Second, \cite{gatys2016image} used a variant of gradient descent to directly optimize the pixels of an output image to minimize the objective function, rather than the previously dominant approach of stitching together small patches from the target style.  This paper sparked the sub-field of 'Neural Style Transfer', an umbrella which covers all of the work I will present as part of my thesis.
\newpage
\section{Thesis Overview}

Since the seminal work of \cite{gatys2016image}, 'Neural Style Transfer' algorithms have split into three major regimes, each sacrificing one of three properties: speed, visual quality, or generalization. Optimization-based algorithms (the same regime originally explored in \cite{gatys2016image}) offer quality and generalization, but are quite slow. Style-specific neural methods offer speed and quality, but overfit to a particular artist/style and do not generalize well. 'Universal' neural methods train on a large dataset of styles and offer generalization and speed, but sacrifice visual quality.  
In Chapter \ref{chpt:background} I give an overview of the research in each of these regimes as well as non-neural style transfer algorithms and earlier research in texture synthesis.

Motivated by the tradeoffs each of these regimes must make, it has been my goal to develop a single style transfer method that is fast, high-quality, and general. I have approached this by first developing optimization-based algorithms with higher visual quality, then replacing optimization with a trained neural network. In an optimization-based framework there is no fitting of an image synthesis model to a training dataset, there is only an objective function based on our two input images ($C$ and $S$) which we seek to minimize by altering the output image $O$. I began by developing algorithms in this regime to focus on developing better quantitative definitions of 'content' and 'style' before determining the many details necessary to successfully train a neural network for image synthesis.

\begin{wrapfigure}{r}{0.48\linewidth}
    \centering
    \vspace{-0.5cm}
    \includegraphics[width=\linewidth]{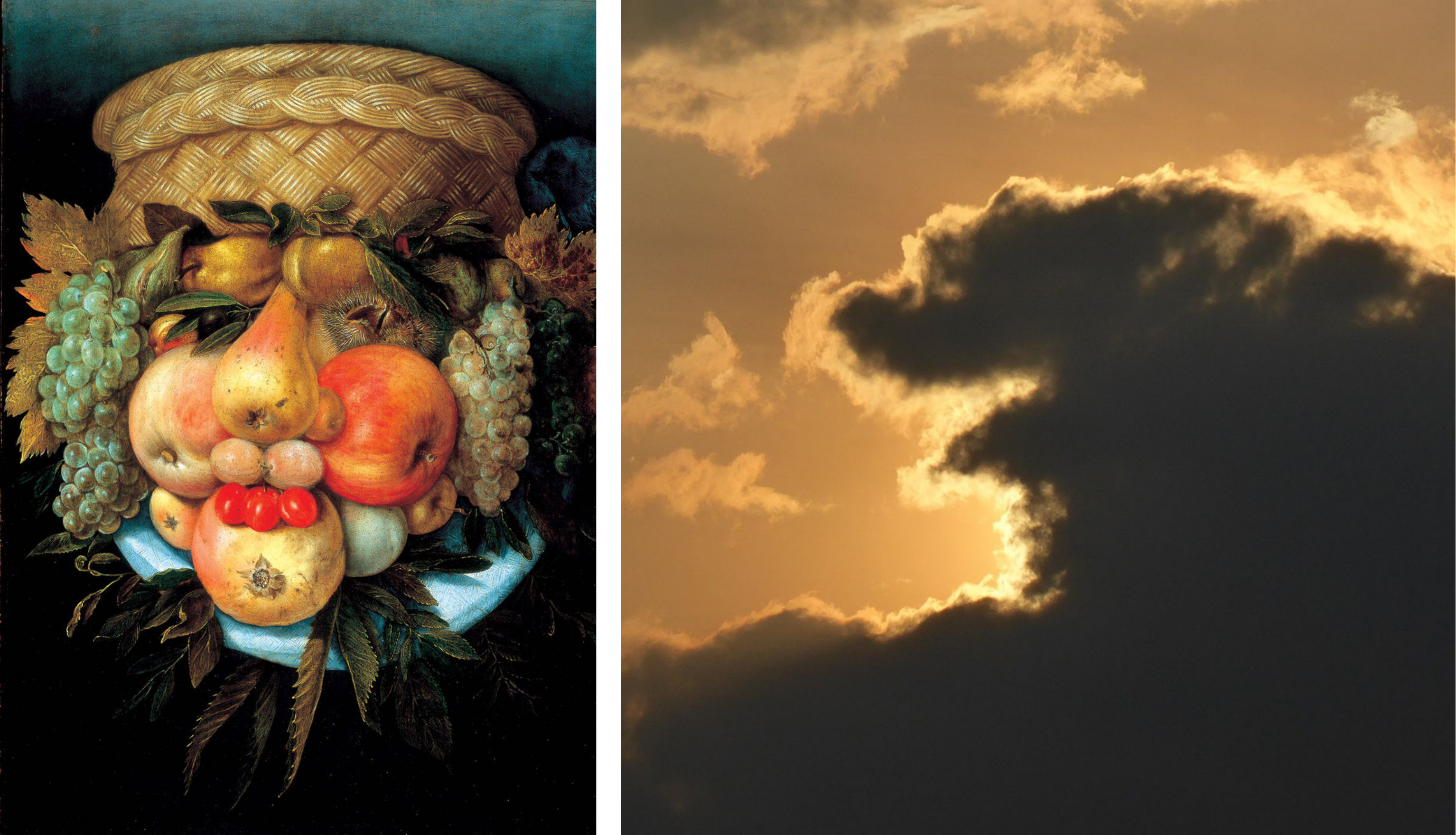}
    \caption{Formulations of 'content' should take into account robust dimensions of human perception. Pictured are examples from art and nature of pareidolia, the human tendency to assign semantic meaning to an ambiguous or random visual pattern.}
    \vspace{-1cm}
    \label{fig:pareid_intro}
\end{wrapfigure}

Chapter \ref{chpt:STROTSS} details my first paper in this area, "Style Transfer by Relaxed Optimal Transport and Self-Similarity" or STROTSS \cite{kolkin2019style}. This work leverages two key ideas which my future work builds upon. The first of these is that the distribution of localized features is too complex to be modelled with simple summary statistics, the same philosophy that lead to patch-driven approaches to texture synthesis. The distinctive textural elements of a style can be modeled far better by explicitly matching local style features to local content features using non-parametric tools such as optimal transport or nearest neighbors. In this work we propose a 'style loss' that measures the distance between two empirical distributions of features using an approximation of the Earth Mover's Distance (EMD). The second key idea is that human perception is extremely robust, and improvements in style transfer's visual quality can be achieved by relaxing definitions of 'content' to take advantage or these robust dimensions of perception. In this work we were inspired by pareidolia (see Figure \ref{fig:pareid_intro}), and propose a content loss based on self-similarity that is invariant to global translations and orthonormal affine transformations of feature space. Together these novel definitions of 'content' and 'style' lead to outputs that not only better match the target style, but also better preserve content by reducing artifacts. Via a large user study we demonstrate that the proposed method offerd a pareto-dominant tradeoff between stylization and content preservation relative to prior work.

However, STROTSS is still quite slow. In our follow-up work "Neural Neighbor Style Transfer" (described in Chapter \ref{chpt:nnst}) we meld techniques from traditional patch synthesis and neural style transfer to develop an algorithm that both produces higher quality visual outputs and is amenable to fast approximation by a neural network. In this work we present a procedure based on nearest neighbors which explicitly constructs a target representation for the stylized output in the feature space of a pretrained neural network. From this representation we describe two methods of retrieving a stylized output image. The first is simply optimizing the pixels of an image until it produces the target representation. The second is training a CNN decoder that can directly synthesize output pixels using the target representation as input. Because the input contains a huge amount of information about the original style image S, we are able to train a neural network that is fast, generalizes well to new styles, and creates aesthetically compelling results. A notable finding of this work is that no explicit content loss is required for high quality style transfer. Because today's style transfer methods primarily focus on copying the texture of the style, content can reliably be preserved via the biased initialization proved by $C$. Via a user study we demonstrate that our fast neural variant can produce outputs of comparable quality to STROTSS, and our optimization based variant surpasses it and sets a new standard of visual quality of general-purpose style transfer.

Both of these works operate in the traditional paradigm of style transfer,  taking texture and style to be essentially interchangeable. To truly develop algorithms that can mimic the abilities of a human artist, we need to broaden our definition of style.  A preliminary effort in this direction is described in Chapter \ref{chpt:dst}, which details our paper "Deformable Style Transfer" \cite{kim2020deformable} (joint work with Sunnie Kim). In this work we propose a framework for learning spatial warp fields which help stylize the output of other algorithms when C and S are well-paired (similar semantics and pose). Perhaps the most notable aspect of this paper is that it extends the definition of 'style' to include form and proportion. Prior work had developed domain-specific approaches to this problem for faces, but in \cite{kim2020deformable} we proposed the first domain-agnostic algorithm. The user study accompanying this paper included two major results. The first was that the warp field produced by our method dramatically improve stylization quality when combined with either STROTSS or the original neural style transfer algorithm of Gatys \cite{gatys2016image}. The second, more interesting, finding was that the deformations produced by our method (which are often fairly substantial) do not dramatically impact perceived content preservation. This work provides evidence that form, like texture, is a rich domain to explore for improving style transfer and makes a first step towards doing so.

Many challenges remain before style transfer can in any way be considered 'solved'. The works presented in this thesis represent tangible progress in improving the ability of 'Neural Style Transfer' algorithms to match a style's textures to the appropriate region of content and synthesize them. However, even if we narrowly consider texture as a proxy for style, there are numerous opportunities for improvement. First, methods driven by patch based synthesis are capable of scaling to extremely high resolution inputs and replicating arbitrarily high frequency features of the style, while neural style transfer algorithms are not. Second, no style transfer method is sufficiently robust, and it can be difficult to predict which combinations of input will result in good or bad outputs. Third, despite the appealing simplicity of viewing a style's texture as local and Markovian, many artworks exhibit textural features, such as brushstrokes, with large spatial extent. These must be better modeled to truly capture an artist's style. This is to say nothing of the many elements of artistic style beyond texture. Our work stylizing the form and proportion of well paired inputs is only a very early step in the very long road towards representing and reproducing stylistic choices such as geometry, subject matter, and composition.
\chapter{Background}
\label{chpt:background}

In this section I will give an overview of the field of artistic style transfer, beginning with its origins in texture transfer and ending with present day. The goal is to provide context for my work, as well as provide readers with a compact entry point for exploring major ideas that have shaped the field. Later chapters will expand on the technical details of the subset of these works which have heavily influenced my own research.

\section{From Texture Analysis To Early Style Transfer}\label{sec:back_tex}
\begin{wrapfigure}{r}{0.48\linewidth}
    \centering
    \vspace{-0.5cm}
    \includegraphics[width=\linewidth]{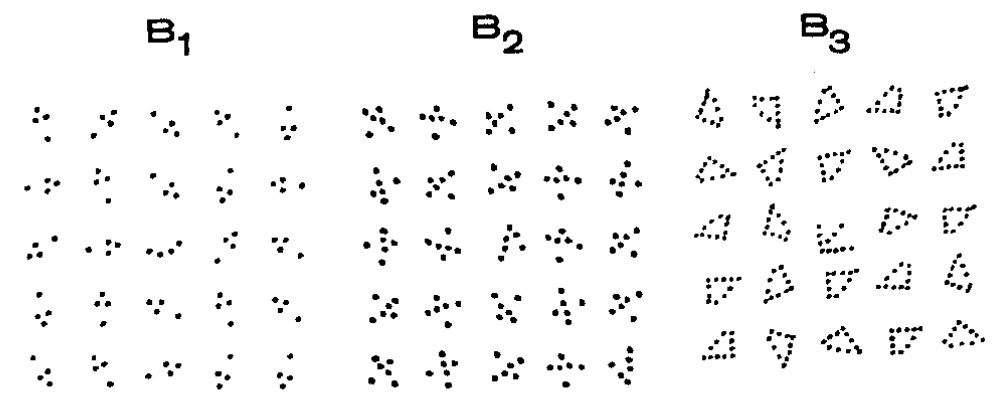}
    \caption{Examples of patterns where pre-attentive discrimination is possible despite shared first and second order statistics. Caelli and Julesz hypothesized that this was due to 'closure', in other words whether or not a pattern is percieved to have an interior. Figure taken from \cite{caelli1978perceptual}}
    \vspace{-1cm}
    \label{fig:julesz_closure}
\end{wrapfigure}

The roots of many algorithms and key ideas in style transfer can be traced to earlier work in texture analysis and texture synthesis. Just as the style transfer community has grappled with defining 'style' and 'content', the texture analysis and synthesis community has grappled with defining 'texture'. Probably the most influential in this effort has been Bela Julesz, who studied humans' pre-attentive texture discrimination abilities and designed algorithms to automatically generate textural stimuli \cite{julesz1965texture, julesz1975experiments}. He theorized that humans relied primarily on the first and second order statistics of visual features to quickly discriminate between textures (exactly defining these features would be a central challenge of later work). While he found that this hypothesis held for many synthetic examples (see Figure \ref{fig:julesz_intro}), in subsequent work he and others  found counterexamples pre-attentive discrimination based on the higher order property of 'closure' was possible \cite{caelli1978perceptual, julesz1981textons} (see Figure \ref{fig:julesz_closure}).

While early efforts approached texture synthesis by simulating the physical phenomena underlying their real world appearance \cite{witkin1991reaction, turk1991generating, fournier1982computer, lewis1989algorithms} (requiring separate algorithms to be developed for different types of texture), later data-driven approaches took their cue from Julesz, often jointly studying the problem of texture synthesis and texture analysis. In this regime it was important not only to synthesize convincing textures perceptually similar to an exemplar, but also develop compact descriptors of texture that could offer insight into textural discrimination and perception. Many works of this flavor took inspiration from what was known of the early visual processing system in mammals, designing features based on oriented linear kernels at multiple spatial scales \cite{bergen1988early,malik1990preattentive,turner1986texture} (for examples of such kernels, see Figure \ref{fig:steer_pyr}).

\begin{wrapfigure}{l}{0.48\linewidth}
    \centering
    \vspace{-0.5cm}
    \includegraphics[width=\linewidth]{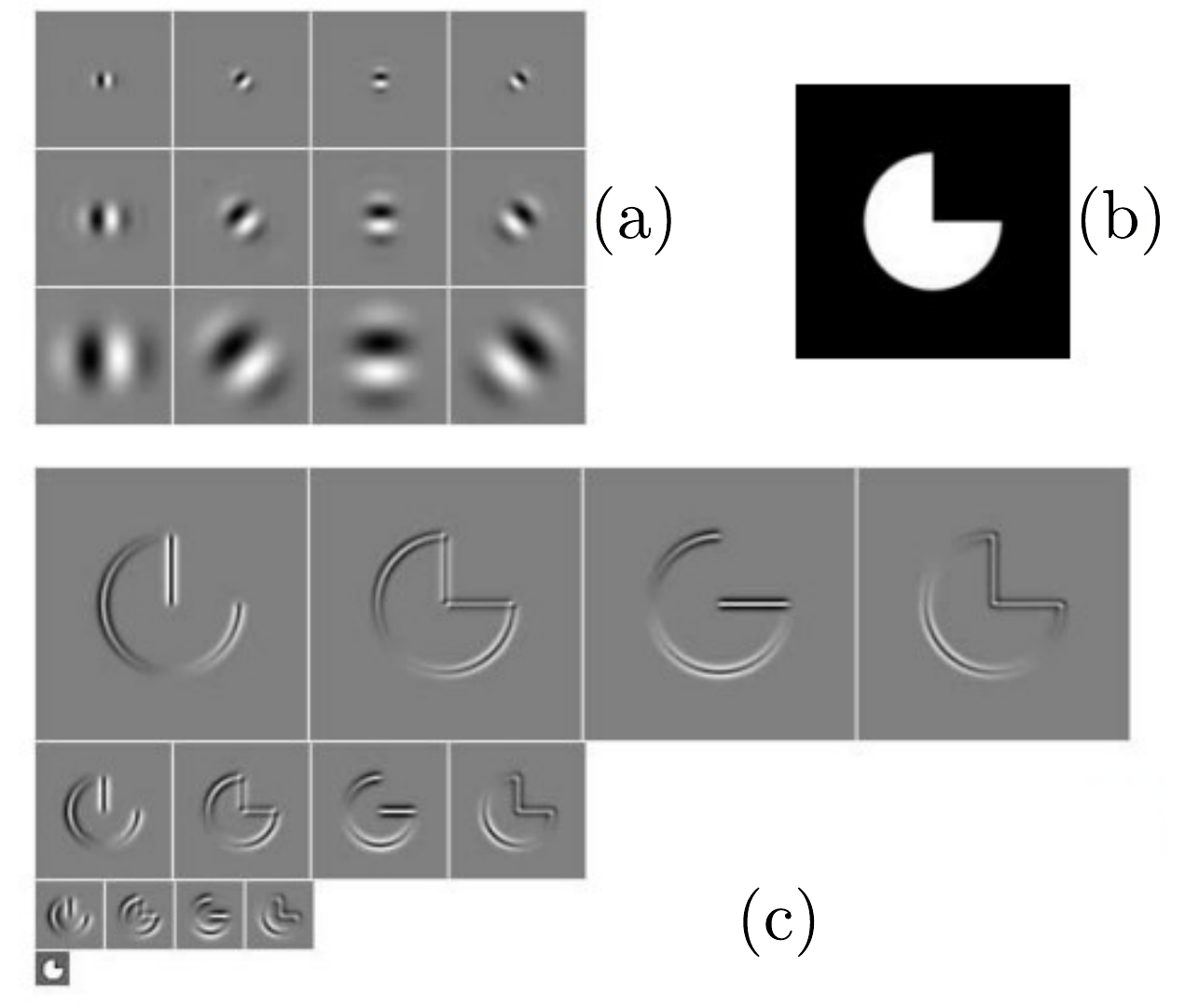}
    \caption{Example of (a) filter kernels from a steerable filter pyramid, (b) an input image, and (c) filter responses when the kernels of (a) are applied to (b). Figure taken from \cite{heeger1995pyramid}.}
    \vspace{-0.5cm}
    \label{fig:steer_pyr}
\end{wrapfigure}
\begin{figure}[htp]
    \centering
    \includegraphics[width=\linewidth]{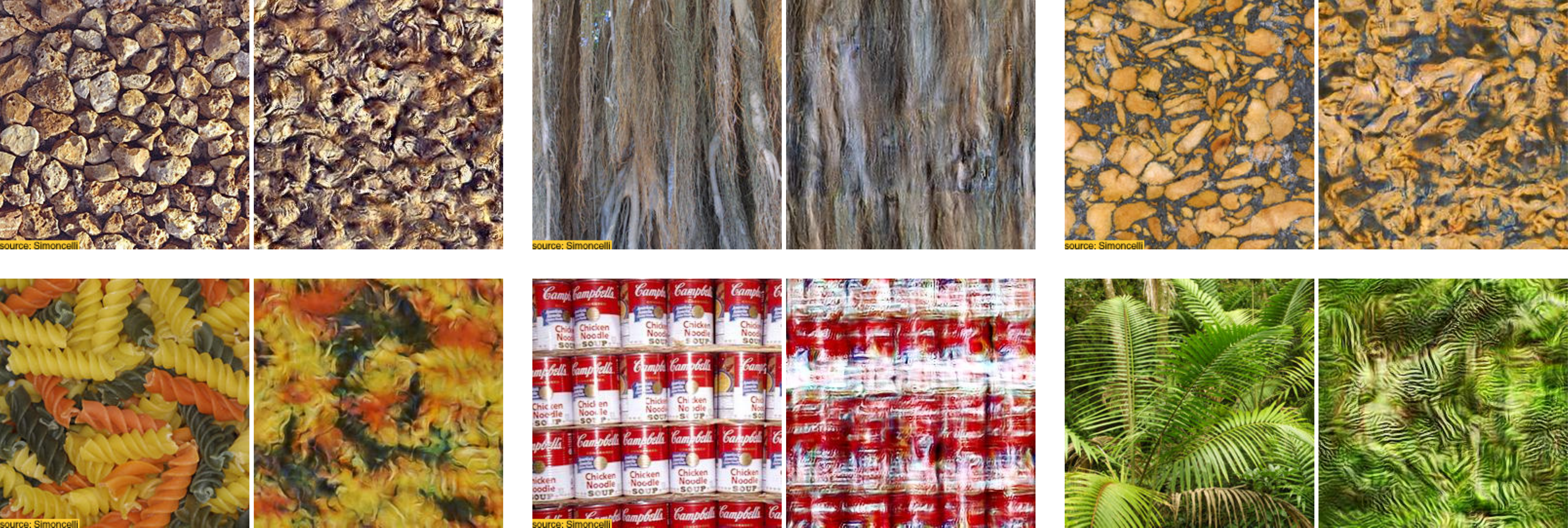}
    \caption{Each pair of images consists of the original texture exemplar on the left, and the synthesized output of \cite{Portilla00} on the right. Textures that do not contain any complex long range structures can be fairly well modeled by parametric texture synthesis models such as \cite{Portilla00} (top row). However, textures with long range dependencies, or composed of perceptually distinct objects, cannot (bottom row). Images taken from the \href{http://www.cns.nyu.edu/~lcv/texture/color/}{project page} of \cite{Portilla00}.}
    \label{fig:og_tex_dist}
\end{figure}

Heeger and Bergin \cite{heeger1995pyramid} were among the earliest to design a texture synthesis algorithm of this type; generating textures by starting from an image filled with random noise, constructing from it a pyramid of steerable filters, then independently matching the 1d histogram of pyramid coefficients for each oriented filter to the corresponding histogram of coefficients extracted from the texture exemplar. Zhu et al. \cite{Zhu98frame} provided a formal framework encompassing and extending the algorithm of \cite{heeger1995pyramid}, proving that an uncountable number of marginals over feature coefficients could define an arbitrary distribution of image textures. They also proposed an algorithm based on Gibbs sampling that allowed synthesis based on non-linear features, and an algorithm for feature selection given a texture exemplar which reduced the tremendous computational cost of their synthesis algorithm. Portilla and Simoncelli \cite{Portilla00} proposed a more efficient synthesis algorithm, essentially using gradient descent with exact line search on the output image's pixels to sequentially satisfy constraints on the output image's feature distribution. They also proposed a universal set of 710 textural feature constraints based on wavelet coefficient marginals, local wavelet auto-correlations, local wavelet magnitude auto-correlations, and cross-scale phase statistics. This effort provided a formalized starting point for evaluating  the original hypothesis of Julesz, that texture could be discriminated based on the Nth-Order statistics (for small N) of a limited set of visual features. Despite the intellectual appeal of these methods, their reliance on a limited number of hand-crafted features makes them poorly suited to synthesizing textures with complex structure (see Figure \ref{fig:og_tex_dist}). This limitation would eventually inspired Gatys et al. \cite{gatys2015texture} to propose a spiritually similar algorithm for texture synthesis using features extracted by a pre-trained neural network. A refinement of Gatys' texture synthesis algorithm would become the first 'neural style transfer' algorithm \cite{gatys2016image}.

\begin{figure}[htp]
    \centering
    \includegraphics[width=\linewidth]{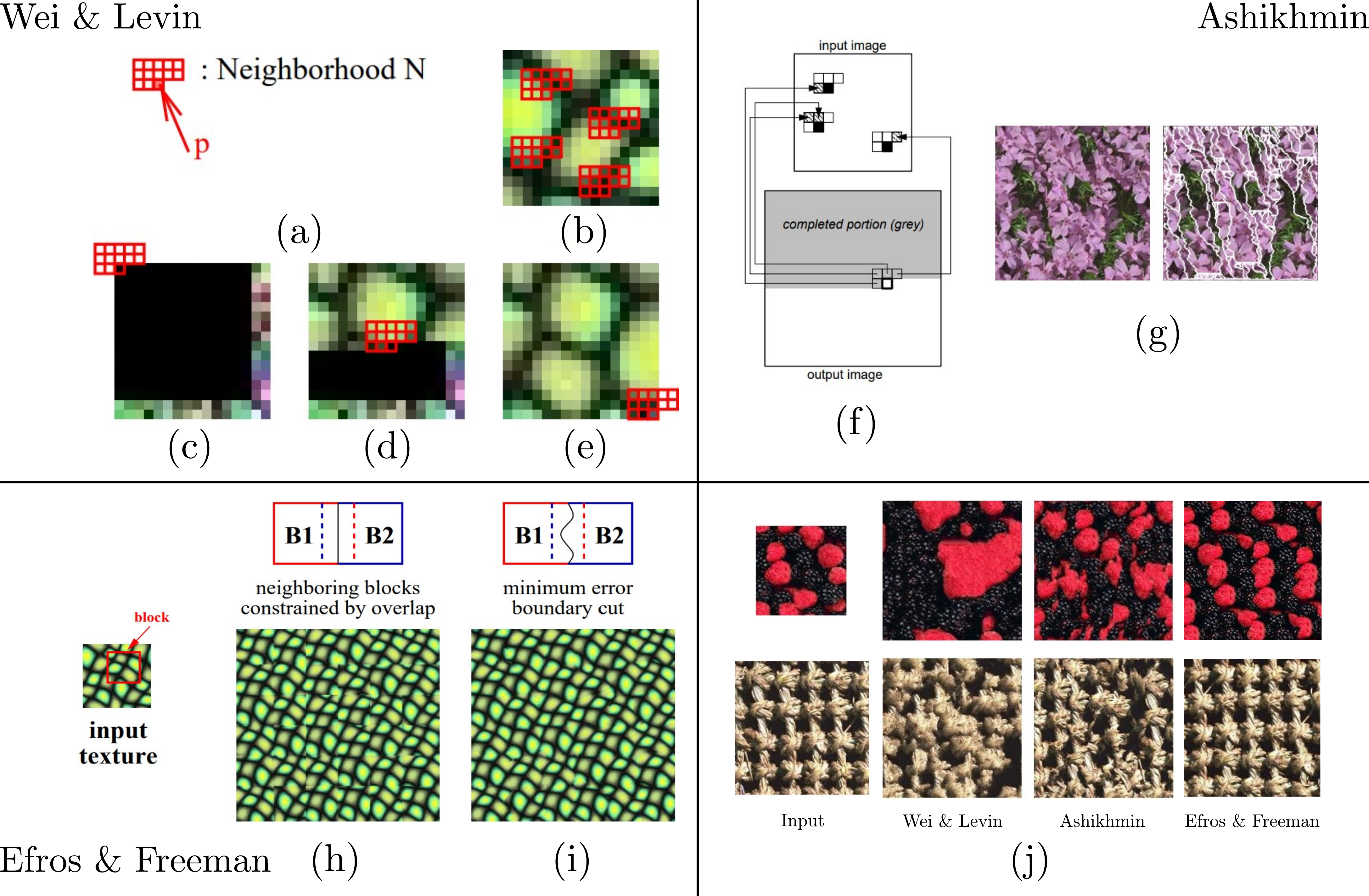}
    \caption{Illustration of some major ideas in non-parametric texture synthesis (a-i) and their qualitative effect (j). Wei and Levin \cite{wei2000fast} proposed a multi-scale auto-regressive texture synthesis model. In this model the next pixel $p$ is based on a set of candidate neighborhoods of shape N taken from the texture exemplar (b). All possible candidates are compared to the local neighborhood of texture synthesized so far. (c) visualizes the neighborhood of the first pixel sampled, note that pixels beyond the object boundary are assumed to be black. (d) visualizes the neighborhood of a pixel in the middle of the synthesis process, and (e) visualizes the neighborhood of the last pixel sampled. Ashikhmin \cite{Ashikhmin01} takes advantage of the fact that the exact spatial index of the pixels taken from the input are known, and used this to only consider candidate neighborhoods which correspond to contiguous regions of the input texture (f), this results in the output being composed of irregular congiguous regions of the exemplar, (g, right) visualizes the border of these regions in a synthesized output. Efros and Freeman \cite{efros2001image} propose an algorithm that is very similar to Wei and Levin, except that blocks/patches of pixels are sampled jointly (h), rather than sampling one pixel at a time. The sampled patches are then joined along a minimum error boundary cut that minimized discontinuity (i). Synthesizing contiguous regions of the exemplar jointly leads to synthesized results that better capture discrete structures and long range correlations in the exemplar (j). Figures taken form \cite{wei2000fast}, \cite{Ashikhmin01}, \cite{efros2001image}}
    \label{fig:auto_syn}
\end{figure}

Other efforts were more concerned with efficiency and synthesis quality than interpretable representations of texture. To better model higher order dependence between features De Bonet \cite{debonnet1997} proposed matching the conditional distribution of the finest level of pyramid coefficients based on all 'parent' coefficients (i.e. pyramid coefficients at roughly the same spatial location in coarser levels), leading to an auto-regressive coarse to fine synthesis algorithm. Many later works would propose refined auto-regressive synthesis strategies. Efros and Leung \cite{efros1999texture} proposed sampling the output pixel-by-pixel at a single scale by conditioning on a large patch 'on the scale of the biggest regular feature', and iteratively 'growing' the synthesized texture from a small patch randomly taken from the exemplar. Wei and Levin \cite{wei2000fast} proposed another auto-regressive procedure with elements of the approaches taken by \cite{debonnet1997} and \cite{efros1999texture}, synthesizing the texture coarse to fine, but within a scale 
synthesizing in raster-order; conditioning on both context from the coarser scale and on already synthesized pixels within the same scale. Ashikhmin \cite{Ashikhmin01} extended the algorithm of \cite{wei2000fast}, and proposed replacing a global search for matching conditioning structures in the texture exemplar with selecting regions with a consistent offset to those already sampled, resulting in synthesized textures consisting of large irregular regions of the texture exemplar. Harrison \cite{harrison2001non} proposed a more principled method for choosing the order in which pixels should be sampled, hypothesizing that the sampling order should be based on the minimizing the entropy of the pixel distribution to be sampled (sample pixels which will be well explained by the synthesized texture so far) normalized by entropy of the neighborhood sampled from in the texture exemplar (to prevent only sampling texture from highly structured regions of the texture). Hertzmann et al. \cite{hertzmann2001image} proposed an extension of the pixel-by-pixel method of \cite{Ashikhmin01} which allowed trading off between matching coherent regions as proposed by Ashikhmin (i.e. cloning large regions of the exemplar, which preserved details well but could introduce seams between cloned regions) and the l2 appearence matching of \cite{wei2000fast} (which was less prone to seams, but synthesized blurrier results), in many cases achieving results with the best properties of both algorithms.

Several influential works proposed making the atomic unit of synthesis patches rather than pixels. This has benefits both in terms of efficiency and synthesis quality, since many local feature statistics are implicitly preserved by copying small patches of the texture exemplar wholesale. The earliest effort in this direction which I'm aware of is the simple but highly effective 'Chaos Mosaic' algorithm \cite{Guo2000ChaosMF}, originally designed as to be a computationally efficient method with a small memory footprint to generate procedural textures for 3d meshes. The algorithm began with a simple tiling of the exemplar texture, then cloning and translating small patches according to 'Arnold's Cat Map' (a chaotic map on coordinate which places the translated blocks in an irregular pattern that appears to be random but is in fact deterministic), the edges of the translated blocks are then blurred with simple cross edge smoothing. 
\begin{wrapfigure}{r}{0.48\linewidth}
    \centering
    \includegraphics[width=\linewidth]{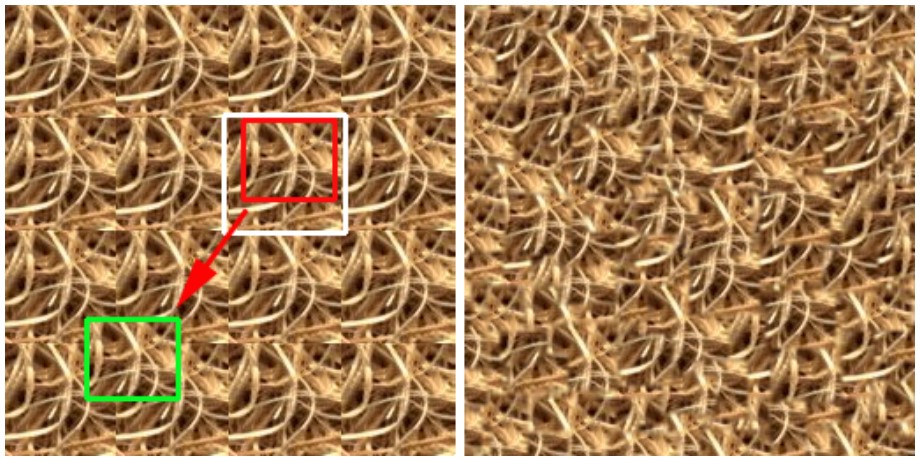}
    \caption{Illustration of the first patch based synthesis algorithm, the 'Chaos Mosaic'. The algorithm begins with a simple tiling of the exemplar texture (left, white box), then cloning and translating small patches (left, red and green boxes) according to 'Arnold's Cat Map'. After translation the patch and background are blended with cross edge smoothing. Repeating this procedure many times (both repeatedly translating and for new random patches) gives the final texture (right). Figure taken from \cite{Guo2000ChaosMF}.}
    \vspace{-0.5cm}
    \label{fig:steer_pyr}
\end{wrapfigure}
Efros and Freeman \cite{efros2001image} and Liang et al. \cite{Liang01} both propose algorithms similar to \cite{efros1999texture} in that they synthesize the the output texture in raster order and condition on a region of overlap with the texture synthesized so far. However, instead of sampling each pixel independently, entire patches are sampled at once. If we consider the size of patches sampled jointly to be the algorithm's 'stride' then the algorithm of \cite{efros1999texture} can be considered to be these algorithms with a 'stride' of one. When the 'stride' is greater than one the question arises of how to blend new patches with those selected so far.  In \cite{Liang01} overlapping patches are blended using cross edge smoothing (i.e. feathering). To avoid synthesizing overly smooth textures \cite{efros2001image}  propose joining the patches along a minimum error boundary cut, which finds the (often irregular) discontinuity minimizing edge to switch between patches.

Pixel-by-pixel and patch-by-patch based methods simultaneously made their first forays into style transfer in SIGGRAPH 2001. Efros and Freeman \cite{efros2001image} proposed extending their patch based method by augmenting a 'content' and 'style' images (referred to a 'target image' and 'source texture' in their work) with correspondence maps that were relatively invariant to style and made it easier to match patches of texture onto sensible areas of content (they used a blurred luminance channel as their correspondence map, but also suggested using other derived features such as local orientation angles). Data-driven image modification was the primary focus of Hertzmann et al. \cite{Hertzmann01}, in which the framework of 'image analogies' was proposed. In essence the approach of 'image analogies' is quite similar to that of \cite{efros2001image}, and suggests improving the matches between a 'target image' and 'source texture' by providing an 'unfiltered' version of the 'source texture' which provides a better feature space for finding matches with the 'target image'. In the ideal case an 'unfiltered source texture' might be a photograph, and the 'source texture' would be a painting or drawing of the photograph's contents (with the exact same viewpoint, lighting, etc.). They demonstrated that this approach allowed the core algorithms of texture synthesis to be applied many image processing tasks, including colorization, super-resolution, and style transfer.

\section{Non-Neural Style Transfer} \label{sec:back_early}

From this point, research began to emerge that focused on style transfer as an independent task, primarily taking inspiration from \cite{Hertzmann01} and refining the framework of 'image analogies'. Rosales et al. \cite{rosales2003unsupervised} relax the 'supervised' requirement of 'image analogies' (requiring an 'unfiltered style image'). They did so by replace the greedy synthesis algorithms common in texture synthesis (which iteratively chose a pixel/patch from the texture/style image to place at each location) with minimizing the energy of a Markov Random Field (MRF) over patch assignments via loopy belief propagation (where unary potentials describe the style patches' distances to a content patch, and binary potentials describe the distance between a selected style patch and a neighboring selected style patch in regions of overlap). This more sophisticated matching algorithm made patch assignments robust enough that the 'unfiltered style' was no longer needed as guidance. Cheng et al. \cite{cheng2008consistent} used the MRF formulation of \cite{rosales2003unsupervised}, but remained the the regime of 'supervised' style transfer \cite{hertzmann2001image} where access to an unfiltered style image was assumed. They improve patch assignments by using metric learning to find a smooth distance function which minimized the distance between matching patches in the filtered and unfiltered style images, then using this data-driven metric to choose which patches to use when stylizing the content image. This work also proposed an elegant implementation detail which I have not seen elsewhere, recommending matching octagonal patches instead of square ones, ensuring that each region of overlap need only harmonize two patches (as opposed to square patches which must harmonize 4 patches at the corners). In addition the overlap of octagonal patches tapers to a point at either end, making them well suited to being joined using the minimum error boundary cut of \cite{efros2001image}. Barnes et al. \cite{Barnes15} focused on improving the computational efficiency of \cite{hertzmann2001image}, and proposed a pre-computed data structure that supports efficiently querying similarity with a predetermined library of patches. Zhang et al. \cite{zhang2013style} use feature engineering to improve the unsupervised algorithm of \cite{rosales2003unsupervised} by decomposing the style image into 'segmentation', 'paint', and 'edge' components (essentially low, middle, and high frequencies), then matching patches from these components independently. Frigola et al. \cite{frigo2016split} also build upon the framework of \cite{rosales2003unsupervised}, but do so by extending it to handle a MRF over patches of varying sizes, noting that often flat regions of a photograph (such as sky) can be matched to large regions of the style images (the large swirls of Starry Night), but details must be matched to small patches (individual brushstrokes).

\begin{figure}[htp]
    \centering
    \includegraphics[width=\linewidth]{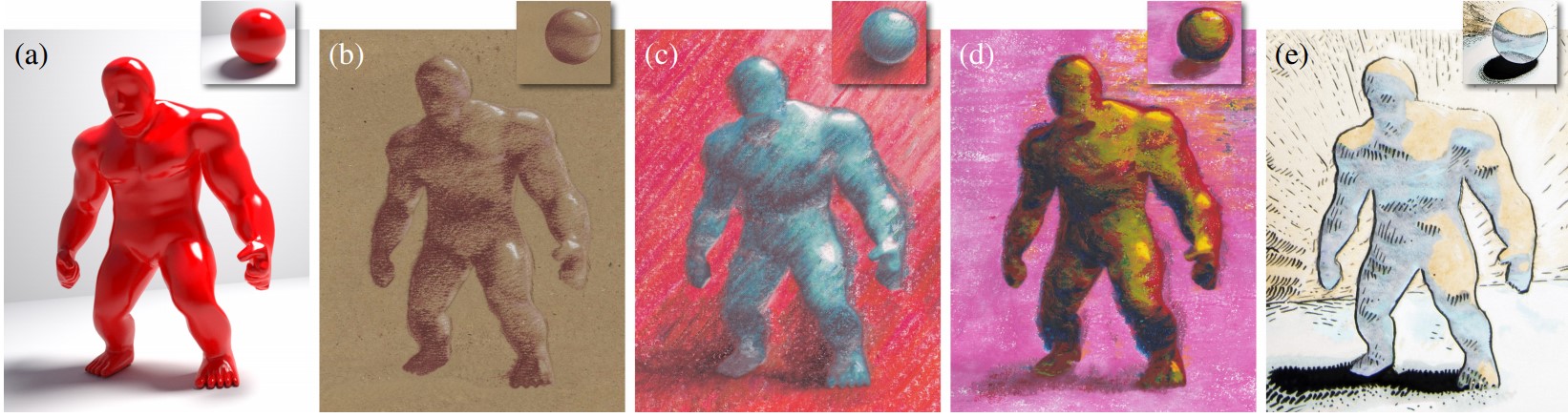}
    \caption{Illustration of data-driven 3d rendering using the 'lit sphere' paradigm. (a) is a rendered 3d model of a golem and the top-right inset is a rendering of a sphere under the same lighting conditions. In (b-e) an artist's manual stylization of the sphere is shown in the top-right inset. This style can then be propagated to an arbitrary 3d model (e.g. the golem). Figure taken from \cite{fivser2016stylit}.}
    \label{fig:auto_syn}
\end{figure}
A related area of research is data-driven 3d rendering, where a 2d rendering of a 3d model undergoes stylization. Unlike style transfer for arbitrary 2d images, in this case additional information, namely ground truth geometry and lighting, is available because the content image is produced using 3d modeling software. The most influential paradigm in this problem is the 'lit sphere' method proposed by Sloan et al. \cite{Sloan01}. In this framework an artist draws or paints over a 3d rendering of a sphere. Because a sphere contains all surface angles visible to a viewer the artist's stylized rendering can be propagated to an arbitrary 3d model under similar lighting conditions. This approach is spiritually  similar to the 'image analogies' paradigm of \cite{hertzmann2001image}, in that the original depth and lighting of the sphere is used as an 'unfiltered' source of features to match patches between the 3d model the artist's rendering. Later work by Fiser et al. \cite{fivser2016stylit} exploited this similarity and improved the method by incorporating improvements in patch assignment using Markov Random Fields similar to those used in style transfer \cite{rosales2003unsupervised}. A follow up paper by Fiser et al. \cite{Fiser17} speeds up the algorithm of \cite{fivser2016stylit} by sparsely searching for matches between the 3d model and the sphere, then filling in the gaps by growing the patch of sphere used until it deviates from the target 3d model's geometry by a significant margin.

While these methods can produce compelling results, they tended to either require high-quality guidance (as in the case of 3d rendering), or only work well for a limited set of painterly styles. Neural style transfer methods leverage the rich feature representations learned by networks pre-trained for tasks such as image classification. This has lead to results of higher visual quality in the regime of 'unsupervised' style transfer, where only the raw content and style images are available.

\section{Neural Image Representations and Their Visualization}

Many of the last decade's greatest successes in computer vision have been driven by Convolutional Neural Networks (CNNs) \cite{fukushima1982neocognitron, lecun1990handwritten, krizhevsky2012imagenet}. Despite difficulty theoretically justifying their ability to generalize beyond the training set, there is no doubt that massive datasets, coupled with efficient training using GPUs, have made it possible to fit discriminative and generative image models with historically unprecedented capabilities. A fundamental property of CNNs' is their locality bias (features at each layer are computed based on features from the previous layer within a small receptive field) and approximate translation equivariance (the parameters of each linear layer are shared across spatial positions, although boundary effects and commonly used pooling layers break exact translation equivariance \cite{zhang2019making}). Their constrained structure leads CNNs to learn a hierarchy of feature extractors useful for minimizing an objective function (for example correctly classifying an input image). Typically this hierarchy of features entangles spatial extent and semantic complexity (early layers contain simple features with small spatial extent such as edges or corners, and deep layers contain complex features with large spatial extent such as face detectors), although there have been efforts to design architectures which break the entanglement of features' spatial extent and complexity \cite{ke2017multigrid}. 

\begin{figure}[htp]
    \centering
    \includegraphics[width=\linewidth]{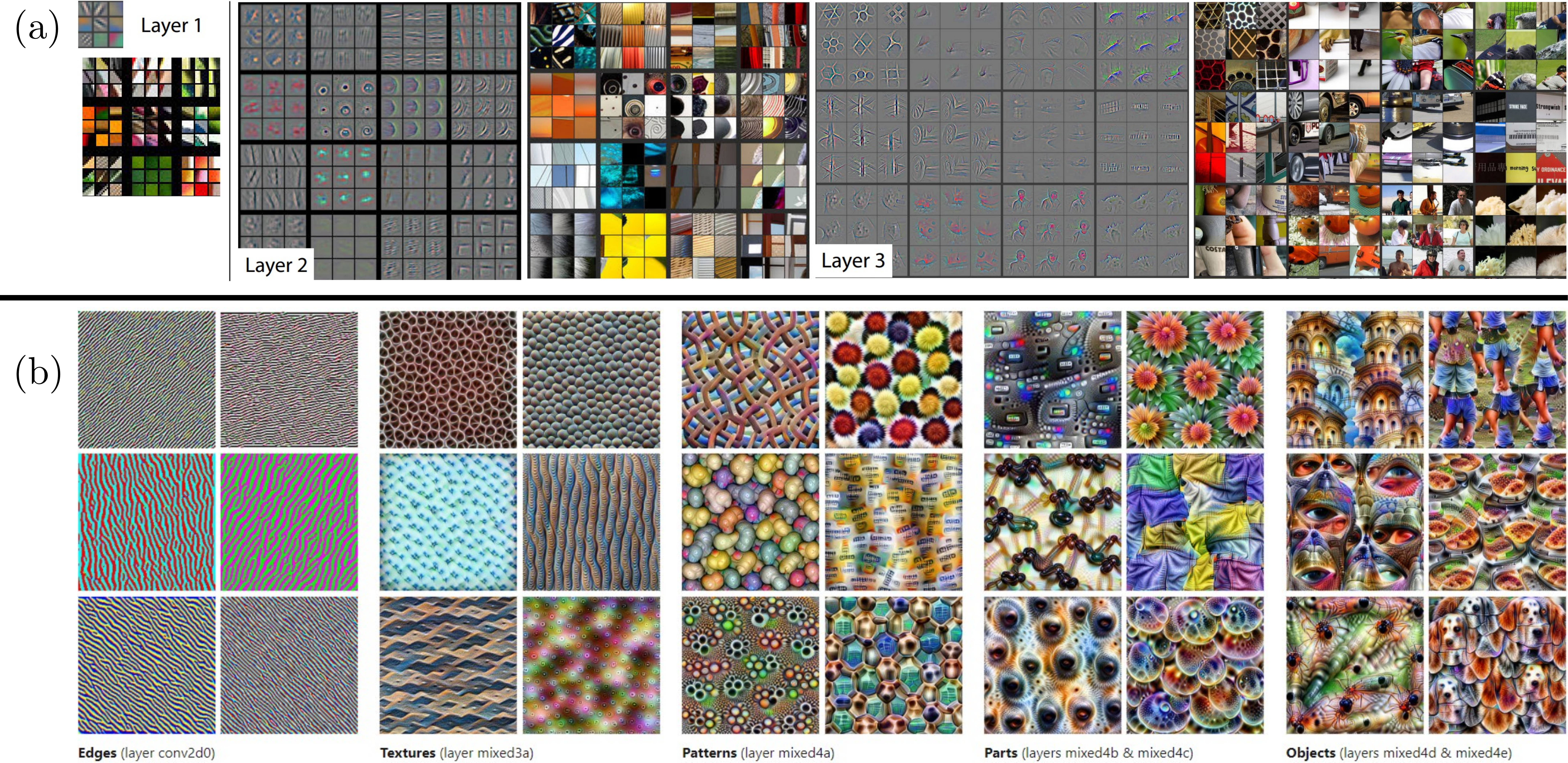}
    \caption{Visualization strategies for learned CNN features. (a) shows image patches which highly activate filters in a particular layer, alongside visualizations generated by approximately inverting the network of which aspects of the patch activated the filter  \cite{zeiler2014visualizing}. (b) shows images optimized from scratch using gradient descent to maximize the filter activations at a given layer \cite{olah2017feature}. In a similar spirit to (b) optimization-based style transfer algorithms optimize the pixels of an output to activate filters corresponding to a stylized image. Figures taken from \cite{zeiler2014visualizing} and \cite{olah2017feature}.}
    \label{fig:auto_syn}
\end{figure}

How do we know what is captured by specific features in the deep layers of a CNN? While it is trivial to visualize the linear filters in the first layer, its less obvious how we can do so for features in later layers. One method is to pass a large number of images through a CNN, and keep track of the patches which trigger high activation values for a given filter, then inspect them and qualitatively identify their shared attributes \cite{zeiler2014visualizing, selvaraju2017grad}. Another method, which has had a much greater impact on the field of style transfer, is to produce the 'stimuli' which maximizes the activation of a certain feature by using gradient descent to directly optimize the pixels of an input image \cite{erhan2009visualizing, simonyan2013deep, zeiler2014visualizing, olah2017feature}. This same technique can be used to 'invert' the feature representation extracted by a CNN from a particular image, allowing some qualitative intuition about which image details are maintained or discarded at different depths of the CNN \cite{mahendran2015understanding}.

Optimization-based neural style transfer algorithms use a similar technique, in that they directly optimize the pixels of an output image using gradient descent. However, rather than optimizing the image to maximize an individual feature activation, or recreate an image that already exist, these algorithms aim to generate novel images by manipulating the representations produced by pretrained neural networks to alter 'style' while maintaining 'content'.

\section{Optimization Based Neural Style Transfer}

\begin{figure}[htp]
    \centering
    \includegraphics[width=\linewidth]{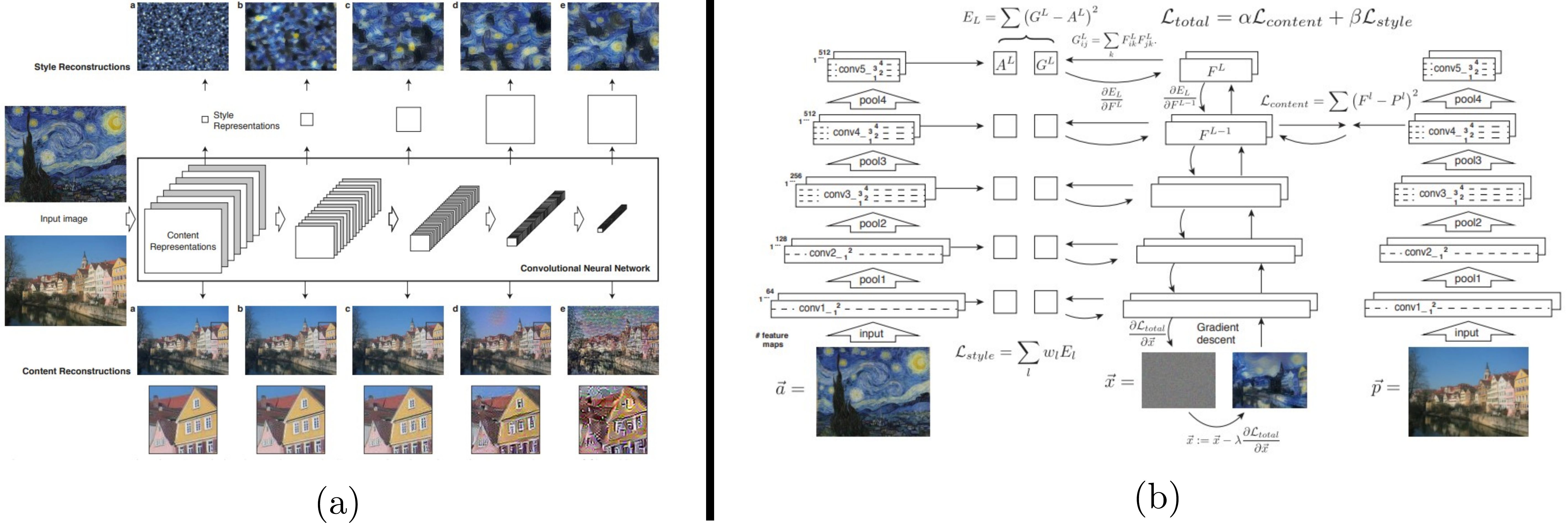}
    \caption{Illustration of major ideas in neural style transfer introduced by Gatys et al \cite{gatys2016image}. (a) visualizes the proposed representations of content (neural network activations) and style (gram matrices constructed from neural network activations aggregated over spatial indexes). (b) visualizes the stylization procedure, in which an output image is optimized to have the desired content and style representation (taken from the content and style image respectively). For more detailed discussion see Section \ref{sec:strotss_gatys}. Figures taken from \cite{gatys2016image}.}
    \label{fig:auto_syn}
\end{figure}

`Neural Style Transfer' was introduced by Gatys et al. in `A Neural Algorithm of Artistic Style' ~\cite{gatys2016image} , and represented a departure from contemporary stylization algorithms in two major ways. First, eschewing hand-crafted features in favor of those extracted by convolutional neural network pretrained for image classification (typically VGG \cite{simonyan2014very}). Second, synthesizing the stylized output by directly optimizing pixels using gradient descent, rather than by stitching together patches of the style image like previous work.

In addition this work proposed an influential framework for designing style transfer objective functions. In this framework (described in detail in Section \ref{sec:strotss_gatys}) objective functions consist of two terms, a `style loss' $\mathcal{L}_s(O,S)$ which measures the stylistic similarity between the output image $O$ and the style image $S$, and a `content loss' $\mathcal{L}_c(O,C)$ which measures the similarity in semantic content and layout between $O$ and the content image $C$. The original objective function proposed by ~\cite{gatys2016image} defined style loss based on the Gram matrices of features extracted from early layers (i.e. the local correlation or co-occurrence of simple visual features with small spatial extent). The content loss was based on the difference between the output image and the style image in the feature space of a deeper layer of the same network. The intuition being that, to facilitate generalization, features deep in an image classification network should be stable under changes to local texture or color. Proposing alternate definitions for one or both of these terms, in other words refining the quantitative definitions of content and style, has been one of the main forms of innovation in subsequent work.

For example, in order to capture long-range spatial dependencies Berger et al. ~\cite{berger2016incorporating} propose computing multiple Gram matrices using translated feature tensors (so that the outer product is taken between feature vectors at fixed spatial offsets). While style losses based on Gram matrices can be seen related to capturing the second order statistics of features (the covariance matrix is proportional to the gram matrix after data is zero-centered), there have been efforts to use more complex parameterizations of the style features' distribution. For example Risser et al. ~\cite{risser2017stable} propose matching the marginal distribution of features using 1d histograms, and our own work (described in Chapter \ref{chpt:STROTSS}) proposed matching the distribution of features using optimal transport.
\begin{landscape}
\begin{table}[]
    \centering
    \begin{tabular}{c|c|c|c}
        \textbf{Methods} & \textbf{Gatys} \cite{gatys2016image} & \textbf{Chen}  \cite{chen2016fast} & \textbf{CNNMRF} \cite{li2016combining}  \\ \hline \hline
        \textbf{Optimization Variables} & Pixels &  Pixels & Pixels \\ \hline
        \textbf{Optimization Algorithm} & L-BFGS &  Adam & L-BFGS \\ \hline
        \textbf{Multiscale} & No & No & Yes\\ \hline
        \textbf{Content Features} & c4\_2 & N/A & c4\_2\\ \hline
        \textbf{Content Loss} & $\ell_2$ & N/A & $\ell_2$\\ \hline
        \textbf{Style Features} &c1\_1,c2\_1,c3\_1,c4\_1,c5\_1 &  c3\_1(3x3) & c3\_1, c4\_1(3x3) \\ \hline
        \textbf{Style Loss} & Gram Loss & $\ell_2$ between NN & $\ell_2$ between NN\\ \hline
        \textbf{Style Image Augmentation} & N/A & N/A & Scales/Rotations \\ \hline
        \textbf{Regularizer} & N/A & TV & TV \vspace{0.2cm}\\  \hline \hline 

        \textbf{Methods} & \textbf{Contextual}\cite{mechrez2018contextual} & \textbf{STROTSS} (Chapter \ref{chpt:STROTSS}) & \textbf{NNST} (Chapter \ref{chpt:nnst}) \\ \hline \hline
        \textbf{Optimization Variables} & Pixels & Laplacian Pyramid & Laplacian Pyramid\\ \hline
        \textbf{Optimization Algorithm} & L-BFGS & RMS-Prop & Adam\\ \hline
        \textbf{Multiscale} & No & Yes & Yes\\ \hline
        \textbf{Content Features} & c4\_2  & c1\_1...c4\_1, c5\_1 & N/A \\ \hline
        \textbf{Content Loss} & $\mathcal{L}_{CX}$ & Self Sim. &  N/A\\ \hline
        \textbf{Style Features}& c2\_2, c3\_2, c4\_2 & c1\_1...c4\_1, c5\_1 & c1\_1...c4\_3 \\ \hline
        \textbf{Style Loss} & $\mathcal{L}_{CX}$ & REMD & Centered Cosine between NN \\ \hline
        \textbf{Regularizer} & N/A & N/A & N/A\\ \hline

    \end{tabular}
    \caption{Taxonomy of optimization-based style transfer algorithms based on directly optimizing the output image. While the algorithms outlined in the table above produce very different visual results, they are all composed of the same building blocks (finding the output image which minimizes the style and/or content loss). All algorithms use features taken from various convolutional layers of VGG, which are denoted as cX\_Y, where X indexes the conv. block (layers at a particular scale), and Y indexes the layer within a block. $\mathcal{L}_{CX}$: A variant of minimizing loss between nearest neighbor content and style features ($\ell_2$ between NN) which up-weights minimizing the distance between uniquely well matched neighbors (i.e. photographic nose to painted nose) and down-weights matching nearest neighbors with many similarly close neighbors (i.e. homogeneous region to homogeneous region), see \cite{mechrez2018contextual} for details. NN: Nearest Neighbors.}
    \label{tab:taxon}
\end{table}
\end{landscape}

Another common direction for redefining the style loss hearkens back to earlier patch-based methods. These methods explicitly match vectors of neural features (or 'patches' of such vectors) extracted from the style to vectors extracted from the content, creating concrete 'target features' for every spatial location in the output. Then the output image is optimized to produce this representation. Li et al. \cite{li2016combining} replaces the style loss of
\cite{gatys2016image} with minimizing the distance between each patch of output features and their current nearest neighbor patch of style features. 
\begin{wrapfigure}{r}{0.48\linewidth}
    \centering
    \includegraphics[width=\linewidth]{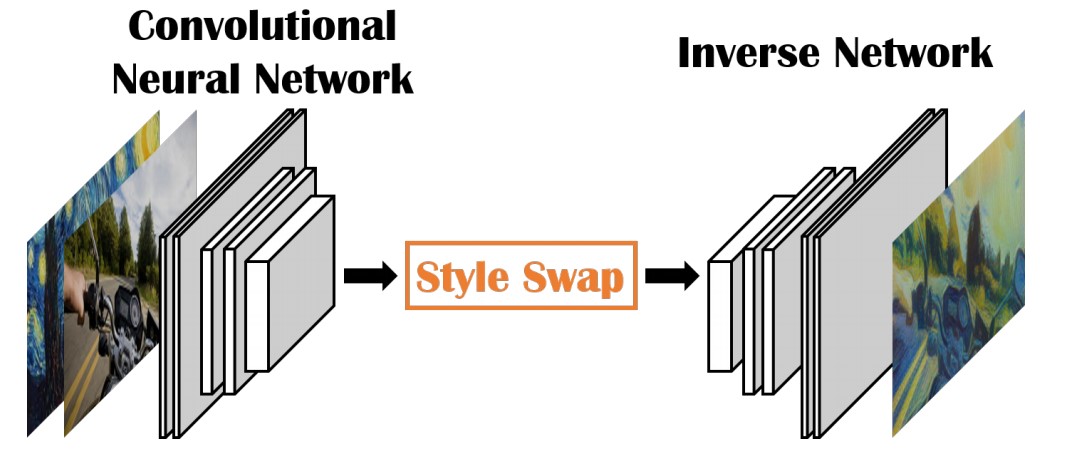}
    \caption{To the best of our knowledge both the first non-parametric neural style transfer algorithm, and the first universal feed-forward style transfer method were proposed by Chen and Schmidt in \cite{chen2016fast}. The 'Style Swap' operation refers to replacing each patch of feature activations extracted from the content with the nearest neighbor patch of activations extracted from the style. The final output can then be found by either by learning a decoder from features to pixels (pictured), or optimizing the output image directly to produce the 'style swapped' features. Figure taken from \cite{chen2016fast}.}
    \vspace{-0.5cm}
    \label{fig:steer_pyr}
\end{wrapfigure}
Chen et al. \cite{chen2016fast} use a very similar style loss to \cite{li2016combining}, but show that the content loss of \cite{gatys2016image} can be omitted. Gu et al. \cite{gu2018arbitrary} add a soft constraint on the matches found by nearest neighbors encouraging each style vector to be used at most most $k$ times. Liao et al. ~\cite{liao2017visual} propose an algorithm for finding feature correspondences between the content and style images that begins by computing matches using neural features from a deep layer of the pretrained network, then spatially refines these matches by examining successively earlier layers until correspondences can be mapped to individual pixels. Their algorithm produces excellent results on style-content pairs with similar semantics and pose, but generally does not work for more disparate style-content pairs. Mechrez et al. ~\cite{mechrez2018contextual} propose a modified nearest neighbors based loss, which increases the importance of matching content features with unique matches in the style image (e.g. matching a nose to a nose) and down-weights matches with many roughly equivalent neighbors (e.g. homogenous texture to homogenous texture). our proposed algorithm 'Neural Neighbor Style Transfer' (described in Chapter \ref{chpt:nnst}) falls within this regime and shares many similarities with \cite{chen2016fast}, but demonstrates how seemingly small design changes can lead to dramatically higher visual quality. 

While they differ in their details, many optimization-based style transfer algorithms follow the same basic framework (directly optimizing the output image to satisfy and style and/or content loss). This allows them to be described, compared, and contrasted using a fairly small set of attributes. We enumerate these attributes in Table \ref{tab:taxon}, and use them to provide a unified taxonomy of several popular methods.

\section{Learning-Based Neural Style Transfer}
Optimization based style transfer has two fundamental flaws. First, because each stylization is solved from scratch as a novel optimization problem it is impossible to learn from previously seen content and style images improve the visual quality of the current output. Second, optimization based methods are quite slow, requiring between twenty seconds and several minutes to produce a single stylization (depending on the algorithm and resolution of input images). Thus far it has been possible to address both these flaws by training models that are specialized for a particular style (or a small set of similar styles), however these methods suffer from the need to be retrained for each new style. Universal feed-forward methods, neural networks trained to produce stylizations for arbitrary input pairs, have been a very active area of research; however, thus far the visual quality of their outputs has lagged behind state-of-the-art optimization based methods.

\subsection{Feed-Forward Methods with Style Specialized Models}
Early attempts to train feed-forward models for style transfer focused on training a single model per style image or per artist/movement. Ulyanov et al. \cite{Ulyanov16} and Johnson et al. \cite{Johnson16} both proposed training models which take a single input (the content) and produce stylizations in one fixed target style. The training loss for these models was simply the original content and style loss proposed by \cite{gatys2016image}. In a follow up work Ulyanov et al. \cite{ulyanov2017improved} proposed a mechanism for enabling style transfer algorithms to produce diverse outputs by using an extra noise vector as input, and adding an additional diversity term to the training objective to encourage outputs with different noise inputs (but the same content input) to be distant in pixel space. Wang et al. \cite{wang2017multimodal} proposed an architecture that produces stylizations at multiple resolutions, and applies a separate style loss at each resolution to encourage capturing both large scale stylistic features and fine details such as brush strokes. Another work by Ulyanov et al. \cite{ulyanov2016instance} proposed Instance Normalization, normalizing an internal representation of the neural network for a particular example by subtracting by the mean and dividing the standard deviation of each channel computed over spatial locations. The purpose of this operation was to provide a closed form mechanism to put neural representations in a canonical 'style-free' feature space, which was easier to modify than the original neural representations which could have arbitrary mean and variance.

Instance normalization proved quite influential. Normalizing the representation extracted from the content allowed training networks which could produce outputs in a multiple predefined styles (typically 30-50). Chen et al. \cite{Chen17} proposed training an architecture where the parameters of most layers were shared, but a few were learned separately for each style, and were swapped in and out depending on the target style. Dumoulin et al. \cite{Dumoulin16} had a similar, but simpler approach, learning a per-style scaling and bias to apply after instance normalization. 

\begin{figure}[htp]
    \centering
    \includegraphics[width=\linewidth]{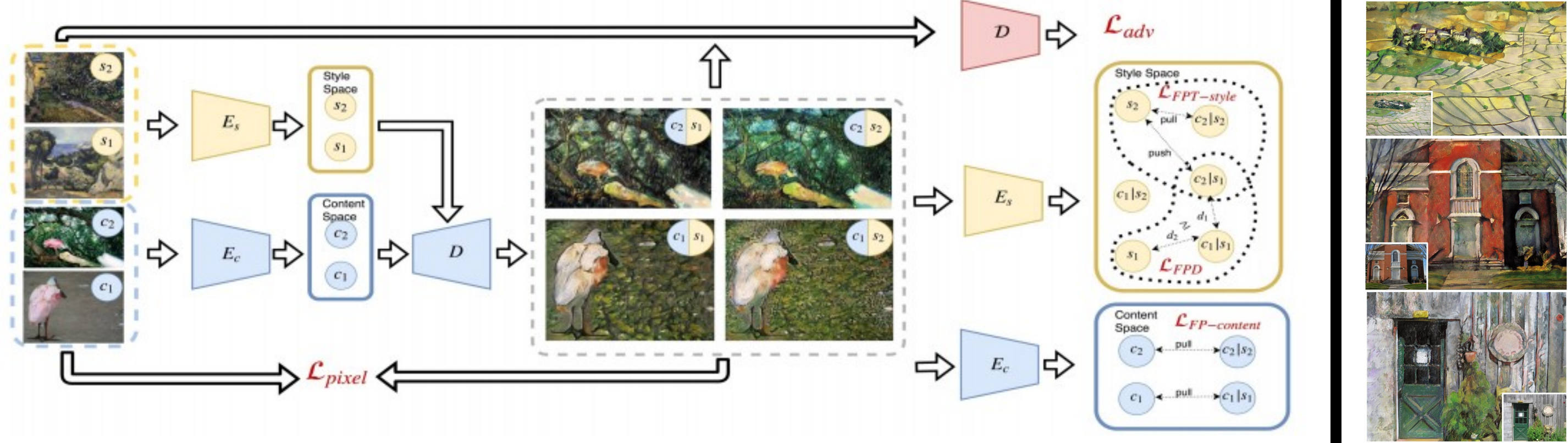}
    \caption{Some of the most impressive stylization results can be achieved by training a stylization network specialized for a particular artist (or a small number of artists). By employing an adversarial loss and a contrastive loss (left) Kotovenko et al \cite{kotovenko2019content} are able to train an efficient stylization model that produce high quality results for a few impressionist artists such as Cezanne (right). This approach works well if only a handful of styles will be used at inference and there are enough ($\approx$ 100) examples from the desired styles, but does not generalize to arbitrary styles without retraining and enough examples. Figures taken from \cite{kotovenko2019content}.}
    \vspace{-0.5cm}
    \label{fig:kotovenko}
\end{figure}

While most subsequent works in feed-forward style transfer focused on the 'universal' setting, where the style can be arbitrary, there have been several papers which consider more sophisticated objective functions for training style transfer models specialized to a particular style. Sanakoyeu et al. \cite{sanakoyeu2018style} departed from the framework of \cite{gatys2016image} in many ways. First, it replaced the gram matrix loss of Gatys with an adversarial loss based on a learned discriminator\cite{Goodfellow14}. Second, rather than relying on a pretrained representaion, this work learned representations amenable to a particular type of stylization from scratch. Third, it attempted to learn a style-invariant content representation by leveraging a cycle loss (recovering the input photo from the stylized result). Kotovenko et al. \cite{kotovenko2019content} refines these ideas, using a contrastive loss on representations extracted from the stylized output and original content to learn style-invariant representations of content, and a contrastive loss on the stylized output and the original style to learn a content-invariant representations of style. While these models must be retrained for new styles, they produce beautiful results within the chosen style.

\subsection{Universal Feed-Forward Methods}
\begin{wrapfigure}{l}{0.48\linewidth}
    \centering
    \includegraphics[width=\linewidth]{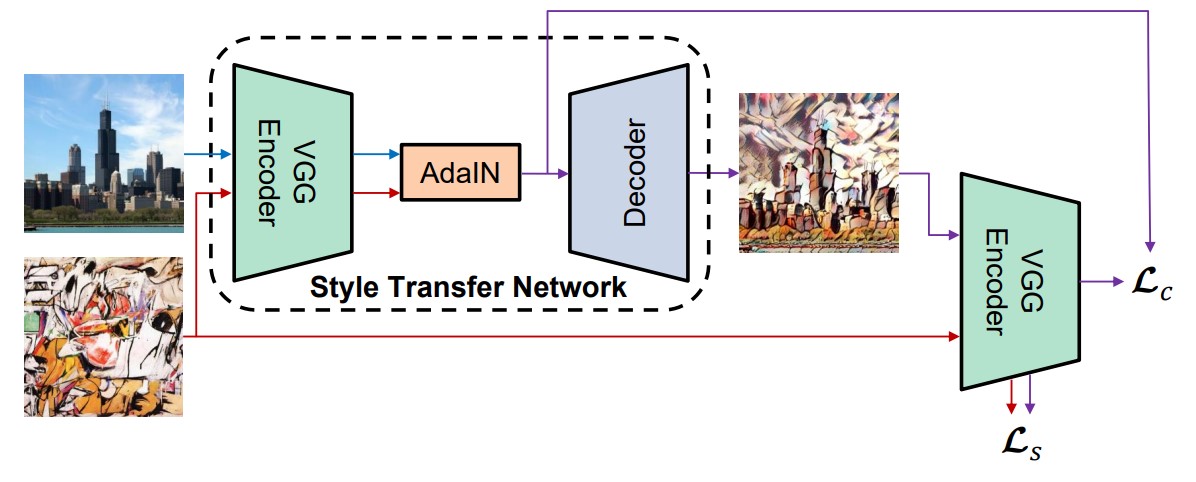}
    \caption{Overview of stylization framework using adaptive instance normalization (AdaIn) \cite{Huang17}, a parametric modification of features at the bottleneck layer based on statistics of the content and sytle features (details in text). Many subsequent works would use the same framework, but experiment with different parametric feature modifications at the bottleneck layer \cite{Li17, lu2019closed} and even combining parametric transformations with non-parametric feature replacement \cite{sheng2018avatar}. Figure taken from \cite{Huang17}.} 
    \vspace{-0.5cm}
    \label{fig:adain}
\end{wrapfigure}
To the best of knowledge the first universal feed-forward style transfer method was proposed by Chen et al. \cite{chen2016fast}. After replacing each content feature with it's nearest neighbor style feature a decoder can be trained to reconstruct pixels from the resulting representation (As an alternative to the optimization based algorithm also proposed in \cite{chen2016fast}). In contrast to this non-parametric approach Huang et al. \cite{Huang17} proposed extending the parametric approach of \cite{Dumoulin16}, replacing a bias and scaling term learned for each style with a bias and scaling terms computed on the fly based on the statistics of features extracted from the style image. This technique, called Adaptive Instance Normalization (AdaIN), was extremely influential and has been applied to many image synthesis tasks \cite{huang2018multimodal, park2019semantic, karras2019style}. To better capture the second order statistics of the style features  Li et al. \cite{Li17} propose replacing the scaling vector of \cite{Huang17} with a 'whitening' matrix derived by applying zero-phase component analysis (ZCA) to the covariance matrix of content features and a 'coloring' matrix similarly derived from the covariance of style features. This pair of affine maps has been termed the whitening and coloring transfrom (WCT). Chiu \cite{chiu2019understanding} theoretically analyzed the WCT and showed that in comparison to transforms derived from PCA and the Cholesky decomposition of the covariance matrix, using the WCT (derived from ZCA) will create features with a lower upper bound on the content loss of Gatys. Lu et al. \cite{lu2019closed} went a step further and derives a modified affine transform inspired by optimal transport that matches second order statistics while minimizing distance from the original content features. In a similar spirit Chiu et al. \cite{Chiu2020Iterative} proposes using gradient descent directly on the neural content features to trade off between the content loss and style loss, while this is more expensive than the closed form solution proposed by \cite{lu2019closed} it allows non-linear modifications to the features. An et al. \cite{an2021artflow} demonstrates that the feature transforms proposed by (WCT, AdaIn) can be applied to the representations extracted by 'flow models'  (invertible neural networks that learn to map a data distribution to gaussian which can easily be sample from). This solves the problems of style transfer being unstable (changing after multiple rounds of stylization) and the uncertainty of whether a trained decoder is perfect (because the flow model is inverible).

To combine the ideas of parametric and non-parametric style transfer Sheng et al. \cite{sheng2018avatar} propose extending the WCT by performing nearest neighbor replacement of (whitened) content features with (whitened) style features in between the whitening and coloring steps. Zhang et al. \cite{zhang2019multimodal} proposes a different intermediate between non-parametric methods based on nearest neighbors and parametric methods assuming the distribution of style features is gaussian \cite{Gatys16, huang2017arbitrary}. They propose modelling the distribution of style features as a mixture of gaussians, then using a graph cut to find the optimal matching between feature clusters. This allows a separate transformation to be applied to each cluster of content features.

Rather than derive closed form modifications to bottleneck features, several works propose to instead predict feature transformations using a trained CNN. Li et al. \cite{li2018learning} proposed learning to predict an affine transformation and bias, resulting in a feature transform with the same form as the (WCT), but which can be computed far more efficiently (since it is generated via a feed-forward pass, rather using an expensive eigendecomposition). Jing et al. \cite{jing2020dynamic} proposed instead directly predicting the parameters of a style-driven convolution and bias layer. This formulation allowed them to take advantage of other advances in architecture such as deformable convolutions \cite{dai2017deformable}.

While many of the feature transformation mechanisms described so far have been linear (i.e. parameterized as an affine map and bias term) there have been several recent efforts to learn quadratic feature modifications using a self attention module. This is the primary contribution of \cite{park2019arbitrary} and \cite{yao2019attention} which were published simultaneously. Svoboda et al. \cite{svoboda2020two} combine this module with the contrastive representation learning of \cite{kotovenko2019content}.

Recently progress has also been made scaling style transfer methods to higher resolution inputs and outputs. While most style transfer algorithms can at best produce results at 1k resolution, the typical standard for high quality art prints is 6k resolution (6000 pixels on the long side). Wang et al. \cite{wang2020collaborative} approach this through model distillation, proposing a two stage procedure where first the small encoder is distilled with a frozen full size decoder, then the small decoder is learned to accommodate the frozen distilled encoder. Texler et al. \cite{Texler20-CAG} take a different approach, proposing to combine initialization produced by neural style transfer with efficient path-based super-resolution to generate outputs of extremely high resolution (this approach can be combined with all of the algorithms presented in this work, and we provide an example of combining it with NNST in Section \ref{sec:ext_nnst}).

\section{Evaluation}
Evaluation in style transfer is a difficult problem, and an open question without widely accepted standards. Given that rigorously measuring style and content is a central challenge of the field, employing automatic metrics is problematic. When this is done it typically involves measuring an algorithm's output's score under the objective function (which measures style loss and/or content loss between the output image and style or content image respectively). Since this is the objective the output was chosen to minimize, this doesn't necessarily measure stylization quality so much as the behaviour of the optimization procedure. An alternate, potentially more meaningful, automatic metric proposed by \cite{sanakoyeu2018style} is to train a classification model on different artists' work, and evaluate style transfer outputs based on whether they are classified in the same way as the target style. 

The most common form of evaluation \cite{gatys2016image, risser2017stable, mechrez2018contextual, li2016combining, Ulyanov16, Johnson16, ulyanov2017improved, wang2017multimodal, ulyanov2016instance, Chen17, Dumoulin16, Aberman2018, chiu2019understanding, Chiu2020Iterative, lu2019closed} is simply showing qualitative comparisons between the proposed algorithm and prior work on several input pairs (typically 5-30 within a paper). Some work has proposed more rigorous evaluation protocols based on user studies that report human's perceptual judgement. One popular way to frame these studies is as measuring output's overall aesthetic quality: In \cite{jing2019neural} study participants are asked to give an absolute score for output's aesthetic quality, in \cite{gu2018arbitrary} participants are shown the outputs of all benchmarked algorithms simultaneously and asked to rank them based on aesthetic quality, and in \cite{Li17,an2021artflow} participants were similarly shown all outputs but instead asked to choose a single best result. While most studies use participants drawn from the general population of mechanical turk workers, some studies have leveraged the help of experts (holders of advanced degrees in art history) in evaluating style transfer. Sanakoyeu et al. \cite{sanakoyeu2018style} asked experts to choose which algorithms' output best match the style of the input image. Kotovenko et al. \cite{kotovenko2019content} conducted a similar study, but with styles from well known artists. Experts were shown zoomed in crops of output images, and asked which crop best matched the style of the desired artist (but where not shown the style input), they also proposed a more challenging variant where a crop from a real style input was an additional option that could be selected as the 'best'.

While these efforts are helpful in evaluating the quality of style transfer algorithms, they do not separately evaluate content preservation and stylization quality. Many, potentially all, of the evaluation procedures described would give high scores to an algorithm that simply returned the style input image. Motivated by this chapters \ref{chpt:STROTSS} and \ref{chpt:dst} describe forced choice user studies which evaluate both stylization quality and content preservation.

\chapter{Style Transfer via Relaxed Optimal Transport and Self-Similarity (STROTSS)}
\label{chpt:STROTSS}
\begin{figure}[h]
\centering
\includegraphics[width=\linewidth]{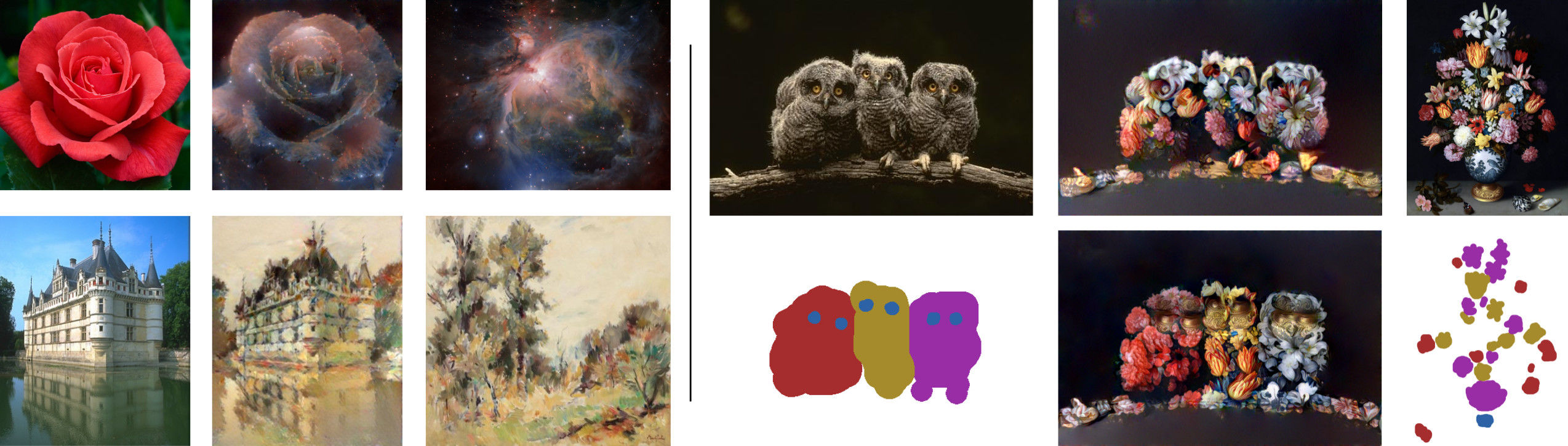}
\caption{Examples of style transfer outputs produced by STROTSS (middle of each triplet). The left triplets show style transfer results without guidance. The right triplets show outputs for the same content and style images without guidance (top) and with guidance (bottom). In this guidance consists of user's annotations of which regions of the style should stylize which regions of the content (indicated by regions of the same color in the segmentation maps of the style and content).}
\end{figure}

As previously discussed, one of the main challenges of style transfer is formalizing
'content' and 'style', terms which evoke strong intuitions but 
are hard to even define semantically. This work focused on formulations of
each term which were, at the time of publication, novel in the domain of style transfer, but had long history of successful application elsewhere in computer vision. The paper this chapter is based on appeared in CVPR 2019 and was joint work with Jason Salavon, and my advisor Greg Shakhnarovich.

\newcommand{\xfig}[1]{\includegraphics[height=0.09\linewidth]{chapters/STROTSS/Figures/Qual_Cross/#1}}
\newcommand{\xfigLine}[1]{\xfig{#1} & \xfig{#1_01} & \xfig{#1_05} & \xfig{#1_03} & \xfig{#1_04} & \xfig{#1_02}}
\newcommand{\xfigLineS}[1]{ & \xfig{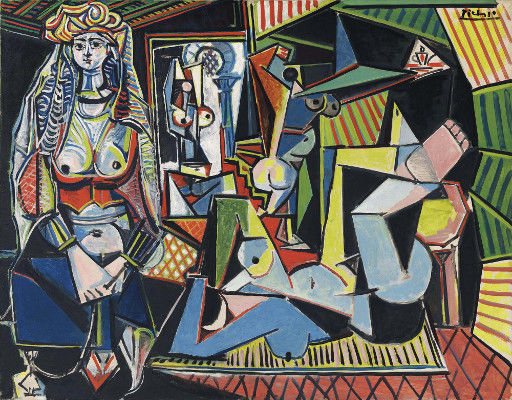} & \xfig{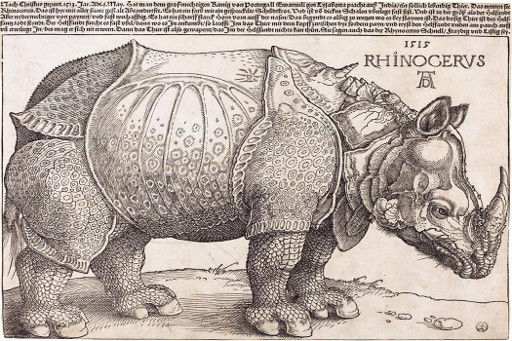} & \xfig{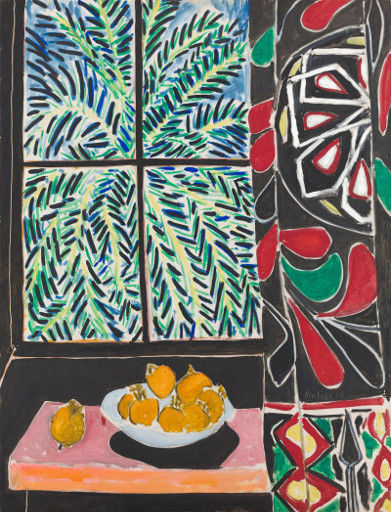} & \xfig{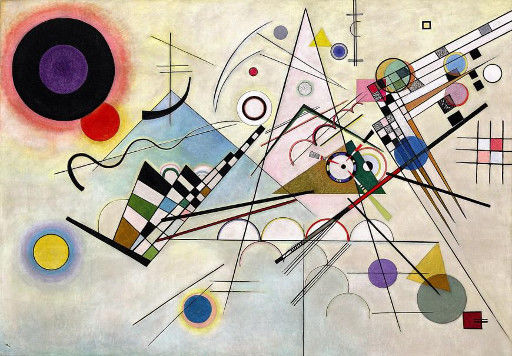} & \xfig{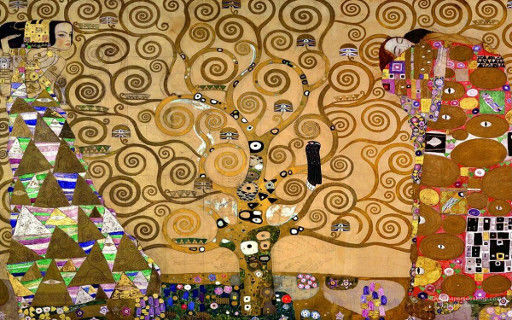}}

\begin{figure}[H]
	\centering
	\begin{tabular}{cccccc}
		\xfigLineS{01}\\
		\xfigLine{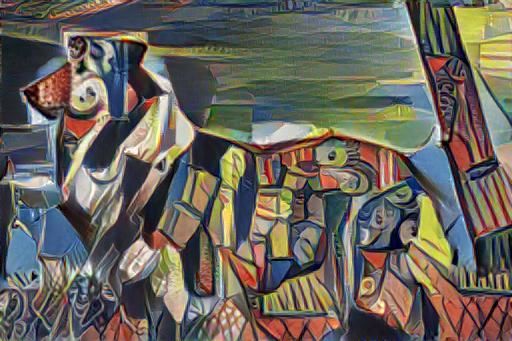}\\
		\xfigLine{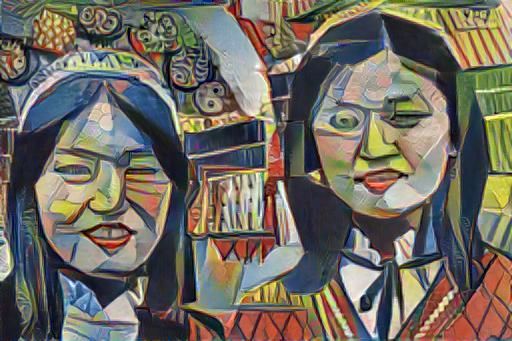}\\
		\xfigLine{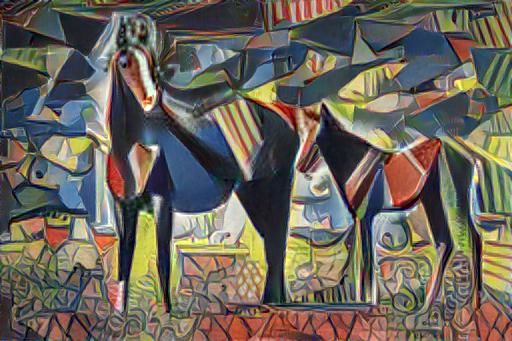}\\
		\xfigLine{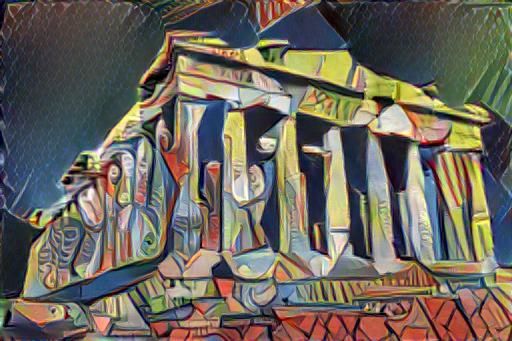}\\
		\xfigLine{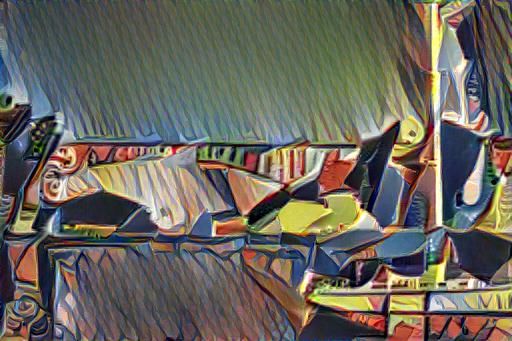}\\
		 & Picasso & D{\"u}rer & Matisse & Kandinsky & Klimt\\
	\end{tabular}
	\caption{Outputs produced by STROTSS which demonstrate the effect of different content images on the same style, and vice-versa}
	\label{fig:qual_cross}
\end{figure}

\section{Introduction}\label{sec:strotss_intro}
Prior neural style transfer algorithms tended to either treat style as multivariate Gaussian distribution over neural features, or give up a probabilistic interpretation of style by turning to non-parametric feature matching (using nearest neighbors) to better capture details of the style. Defining style based on a limited set of summary statistics is an impoverished representation which cannot capture complex stylistic details. On the other hand, while replacing each content feature with it's nearest style neighbor can copy details of the style, not considering a distribution of features can lead to an overly homogeneous result.
Our goal was a formulation of style similarity that captured the best of both worlds, the detail preservation of non-parametric approaches and the diversity of probabilistic approaches.
We defined style as an empirical distribution over features extracted by
a deep neural network, and measured the similarity between such
distributions using an efficient approximation of the Earth Movers
Distance initially proposed in the Natural Language Processing
community~\cite{kusner2015word}. The Earth Movers Distance is not only a useful measure of similarity between distributions which encourages using all the features in the style image, it is also non-parametric and therefore not constrained to matching the limited summary statistics of a Gaussian.

Our definition of content similarity (or content loss) was motivated by the phenomena of pareidolia, which is when humans perceive arrangements of arbitrary shapes and textures to represent a semantic object (for example seeing animals made of clouds, or a face in a car's grille). This phenomena implies that our perception is at least partially based on relative appearance, rather than absolute appearance, which allows us to recognize a shape as representing an object regardless of the texture. Self-similarity \cite{shechtman2007matching} is a formalizes this idea to define a local feature descriptor, where a feature vector at a particular locations
is transformed into a vector of distances from other feature vectors at fixed spatial offsets. In this work we consider self-similarity to be a global phenomena, and a feature vector at a particular locations
is transformed into a vector of distances from \textbf{all} other feature vectors. Content similarity is then the $L_1$ distance between two such tensors of global self-similarity features.
Defining content similarity in this way helped disconnect it from 
pixels' precise values. Concretely it makes the distance between two content representations invariant to translations and orthonormal affine transformations in feature space (when applied across all spatial positions). This additional invariance makes it easier to satisfy than definition of content similarity used in prior work, distance between raw features, and results in outputs which maintain the perceived semantics and spatial layout of the content image despite being highly stylized.

We developed a style transfer algorithm using these definitions of content similarity and style similarity called "Style Transfer via Relaxed Optimal Transport and Self-Similarity" or STROTSS.

Through a large user study we demonstrated that for any desired level of content 
preservation, STROTSS provided higher quality stylization than prior work. The structure of this user study itself was a contribution of this paper, attempting to provide a more nuanced evaluation of style transfer than prior work. We asked users to evaluate both 'content preservation' and 'stylization quality', because evaluating only one of these questions leads to a degenerate 'optimal algorithm' which simply returns the content or style image.

To increase utility of STROTSS as an artistic tool, it was
important that users could easily and intuitively control the
algorithm's output. We extended our formulation to allow
region-to-region constraints on style transfer (e.g., ensuring that
hair in the content image is stylized using clouds in the style image)
and point-to-point constraints (e.g., ensuring that the eye in the
content image is stylized in the same way as the eye in a
painting).

\section{Background: A Neural Algorithm of Artistic Style}\label{sec:strotss_gatys}
\subsection{Feature Extraction}
Optimization-Based Neural Style Transfer algorithms rely on the rich
feature representations extracted by a pre-trained neural networks.
All works in this regime which I am aware of use VGG pre-trained
for ImageNet classification ~\cite{simonyan2014very}. I will define
$\Phi(X)^l$ be the tensor of feature activations extracted from input
image $X$ by layer $l$ of network $\Phi$. 

\subsection{Objective Function}
The style loss of Gatys et al. \cite{gatys2016image} had previously been proposed by the same authors for the purpose of texture synthesis \cite{gatys2015texture}, and that work was inspired by the parametric matching of hand-crafted features (for texture synthesis) proposed by Portilla and Simoncelli \cite{Portilla00}. The loss is based on Gram matrices computed over the spatial dimensions of $\Phi(X)^l$, the tensor of neural activations in layer $l$. If $\Phi(X)^l$ has $F$ feature channels then the gram matrix at that layer $\mathcal{G}^l$ is a positive semi-definite matrix $\mathcal{G}^l\in \mathbb{R}^{F\times F}$.  Each entry of such a matrix is defined as:
\begin{equation}
    \mathcal{G}^l_{i,j} = \sum_k \Phi(X)^l_{i,k} \cdot \Phi(X)^l_{j,k} 
\end{equation}
Where $i/j$ index over channels of $\Phi(X)^l$ and $k$ indexes over spatial coordinates of $\Phi(X)^l$. The gram matrix can be seen as related to the covariance of features (it differs in being uncentered and unnormalized), and captures the co-occurence of features at the same spatial location.

Given $\mathcal{G}^l_O$ and $\mathcal{G}^l_S$ (computed respectively from the output image $O$ and style image $S$), the style loss at layer $l$ can is defined defined as the froebenius norm of the difference between these matrices:
\begin{equation}
    \mathcal{L}_s^l(O, S) = \|\mathcal{G}^l_O - \mathcal{G}^l_S \|_2^2 
\end{equation}

Given a weighting $w_l$ for each layer (for example weighting an early layer highly would prioritize matching features based on color, while weighting a slightly deeper layer might prioritize matching features based on texture), the total style loss can be defined as: 

\begin{equation}
    \mathcal{L}_s(O, S) = \sum_l w_l \cdot \mathcal{L}_s^l(O, S)
\end{equation}

In practice the algorithm uniform weights layers 'conv1\_1', 'conv2\_1', 'conv3\_1', 'conv4\_1',  and 'conv5\_1' from VGG ~\cite{simonyan2014very} pre-trained for imagenet classification. Notably this style loss can only capture the magnitude of features present, and their second order interactions.

The given current output O and content C, the content loss for layer $l$ is simply defined as: 
\begin{equation}
    \mathcal{L}_c(O,C) = \|\Phi(O)^l - \Phi(C)^l \|^2 
\end{equation}

In practice the algorithm only uses one VGG layers, 'conv4\_2', for this loss.

The optimization problem for finding the output image $O$ can then be expressed as the differentiable (w.r.t. $O$) objective function:
\begin{equation}
    \min_O \alpha \mathcal{L}_c(O,C) + \mathcal{L}_S(O,S) \label{eq:gatys_obj}
\end{equation}
This objective function can be minimized by directly updating the output image using gradient descent or a variant (the original implementation of \cite{gatys2016image} uses L-BFGS-B \cite{zhu1997algorithm}).

where $\alpha$ controls the tradeoff between stylization and content preservation.

We are now ready to introduce STROTSS, which follows the general form of Equation \ref{eq:gatys_obj}, but uses very different definitions of content and style loss.

\section{STROTSS}\label{sec:meth}
Like the original Neural Style Transfer algorithm proposed by Gatys et al.~\cite{gatys2016image} our method takes two inputs, a style image $S$ and a content image $C$, and uses the gradient descent variant RMSprop \cite{hinton2012neural} to minimize our proposed objective function (equation \ref{eq:full_obj}) with respect to the output image $X$.
\begin{align}
L(O,C,S)= \frac{(\alpha \cdot \ell_C) + \ell_m + \ell_r + (\frac{1}{\alpha} \cdot \ell_p)}{2+\alpha+\frac{1}{\alpha}}\label{eq:full_obj}
\end{align}

We describe the content term of our loss $\alpha \cdot \ell_C$ in Section \ref{sec:sl}, and the style term $\ell_m + \ell_r + (\frac{1}{\alpha}  \cdot \ell_p)$ in  Section \ref{sec:cl}. The hyper-parameter $\alpha$ controls the trade-off between content preservation to stylization (demonstrated in Figure \ref{fig:knob_fig}). Our method iteratively optimizes the output image $O$, let $O^{(t)}$ refer to the stylized output image at timestep $t$. We describe our initialization $O^{(0)}$ in Section \ref{subsec:impl}.

\newcommand{\cKIm}[3]{\includegraphics[height=#3cm]{chapters/STROTSS/Figures/content_knob/#1_#2}}

\newcommand{\contentKnobRow}[2]{\cKIm{#1}{c}{#2} & \cKIm{#1}{2000}{#2} & \cKIm{#1}{1000}{#2} & \cKIm{#1}{0500}{#2} & \cKIm{#1}{0250}{#2} & \cKIm{#1}{s}{#2}}

\begin{figure}[tb]
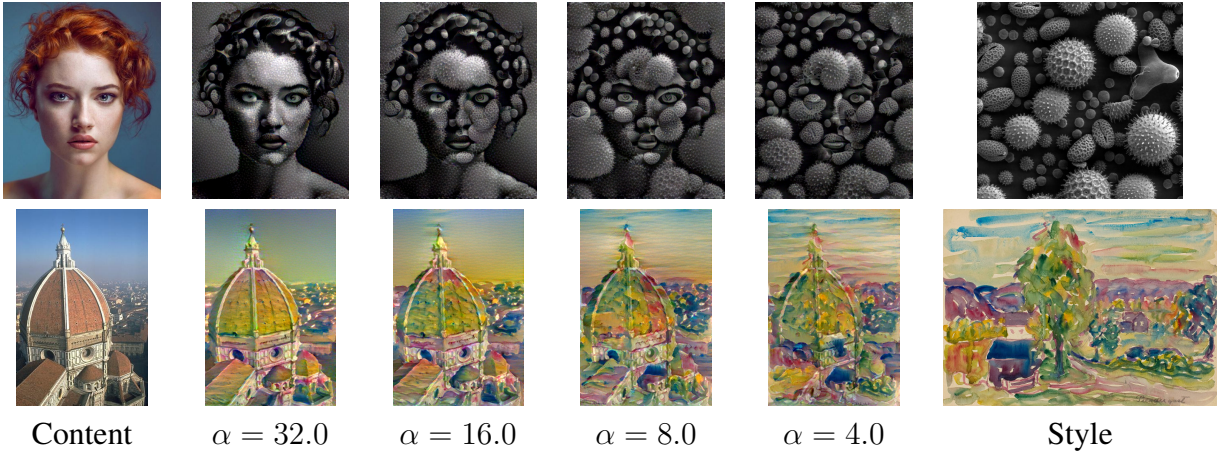

	\centering
\begin{tabular}{cccccc}
	\contentKnobRow{pollen}{2.6}\\
	\contentKnobRow{dome}{2.6}\\
	Content & $\alpha=32.0$ & $\alpha=16.0$ & $\alpha=8.0$ & $\alpha=4.0$ & Style
\end{tabular}\vspace{-.5em}
	\caption{Effect of varying $\alpha$, the content loss weight, on our unconstrained style transfer output, because we stylize at four resolutions, and halve $\alpha$ each time, our default $\alpha=16.0$ is set such that $\alpha=1.0$ at the final resolution.}
	\label{fig:knob_fig}
\end{figure}

\subsection{Feature Extraction}
Both our style and content loss terms rely upon extracting a rich
feature representation from an particular spatial location. In this
work we use hypercolumns \cite{mostajabi2015feedforward,
  hariharan2015hypercolumns} extracted from a subset of layers of VGG16 trained
on ImageNet~\cite{simonyan2014very}. Let
$\Phi(X)^i$ be the tensor of feature activations extracted from input
image $X$ by layer $i$ of network $\Phi$. Given the set of layer
indices $l_1,..,l_L$ we use bilinear upsampling to match the spatial
dimensions of $\Phi(X)^{l_1}...\Phi(X)^{l_L}$ to those of the original
image ($X$), then concatenate all such tensors along the feature dimension. This yields a hypercolumn at each pixel which contains low-level edge and color features, mid-level texture features, and high-level semantic features \cite{zeiler2014visualizing}. The content and  style losses of STROTSS used all convolutional layers of VGG16 except layers  'conv4\_2', 'conv4\_3', 'conv5\_2', and 'conv4\_3', which we excluded due to memory constraints at the time of publication.

\subsection{Style Loss}\label{sec:sl}


Optimal transport~\citep{monge1781memoire,villani2008optimal} can be framed as finding the transportation map that minimizes that cost of moving probability mass between two histograms (i.e. probability mass functions).
\begin{align}
\min_T & \hspace{1cm} \sum_{ij} T_{ij}C_{ij} \label{eq:oto} \\ 
\text{s.t.} & \hspace{1cm} T_i = \alpha_i \label{eq:constrain_i}\\ 
& \hspace{1cm} T_j = \beta_j \label{eq:constrain_j} \\ 
& \hspace{1cm} T_{ij} \geq 0 \label{eq:constrain_z}
\end{align}
$C_{ij}$ is the cost of transporting one unit of probability mass between bin $i$ of histogram $\alpha$ to bin $j$ of histogram $\beta$, $T_i$ and $T_j$ are respectively the row sums and column sums of $T$ (the transport map), and $T_{ij}$ is the amount of probability mass being transported between bin $\alpha_i$ and bin $\beta_j$. Contraints \ref{eq:constrain_i}, \ref{eq:constrain_j}, and \ref{eq:constrain_z} ensure that transport map conserves probability mass. The cost of the minimizing transport map (Equation \ref{eq:oto}) is also called the Earth Movers Distance (EMD).

Let $A=\{A_1,\ldots,A_n\}$ be a set of $n$ feature vectors extracted
from $O^{(t)}$, and $B=\{B_1,\ldots,B_m\}$ be a set of $m$ features
extracted from style image $S$. The style loss of STROTSS is derived from the EMD, but since we consider all features to have equal mass, we can use a simplified version of the EMD:
\begin{align}
\textsc{EMD}(A,B)=&\min_{T} \sum_{ij} T_{ij}C_{ij}\\
s.t. &\sum_j T_{ij} = 1/m\label{eq:emda}\\
&\sum_i T_{ij} = 1/n\label{eq:emdb}\\
& \hspace{1cm} T_{ij} \geq 0
\end{align}
In this case $C$, the 'cost matrix', contains how far each
element in $A$ is from each element in $B$. $\textsc{EMD}(A,B)$ captures
the distance between sets $A$ and $B$, but finding the optimal $T$ costs $O(\max(m,n)^3)$, and is therefore untenable for gradient descent based style
transfer (where it would need to be computed at each update
step). Instead we will use the Relaxed EMD~\cite{kusner2015word}. To
define this we will use two auxiliary distances, essentially each is
the EMD with only one of the constraints ~\eqref{eq:emda} or~\eqref{eq:emdb}:
\begin{align}
R_A(A,B)=&\min_{T\geq 0} \sum_{ij} T_{ij}C_{ij}\quad
s.t. &\sum_j T_{ij} = 1/m\\
R_B(A,B)=&\min_{T\geq 0} \sum_{ij} T_{ij}C_{ij}\quad
s.t. &\sum_i T_{ij} = 1/n
\end{align}
we can then define the relaxed earth movers distance as:
\begin{align}
\ell_r = REMD(A,B)=&\max (R_A(A,B),R_B(A,B))
\end{align}

This is equivalent to:
\begin{align}
\ell_r = &\max \left(\frac{1}{n}\sum_i \min_j C_{ij},\frac{1}{m}\sum_j \min_i C_{ij}\right)
\end{align}
The efficiency of computing this is dominated by the efficiency of computing the cost matrix $C$. We define matrix entry $C_{ij}$, the cost of transport probability mass from from $A_i$ to $B_j$ (also called the ground metric) as the
cosince distance between the two feature vectors:
\begin{align}
C_{ij}\,=\,D_{\cos}(A_i,B_j)\,=\,1-\frac{A_i \cdot B_j}{ \|A_i\| \|B_j\|}
\end{align}

While $\ell_r$ does a good job of transferring many features of the style image to the content, the cosine distance ignores the magnitude of the feature vectors. In practice this leads to visual artifacts in the output, most notably over/under-saturation. To combat this we add a moment matching loss:
\begin{equation}
\ell_m\,=\,\frac{1}{d}\|\mu_A-\mu_B\|_1 + \frac{1}{d^2}\|\Sigma_A-\Sigma_B\|_1
\end{equation}
where $\mu_A$, $\Sigma_A$ are the mean and covariance of the feature vectors in set $A$, and $\mu_B$ and $\Sigma_B$ are defined analagously.

We also add a color matching loss, $\ell_p$ to encourage our output and the style image to have a similar palette.
$\ell_p$ is defined using the Relaxed EMD between pixel colors in
$X^{(t)}$ and $I_S$, this time and using Euclidean distance as a ground
metric. We find it beneficial to convert the colors from RGB into a decorrelated colorspace with mean color as one channel when computing this term. Because palette shifting is at odds with content preservation, we weight this term by $\frac{1}{\alpha}$. 

\begin{figure}[H]
	\centering
	\includegraphics[width=0.95\textwidth]{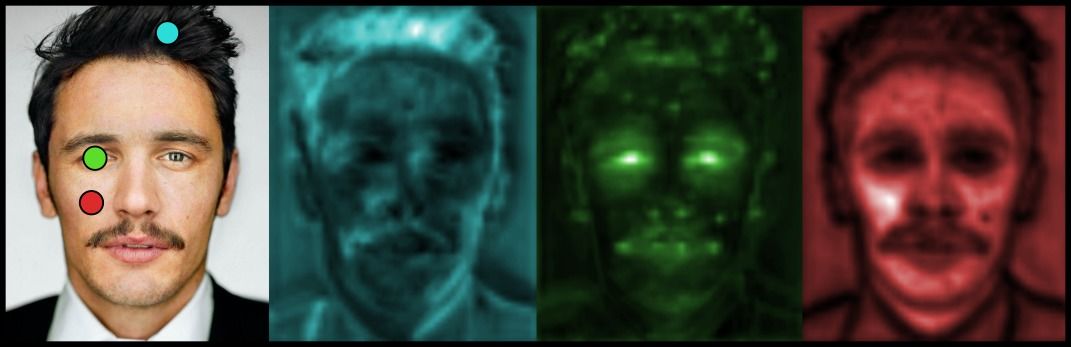}
	\caption{The blue, red, and green heatmaps visualize the cosine similarity in feature space relative to the corresponding points marked in the photograph. Our content loss attempts to maintain the relative pairwise similarities between 1024 randomly chosen locations in the content image}
	\label{fig:conSim}
\end{figure}

\begin{figure}[H]
	\centering
    \begin{tabular}{ccc}
	\includegraphics[height=0.23\textwidth]{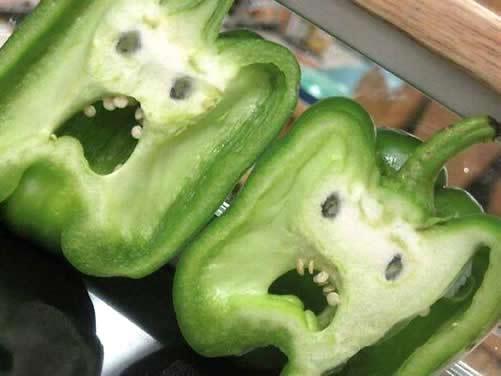} &
	\includegraphics[height=0.23\textwidth]{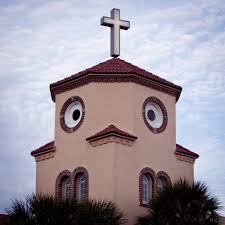} &
	\includegraphics[height=0.23\textwidth]{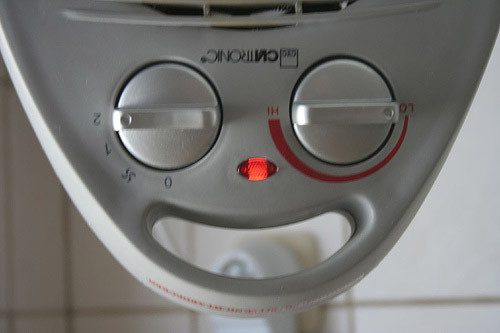}
    \end{tabular}
	\caption{Examples of the perceptual phenomena of pareidolia, the human tendency to incorrectly assign semantic meaning to visual stimuli based on self-similarity}
	\label{fig:pareid}
\end{figure}

\subsection{Content Loss}\label{sec:cl}
Our content loss is motivated by the observation that robust pattern
recognition can be built using local self-similarity descriptors
\cite{shechtman2007matching}. An every day example of this is the
phenomenon called pareidolia, where the self-similarity patterns of
inanimate objects are perceived as faces because they match a loose
template (see Figure \ref{fig:pareid}). Formally, let $D^O$ be the pairwise cosine
distance matrix of all (hypercolumn) feature vectors extracted from $O^{(t)}$, and
let $D^{C}$ be defined analogously for the content image. We visualize several potential rows of $D^C$ in Figure \ref{fig:conSim}. We define our content loss as:
\begin{align}
\mathcal{L}_{content}(O,C) = \frac{1}{n^2} \sum_{i,j} \left|  \frac{D^O_{ij}}{\sum_i D^O_{ij}}- \frac{D^{C}_{ij}}{\sum_i D^{C}_{ij}}\right|
\end{align}
In other words the normalized cosine distance between feature vectors
extracted from any pair of coordinates should remain constant between
the content image and the output image. This constrains the structure
of the output, but allows global translations and rotoreflections in feature space relative to the original content image. As a result semantics and spatial layout are broadly preserved, but colors and texture of our output $O^{(t)}$ can
drastically differ from those in $C$.

\begin{figure}[H]
	\centering
	\includegraphics[width=\linewidth]{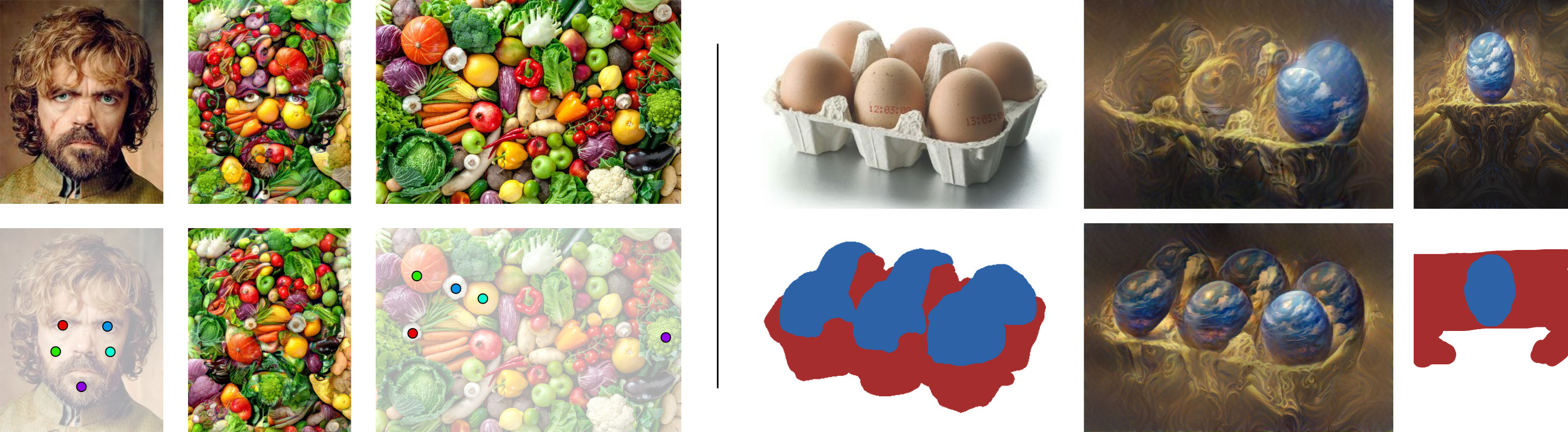} 
	\caption{Examples of using guidance for
          aesthetic effect (left, point-to-point)) and error
          correction (right, region-to-region). In the top row the
          images are arranged in order of content, output,
          style. Below each content and style image we show the
          guidance mask, and between them the guided output.}
	\label{fig:control_fig}
\end{figure}

\subsection{User Control}
\begin{figure}[ht]
\includegraphics[width=\textwidth]{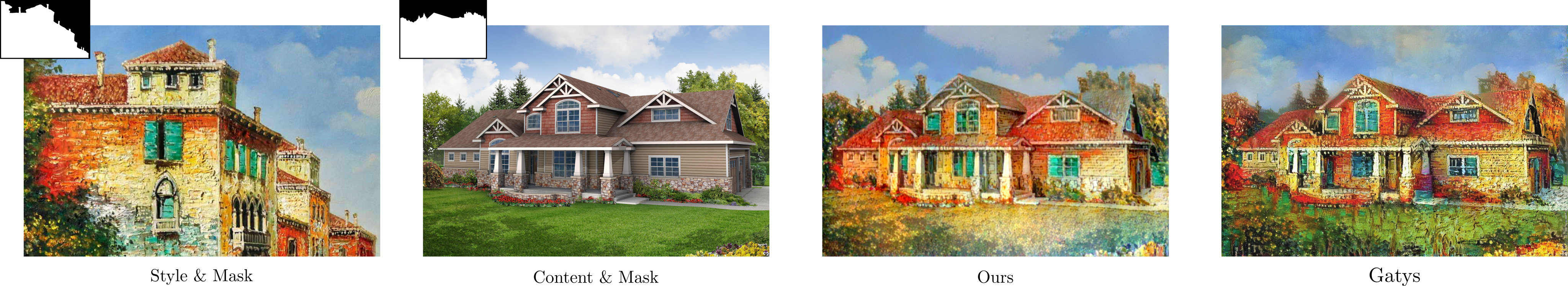}
	\caption{Qualitative comparison of the resulting output of our spatial guidance and that proposed in ~\cite{gatys2017controlling}}
	\label{fig:mturk_guid_comp_fig}
\end{figure}

We incorporate user control as constraints on the style of the output. Namely the user defines paired sets of spatial locations (regions) in $O^{(t)}$ and $S$ that must have low style loss. In the case of point-to-point user guidance each set contains only a single spatial location (defined by a click). Let us denote paired sets of spatial locations in the output and style image as $(O_{t1},S_{s1})...(O_{tK},S_{sK})$. We redefine the ground metric of the Relaxed EMD as follows:
\begin{align}
C_{ij} =
	\begin{cases}
	\beta*D_{cos}(A_i,B_j), \text{ if } i \in O_{tk}, j \in S_{sk} \\
	\infty, \text{ if } \exists k \text{ s.t. } i \in O_{tk} ,j \not \in S_{sk} \\
	D_{cos}(A_i,B_j) \text{ otherwise},
	\end{cases}
\end{align}
where $\beta$ controls the weight of user-specified constraints relative to the unconstrained portion of the style loss, we use $\beta=5$ in all experiments. In the case of point-to-point constraints we find it useful to augment the constraints specified by the user with 8 additional point-to-point constraints, these are automatically generated and centered around the original to form a uniform 9x9 grid. The horizontal and vertical distance between each point in the grid is set to be 20 pixels for 512x512 outputs, but this is is a tunable parameter that could be incorporated into a user interface. For examples of region-to-region and point-to-point user control, see Figure ~\ref{fig:control_fig}. Using one of the examples from ~\cite{gatys2017controlling}, which extended the algorithm of \cite{gatys2016image} to allow region-to-region control, we demonstrate in Figure \ref{fig:mturk_guid_comp_fig} that our proposed control mechanism performs similarly.

\begin{figure}[H]
\includegraphics[width=\textwidth]{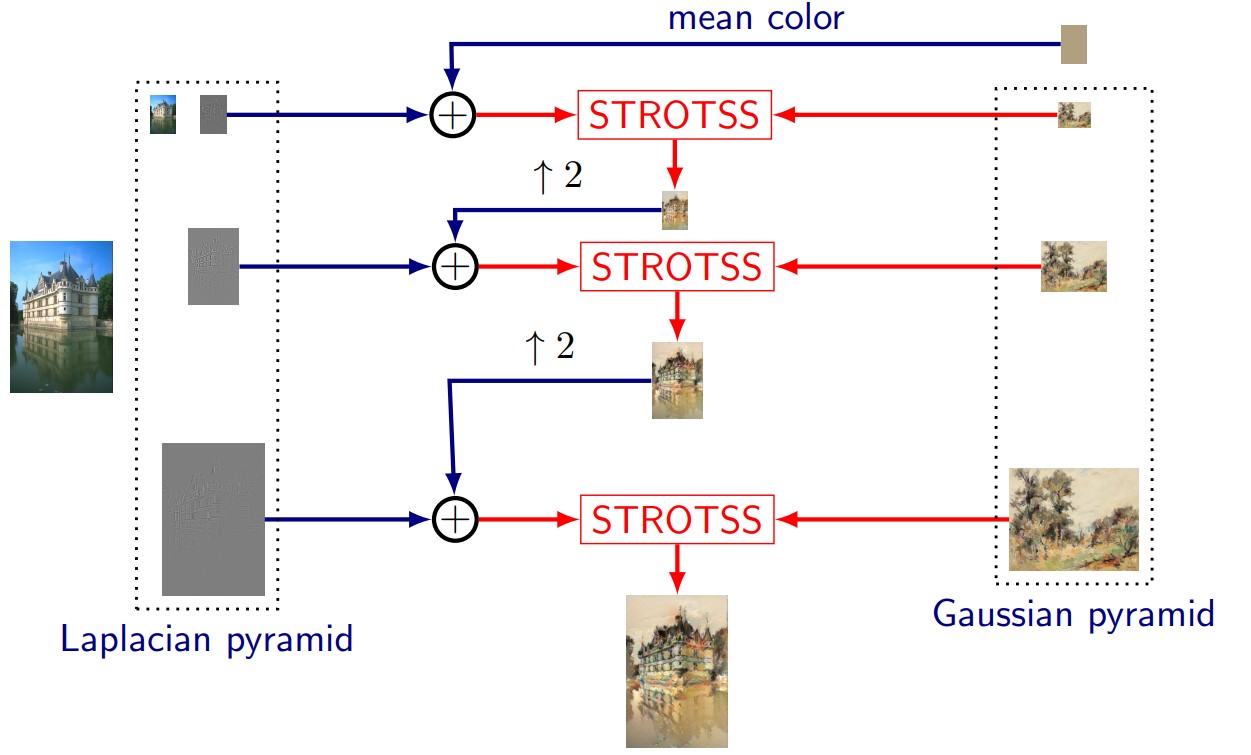}
	\caption{Overview of multi-scale STROTSS. At the lowest resolution we initialize using the mean color of the style image plus the high frequencies of the content image at that resolution. After using STROTSS to stylize a given scale we initialize stylization at the next scale by upsampling the result and adding the high frequencies of the content image at the new scale.}
	\label{fig:strotss_overview}
\end{figure}

\subsection{Implementation Details}\label{subsec:impl}
We apply our method iteratively at increasing resolutions, halving $\alpha$ each time. We begin with the content and style image scaled to have a long side of 64 pixels. The output at each scale is bilinearly upsampled to twice the resolution and
used as initialization for the next scale. By default we stylize at four resolutions, and because we halve $\alpha$ at each resolution our default $\alpha=16.0$ is set such that $\alpha=1.0$ at the final resolution. We use this schedule of altering $\alpha$ to prevent over-stylization at low resolutions, because low-frequencies are generally more important for content preservation and once they are destroyed they cannot be recovered by STROTSS at finer resolutions.

At the lowest resolution we initialize using the bottom level of a Laplacian pyramid constructed from the content image (high frequency gradients) added to the mean color of the style image. We then decompose the initialized output image into a five level Laplacian pyramid. We update the pyramid coefficients using RMSprop~\cite{hinton2012neural} to minimize our objective function. We find that optimizing the Laplacian pyramid, rather than pixels directly, dramatically speeds up convergence. At each scale we make 200 updates using RMSprop, and use a learning rate of 0.002 for all scales except the last, where we reduce it to 0.001. 

The pairwise distance computation required to calculate the style and content loss precludes extracting features from all coordinates of the input images, instead we sample 1024 coordinates randomly from the style image, and 1024 coordinates in a uniform grid with a random x,y offset from the content image. We only differentiate the loss w.r.t the features extracted from these locations, and resample these locations after each step of RMSprop. 
\begin{figure}[!t]
	\centering
	\includegraphics[width=0.9\linewidth, frame]{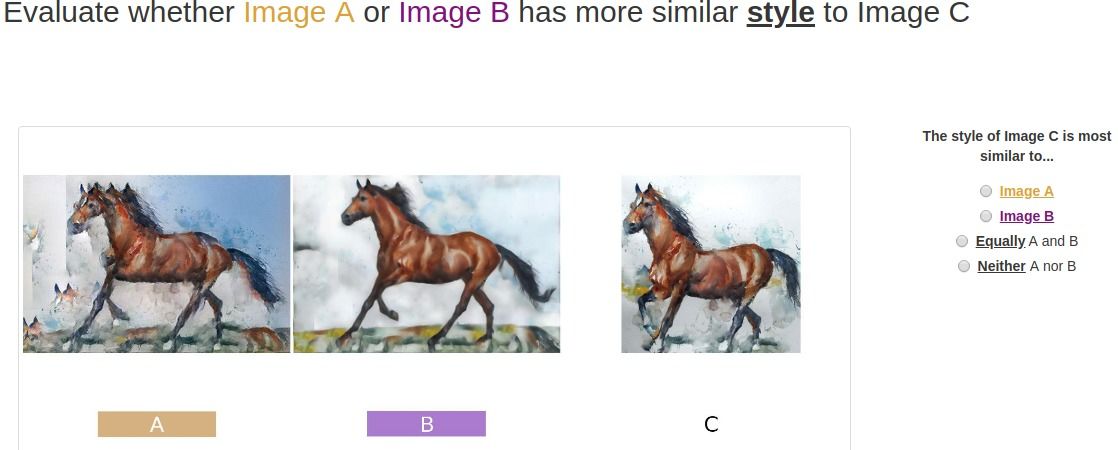}
	\caption{Example of User Study Interfaced (Details in Section \ref{sec:h_eval})}
	\label{fig:mturk_gui}
\end{figure}

\newcommand{\qCIm}[3]{\includegraphics[height=#3\textwidth]{chapters/STROTSS/Figures/Compare/#1_#2} }

\newcommand{\qCRow}[2]{\qCIm{#1}{Content}{#2} & \qCIm{#1}{Style}{#2} & \qCIm{#1}{Ours}{#2} & \qCIm{#1}{Reshuffle}{#2} & \qCIm{#1}{Gatys}{#2} & \qCIm{#1}{CNNMRF}{#2} & \qCIm{#1}{Contextual}{#2} }

\begin{figure}[!t]
	\centering
	
	\begin{tabular}{ccccccc}
	\qCRow{00}{0.13}\\
	\qCRow{01}{0.06}\\
	\qCRow{02}{0.13}\\
	\qCRow{03}{0.07}\\
	\qCRow{04}{0.13}\\
	\qCRow{05}{0.13}\\
	Content & Style & Ours & Reshuffle \cite{gu2018arbitrary} & Gatys \cite{gatys2016image} & CNNMRF \cite{li2016combining} & Contextual \cite{mechrez2018contextual} 
\end{tabular}

	\caption{Qualitative comparison between our method and prior work. Default hyper-parameters used for all methods}
	\label{fig:comp_fig}
\end{figure}

\begin{figure}
	\centering
	\includegraphics[width=\linewidth]{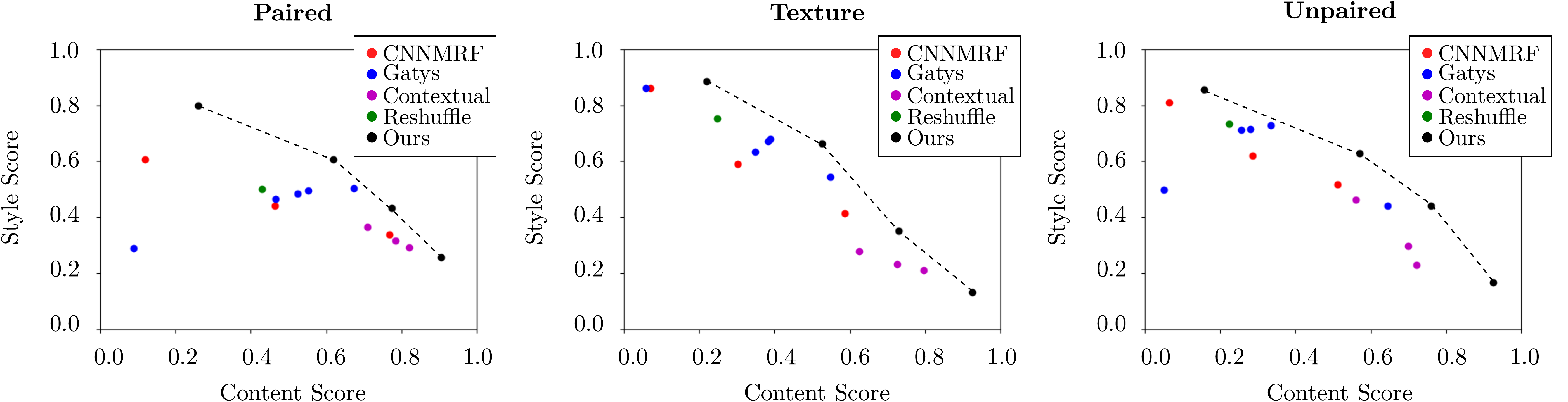}
	\caption{Quantitative evaluation of our method and prior work, we estimate the Pareto frontier of the methods evaluated by linearly interpolation (dashed line)}
	\label{fig:mturk_results_fig_strotss}
\end{figure}

\section{Evaluation}
Unlike discriminative tasks like classification and semantic segmentation, there are no widely accepted metrics by which to benchmark style transfer. Quantitative measurements of synthesized images' absolute quality, let alone whether they satisfy complex criteria like perceptual content preservation or stylistic similarity, cannot (so far) compare the judgement of humans. Therefore user studies are the gold-standard for style transfer evaluation.

\subsection{Large-Scale Human Evaluation}\label{sec:h_eval}
Because style transfer between arbitrary content and style pairs is such a broad task, we proposed an evaluation dataset, broken into three regimes, to cover major use cases of style transfer. In the 'Paired' regime the content image and style image are represent similar contents, this is mostly images of the same category (e.g. both images of dogs), but also includes images of the same entity (e.g. both images of the London skyline). In the 'Unpaired' regime the content and style image are \textbf{not} representations of the same thing (e.g. a photograph of a Central American temple, and a painting of a circus). In the 'Texture' regime the content is a photograph of a face, and the style is a homogeneous texture (e.g. a brick wall, flames). Each regime's evaluation set consisted of 30 content/style pairs (a total of 90).

To quantitatively compare STROTSS with prior work we
performed several studies using Amazon Mechanical Turk (AMT) across the three regimes. An example of the workers' interface is shown in
Figure~\ref{fig:mturk_gui}. Images A and B are the product of two different style transfer algorithms given the same inputs. We consider the algorithms proposed
in~\cite{gatys2016image},\cite{gu2018arbitrary},~\cite{li2016combining},~\cite{mechrez2018contextual},
and STROTSS. In Figure~\ref{fig:mturk_gui} image C is the corresponding
style image, and workers were asked to choose whether the style of
image is best matched by: 'A', 'B', 'Both Equally', or 'Neither'. If
image C is a content image, workers are posed the same question with
'content' replacing 'style'. For each competing
algorithm except \cite{gu2018arbitrary} we test three sets of
hyper-parameters, the defaults recommended by the authors, the same
with $\frac{1}{4}$ of the content weight (high stylization), and the
same with double the content weight (low stylization). Because these modifications to content weight did not alter the behavior of ~\cite{gatys2016image} significantly we also tested ~\cite{gatys2016image} with $\frac{1}{100}$ and $100\times$ the default content weight. We also test our method with $4\times$ the content weight. We could only test the default hyper-parameters of ~\cite{gu2018arbitrary} because
the author's code did not expose content weight as a
parameter to users. We test all possible pairings of A and B between
different algorithms and their hyperparameters (i.e. we do not compare
an algorithm against itself with different hyperparameters, but do
compare it to all hyperparameter settings of other algorithms). In
each presentation, the order of output (assignment of methods to A or B in
the interface) was randomized. Each pairing was voted on by an average
of 4.98 different workers (minimum 4, maximum 5), 662 workers in
total. On average, 3.7 workers agreed with the majority vote for each
pairing.

For an algorithm/hyper-parameter combination we defined its content score to be the number of times it was selected by workers as having better or equal content preservation relative to the other output it was shown alongside, divided by the total number of experiments it appeared in. This is always a fraction between 0 and 1. The style score is defined analogously. We plot these scores in Figure \ref{fig:mturk_results_fig_strotss}, separated by regime. The score of each point is computed over 1580 pairings on average (including the same pairings being shown to distinct workers, minimum 1410, maximum 1890). We also provide qualitative examples of outputs produced by each algorith in Figure ~\ref{fig:comp_fig}. For a given content score, STROTSS provided a higher style score than prior work. 

\newcommand{\texpath}{chapters/STROTSS/Figures/texture_syn}
\newcommand{\ablationrowApp}[2]{
 & 
 \includegraphics[height=#2\textwidth]{\texpath/#1_moment}&
 \includegraphics[height=#2\textwidth]{\texpath/#1_pixDirect}&
 \includegraphics[height=#2\textwidth]{\texpath/#1_singleScale}&
 \includegraphics[height=#2\textwidth]{\texpath/#1_l2Ground} & \includegraphics[height=#2\textwidth]{\texpath/#1_c} \\
 & $\ell_{m}$ & Optimize Pixels & Single Scale & $L_2$ Ground Metric & Content\\
\includegraphics[height=#2\textwidth]{\texpath/#1_remd1} & \includegraphics[height=#2\textwidth]{\texpath/#1_remd2} & \includegraphics[height=#2\textwidth]{\texpath/#1_remd} & \includegraphics[height=#2\textwidth]{\texpath/#1_MomRemd}&
\includegraphics[height=#2\textwidth]{\texpath/#1_standard}&
\includegraphics[height=#2\textwidth]{\texpath/#1}\\
$\ell_{R_A}$ & $\ell_{R_B}$ & $\ell_{r}$ & $\ell_{r}$ + $\ell_{m}$& Ours & Style}

\begin{figure}[H]
	\centering
	\begin{tabular}{ccccccc}
	\ablationrowApp{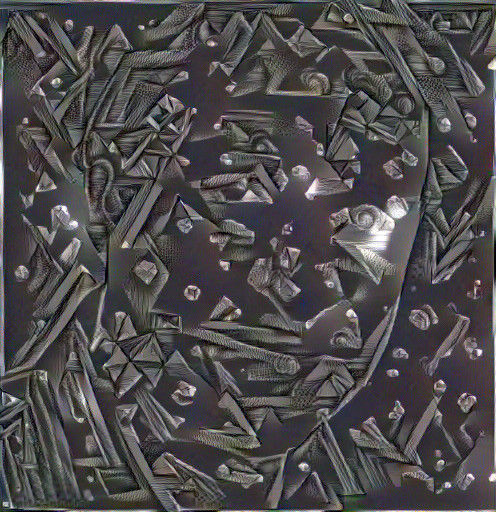}{0.13} \\
	\ablationrowApp{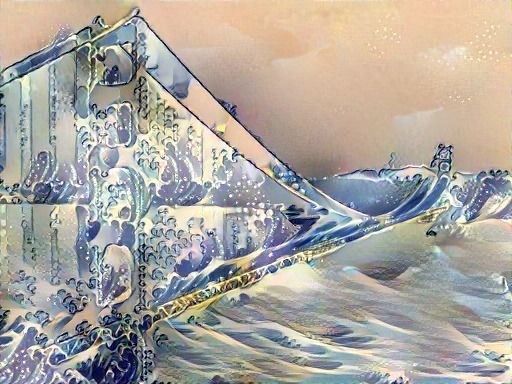}{0.09} \\
	\ablationrowApp{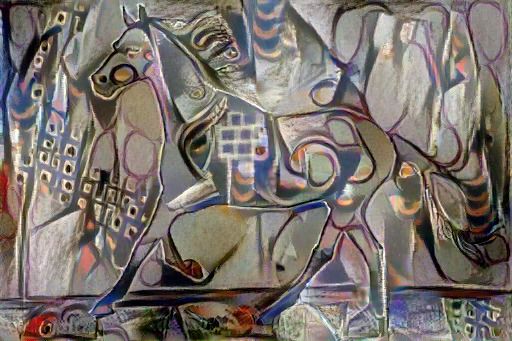}{0.09} \\
	\end{tabular}
	\caption{Ablation study of effects of our proposed style terms with low content loss ($\alpha=4.0$). See Section \ref{sec:strotss_ab} for discussion.}
	\label{fig:mturk_ab_fig}
\end{figure}

\subsection{Ablation Study} \label{sec:strotss_ab}

In Figure \ref{fig:mturk_ab_fig} we explored the effect of different
terms of our style loss, which is composed of a moment-matching loss
$\ell_m$, the Relaxed Earth Movers Distance $\ell_r$, and a color
palette matching loss $\ell_p$. As seen in Figure
\ref{fig:mturk_ab_fig}, $\ell_m$ alone does a decent job of
transferring style, but fails to capture the larger structures of the style image. $\ell_{R_A}$ alone does not make use of the entire distribution of style features, and reconstructs content more poorly than $\ell_{r}$. $\ell_{R_B}$ alone encourages every style feature to have a nearby output feature, which is too easy to satisfy. Combining $\ell_{R_A}$ and $\ell_{R_B}$ in the relaxed earth movers distance $\ell_r$ results in a higher quality output than either term alone, however because the ground metric used is the cosine distance the magnitude of the features is not constrained, resulting in saturation issues. Combining $\ell_{r}$ with $\ell_{m}$ alleviates this, but some issues with the output's palette remain, which are fixed by adding $\ell_p$. We also explore the effect of several of STROTSS implementation details. 'Optimize Pixels' refers to performing gradient descent on pixel values of the output directly, instead of the entries of a laplaccian pyramid (STROTSS's default). In 'Single Scale' we perform 800 updates at the final resolution, instead of 200 updates at each of four increasing resolutions. In '$\ell_2$ Ground Metric' we replace the ground metric of the Relaxed EMD with euclidean distance (instead of STROTSS's default, cosine distance). 

\subsection{Relaxed EMD Approximation Quality}\label{subsec:approx}
To measure how well the Relaxed EMD approximates the exact Earth
Movers Distance we take each of the 900 possible
content/style pairings formed by the 30 content and style images used in our
AMT experiments for the unpaired regime. For each pairing we compute the REMD between 1024 features
extracted from random coordinates, and the exact EMD
based on the same set of features.
We then analyze the distribution of $\frac{REMD(A,B)}{EMD(A,B}$
Because the REMD is a lower bound, this quantity is always
$\le$1. Over the 900 image pairs, its mean was 0.60, with standard deviation
0.04. On one hand the REMD seems to be a loose lower bound, on the other, the small standard deviation indicates it tends to be off by a constant factor in this setting, making it a good proxy to optimize. A better EMD approximation, or
one that is an upper bound rather than a lower bound, may yield better
style transfer
results. On the other hand the REMD is simple to compute, empirically easy to optimize, and yields good results.

\subsection{Timing Results}
We compute our timing results using a Intel i5-7600 CPU @ 3.50GHz CPU,
and a NVIDIA GTX 1080-TI GPU. We use square style and content images
scaled to have the edge length indicated in the top row of Table
\ref{tab:timing}. For inputs of size 1024x1024 the methods from
\cite{li2016combining} and \cite{mechrez2018contextual} ran out of
memory ('X' in the table). Because the code provided by the authors \cite{gu2018arbitrary} only runs on Windows, we had to run it on a different computer. To approximate the speed of their method on our hardware we project the timing result for 512x512 images reported in their paper based on the relative speedup for \cite{li2016combining} between their hardware and ours. For low resolution outputs our method is relatively slow, however it scales better for outputs with resolution 512 and above relative to \cite{li2016combining} and \cite{mechrez2018contextual}, but remains slower than \cite{gatys2016image} and our projected results for \cite{gu2018arbitrary}.

\begin{table}[t]
\centering
\begin{tabular}{c||c|c|c|c|c}
	\textbf{Image size} & \textbf{64} & \textbf{128} & \textbf{256} & \textbf{512} & \textbf{1024}\\
	\hline \hline
	Ours & 20 & 38 & 60 & 95 & 154\\
	Gatys & 8 & 10 & 14 & 33 & 116\\
	CNNMRF & 3 & 8 & 27 & 117 & X\\
	Contextual & 13 & 40 & 189 & 277 & X\\
	Reshuffle & - & - & - & \textit{69*} & -\\
\end{tabular}
	\caption{Timing comparison (in seconds) between our methods
          and others. The style and content images had the same
          dimensions and were square. *: a projected
          result, see text for details. -: we were not able to project these results. X: the method ran out of memory.}
	\label{tab:timing}
\end{table}

\section{Conclusion}
"Style Transfer by Relaxed Optimal Transport and Self-Similarity" proposed novel definitions of style and content and demonstrated that the resulting algorithm compared favorably to prior work, both in terms of stylization quality and content preservation. Through our user study and ablations we demonstrate that style-similarity losses which more accurately measure the distance between distributions of features leads to style transfer outputs of higher visual quality. However, the absolute quality of STROTSS's outputs leave something to be desired, and rarely are entirely free of artifacts or create the convincing illusion of having been created by the same media as the style (i.e. being a painting rather than the result of an image processing algorithm). Another shortcoming is it's speed, which is much too slow to be used in most creative pipelines. In Chapter \ref{chpt:nnst} we describe our efforts to make progress on both fronts (quality and speed). Another notable flaw (shared with other style transfer algorithms up to this point) is the narrow definition of style as texture, in the next chapter we will explore endowing style transfer algorithms with the ability to alter geometry and proportion as well.

\chapter{Deformable Style Transfer}
\label{chpt:dst}

\newcommand{\lcon}{L_{\text{content}}} 
\newcommand{\lsty}{L_{\text{style}}} 
\newcommand{\lmatch}{L_{\text{warp}}} 

\begin{figure}[h]
\centering
\includegraphics[width=0.95\linewidth]{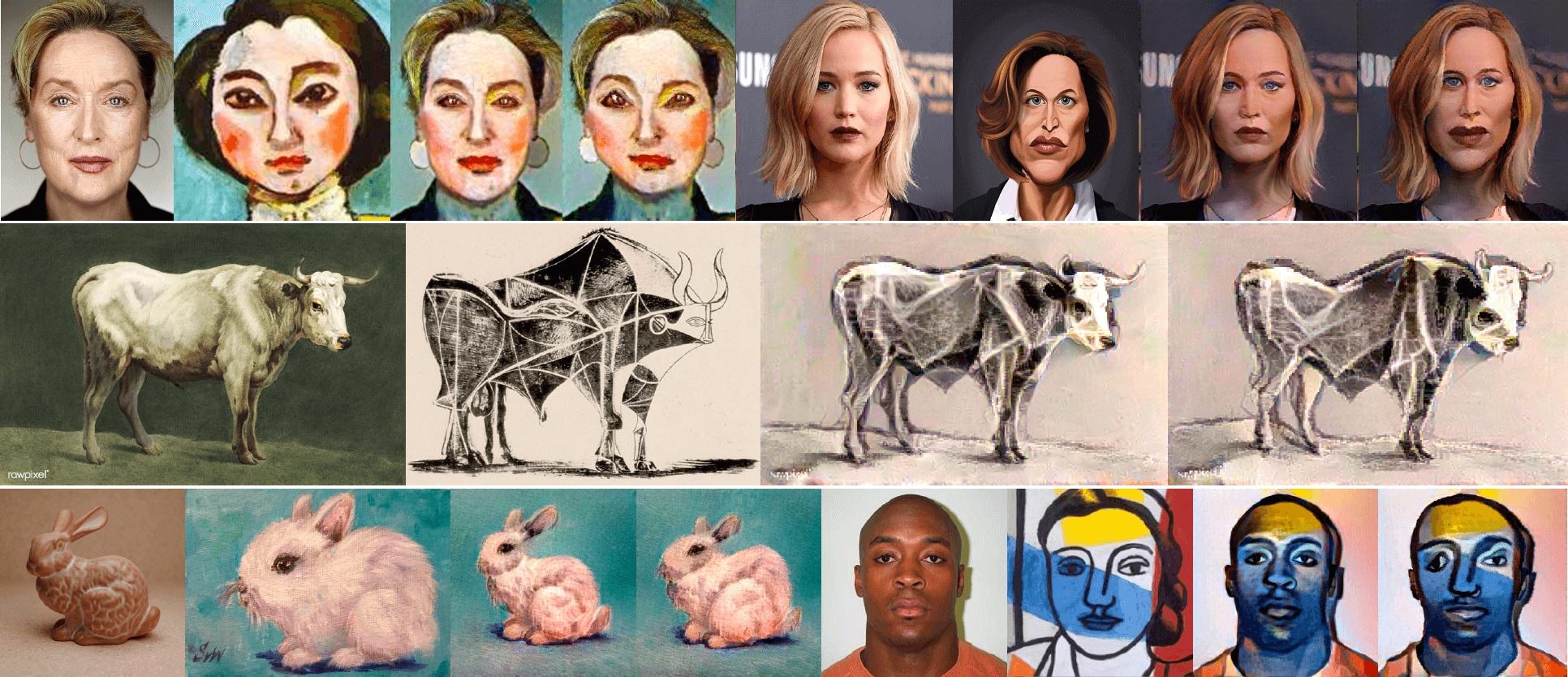}
\caption{Examples of the deformations produced by our proposed style transfer framework 'Deformable Style Transfer' (DST). Each quartet of images consists of: (1) the content input, (2) the style input, (3) the output of DST without geometric deformation, and (4) the output of DST with geometric deformation.}
\end{figure}

  Shape and form play a vital role in defining the distinctive style of artists across many types of media, for example painting (e.g. Picasso, Modigliani, El Greco) and sculpture (e.g., Botero, Giacometti). Indeed, the formal discussions of art historians and other experts on image creation almost always include shape (in 2D artwork) and form (in 3D artwork) as important topics when analyzing a work of art~\cite{hofstadter1983metamagical,elkins1996style}. However, while stylization algorithms specifically focused on faces had begun to incorporate geometry into their definition of style, no solution for arbitrary domains existed. Motivated by this we proposed 'Deformable Style Transfer'. A framework for endowing existing style transfer methods with the ability to make changes to a content input's shape, guided by the style input. The paper this chapter is based on appeared in ECCV 2020 was joint work with Sunnie Kim, Jason Salavon, and my advisor Greg Shakhnarovich.

\section{Introduction}

Style transfer methods which do not explicitly include geometry in their definition of style almost always keep the perceived geometry of the content unchanged in the final output. As a result these outputs are easily identified as altered or ``filtered" versions of the content image, rather than novel works of art created using the content image as a reference. Most optimization-based style transfer methods directly optimize the pixels of an output image using gradient descent. However, modifications to the spatial position or relative size of features in the output image require altering many pixels in sync, making a representation based only on raw pixels poorly suited to stylizing shape and proportion. We take a first step towards solving this problem by optimizing over an additional set of variables, the parameters of a deformation field defining a warped image.

Deformable Style Transfer (DST), takes two images as the input: a content image and a style image. We assume both images share a domain and have some approximate alignment (e.g. both are images of sitting animals). This is a general scenario likely to arise in recreational or artistic uses of style transfer, as well as in tasks such as data augmentation. The nature of this problem makes \emph{learning} to transfer style challenging since the variation in unconstrained domains and styles is difficult to capture in any feasible training set. Therefore, like other style transfer work in this setting, we develop an optimization-based method, leveraging a pre-trained and fixed feature extractor derived from a convolutional network (CNN) trained for ImageNet classification.

Prior to publishing DST there had been some recent work on learning geometric style, using an explicit model of landmark constellations~\cite{Yaniv_2019_ACM} or a deformation model representing a specific style~\cite{Shi_2019_CVPR}. These methods required a collection of images in the chosen style, and were specialized to face and not applicable to our more general setting. Nonetheless, we compared DST with these as a baseline on faces (see Section~\ref{sec:results}), and found DST's outputs to be of equal if not better visual quality.

The key idea of DST is to find a smooth deformation (i.e. spatial warping) of the content image that brings it into spatial alignment with the style image. This deformation is guided by a set of matching keypoints, chosen to maximize the feature similarity between paired keypoints of the two images. After roughly aligning the paired keypoints with a rigid rotation and scaling, a simple $L_2$ loss encourages warping our output image in such a way that the keypoints become spatially aligned. This deformation loss is regularized with a total variation penalty to reduce artifacts due to drastic deformations, and combined with the more traditional style and content loss terms. DST's joint, regularized objective simultaneously encourages preserving content, minimizing the style loss, and obtaining the desired deformation, weighing these goals against each other. This objective can be minimized using gradient descent.

To summarize the contributions of this work:
\begin{itemize}
\item We proposed an optimization-based framework that endowed existing style transfer algorithms with the ability to deform a content image to match the geometry of a style image. Our flexible formulation also allows explicit user guidance and control of stylization tradeoffs.
\item We demonstrated, for the first time, geometry-aware style transfer in a one-shot scenario. In contrast to previous works limited to human faces, DST worked for images in arbitrary domains, with the assumption that the content and style have similar contents (the 'Paired' regime used to evaluate STROTSS) and have some approximate alignment.
\item We evaluated DST on a range of style transfer instances, with images of faces, animals, vehicles, and landscapes, and through a user study demonstrate that augmenting existing style transfer algorithms with DST led to dramatically improve the perceived stylization quality at minimal cost to the perceived content preservation.
\end{itemize}

\section{Related Work}\label{sec:related}

Until shortly before 'Deformable Style Transfer' was published, style transfer methods could not transfer geometric style at all and were limited to transferring color and texture. However, several recent works had tackled geometric style transfer for faces. Cole et al.~\cite{Cole_2017_CVPR} focused on generating frontal, uniformly lit, views of faces from normal photographs. One component of their method was estimating a face 'texture' by predicting a sparse set of face landmarks, then using these as keypoints to warp the input face onto a canonical set of front facing landmarks using thin-plate cubic splines (which interpolate the sparse offsets defined by keypoint pairs to form a dense warping field which defines the new location in the output of every pixel in the input). Their proposed warping module was differentiable and thus easily added to a system trained end-to-end through backpropagation. Subsequent works combined image warping with feed-forward style transfer techniques to learn both textural and geometric stylizations of faces. CariGAN~\cite{Li_2018} translates a photo to a caricature by training two GANs. One is a standard texture transfer GAN, which alters the texture of a photograph to resemble a caricature (similar to the stylization loss of \cite{sanakoyeu2018style} except the style is caricatures instead of Van Gogh or Cezanne); The second GAN learns to predict offsets to facial landmarks extracted from the input photo which results in transformed landmarks that match the distribution of caricature landmarks. WarpGAN ~\cite{Shi_2019_CVPR}, tackle the same task but with a more flexible warping module, learning to predict both the locations of keypoints (rather than using a canonical set of landmarks) and their displacements. Both CariGAN and WarpGAN rely on a dataset of caricatures with manually labeled keypoints. Face of Art (FoA)~\cite{Yaniv_2019_ACM} trains a neural network model to automatically detect 68 canonical facial landmarks in artistic portraits (supervised using a labeled dataset of landmarks in various artists' portraits). These landmarks can then be used as the keypoints of a warp field to match the geometry of an input face to a target artistic portrait.

The main distinction of DST from these efforts was that it was not limited to human faces (or any other particular domain) and did not require offline training on a specially prepared dataset. In terms of methodology, FoA and CariGan separately transfer texture and geometry, while DST transfers them jointly. WarpGAN treats texture and geometry jointly, but had to learn a warping module from paired examples of face photos and caricatures. We show in Section~\ref{sec:results} that results of our more general method, even when applied to faces, are competitive or even superior to the results of these two face-specific methods.

Deformable Style Transfer relied on finding correspondences (i.e. paired keypoints) between two input images in an arbitrary domain. Several recent works had developed techniques leveraging CNN-derived features to find such correspondences. Fully Convolutional Self-Similarity ~\cite{kim2017fcss} is a descriptor for dense semantic correspondence that uses local self-similarity to match keypoints among different instances within the same object class. Neural Best-Buddies (NBB)~\cite{Aberman_2018_ACM} is a more robust method (which can operate across object classes) for finding a set of sparse correspondences by finding mutual nearest neighbors in the deep layers of a pretrained network, then spatially refining these correspondences by examining activations in earlier layers of the network. We used NBB as the basis for generating correspondences between the content and style images, and give a more detailed description of NBB and our modifications in next section.
\section{Geometry Transfer via Correspondences} \label{sec:points}

One path for introducing geometric style transfer is establishing spatial associations between the content and style images, then finding a \emph{deformation} that brings the content image into (approximate) alignment with the style image. Assuming they share a domain and have similar geometry (e.g. both are images of front-facing cars), we can aim to find meaningful spatial correspondences to define the deformation. The correspondences specify displacement ``targets'', derived from the style image, for keypoints in the content image. Thin-plate spline interpolation~\cite{glasbey1998review} can extend this sparse set of displacements to a full displacement field specifying how to deform every pixel in the output image. 

\begin{figure}
\includegraphics[width=0.95\linewidth]{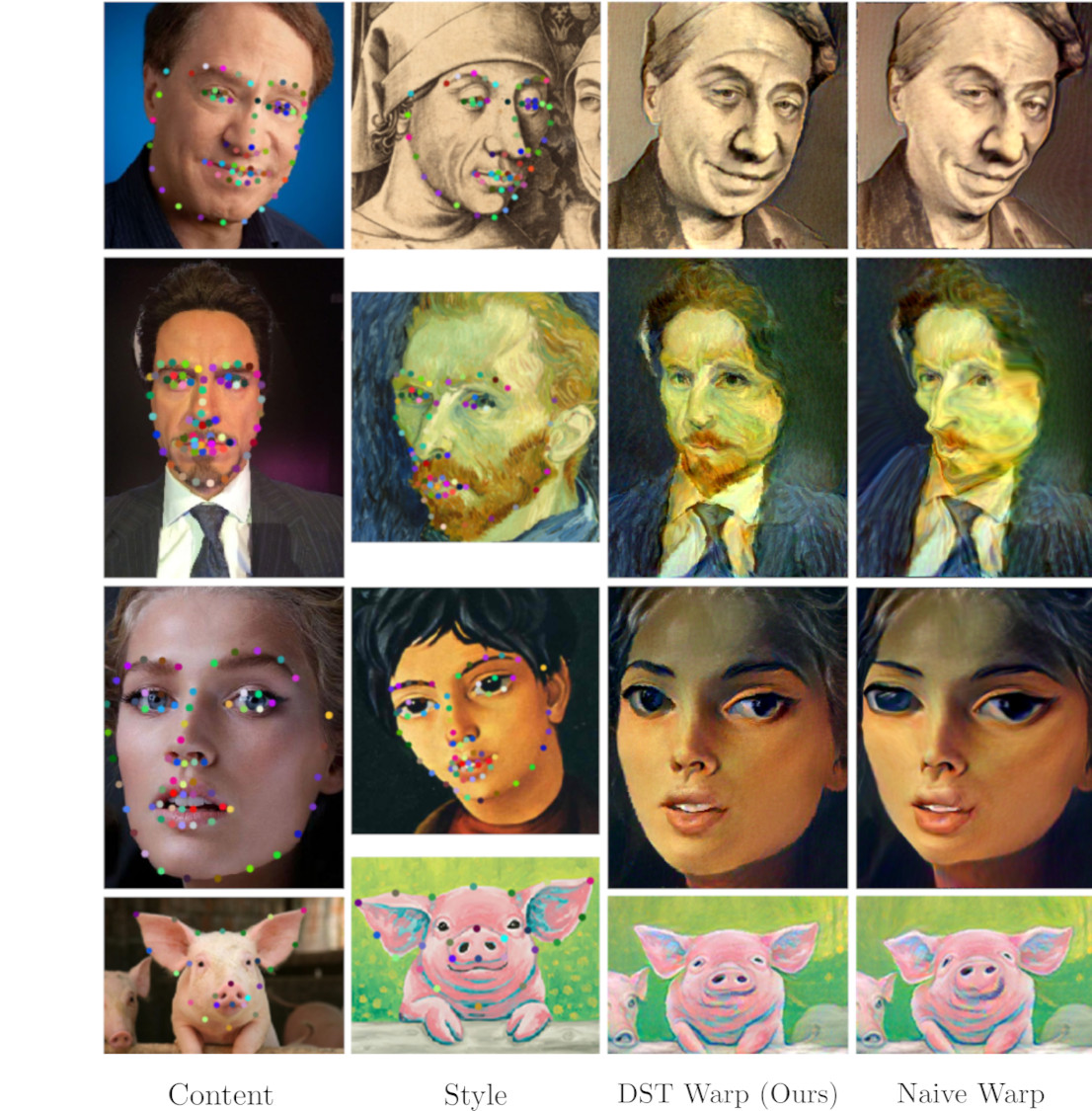}
\caption{DST can produce geometric stylizations using keypoints from a variety of sources. Rows 1-3 show the outputs generated by DST using keypoints taken from FoA, Row 4 shows an output generated by DST using manually selected keypoints. Keypoints are overlayed on the content and style images with matching points in the same color. Naive warp indicates output of style transfer warped source points on top of target points, rather than jointly optimizing the warp with the content and style loss (DST Warp)}\label{fig:foapoints}
\end{figure}

\subsection{Finding and Cleaning Keypoints}
If we fix a domain and assume availability of a training set drawn from the domain, we may be able to learn a domain-specific mechanism for finding salient and meaningful correspondences. This can be done through facial landmark detection~\cite{Yaniv_2019_ACM} or through learning a data-driven detector for relevant points~\cite{kim2017fcss,Shi_2019_CVPR}. Alternatively, we could expect a user interacting with a style transfer tool to manually select points they consider matching in the two images. If matching points are provided by such approaches, they can be used in DST as we show in Figure~\ref{fig:foapoints}. However, we were interested in a more general scenario, a one-shot, domain-agnostic setting where we may not have access to such points. In this setting we turned to Neural Best Buddies (NBB), a generic method for point matching between images. 

NBB finds a sparse set of correspondences between two images that could be from different domains or semantic categories. It utilizes the hierarchy of features extracted by a pre-trained CNN. Starting from the deepest layer, NBB searches for pairs of correspondences that are mutual nearest neighbors, filters the matches to only keep those those between feature vectors with high norm (high norm features typically correspond to 'interesting' regions of the image such as perceptually salient edges or distinctive semantic objects, while low norm features typically correspond to regions of homogeneous texture), then percolates the matches through sequentially shallower layers until reaching the original pixels. The deepest layer has coarse spatial resolution, but in shallower layers the spatial resolution increases and allows refining the matches spatial location. After the matches have been propagated back to the pixel level they spatially clustered using $k$-means and $k$ keypoint pairs are returned. 

However, the keypoint pairs returned by NBB were too noisy and not sufficiently spread out for our purposes. To provide better guidance for geometric deformation, we modified NBB to get a cleaner and better spatially-distributed set of pairs. Specifically, we remove the final clustering step and return all pixel-level correspondences, usually on the order of hundreds of correspondence pairs. Then we greedily select the keypoint pair with the highest activation value (i.e. the sum of the norm of the matched features through the feature hierarchy) that is at least $10$ pixels away from any already selected keypoint. We select up to $80$ keypoint pairs and filter out keypoints with small activation values. After the initial selection, we align the keypoints in the style image with the content image by finding an linear map (on coordinates) that minimizes the squared distance between the two point clusters~\cite{Umeyama1991}. After superimposing the aligned style keypoints onto the content image (with the content keypoints) consider paired keypoints being connected by a line, if two lines cross we remove the pair of keypoints with lower activation value. If more than two lines cross we continue this repeat this process there are no crossing lines. This prevents using a set of target keypoints for DST which parameterize a discontinuous deformation field.

\begin{figure}[htbp!]
\begin{center}
\begin{tabular}{ccccc}
	\includegraphics[height =0.14\linewidth]{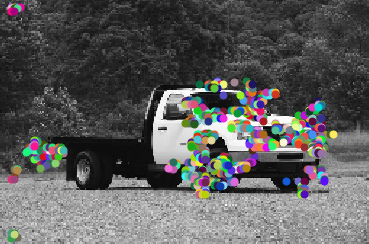} &   
	\includegraphics[height=0.14\linewidth]{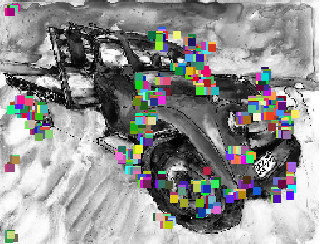} &
	\includegraphics[height=0.14\linewidth]{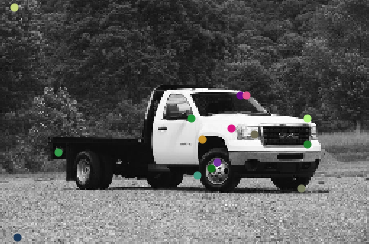} &
	\includegraphics[height=0.14\linewidth]{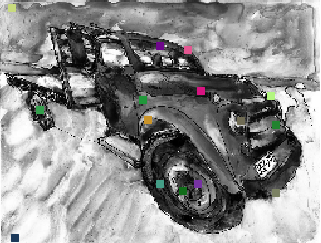} \\
	(a) & (b) & (c) & (d) \\
	\includegraphics[height=0.14\linewidth]{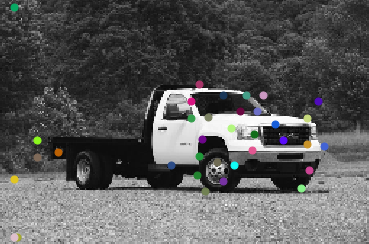} &
	\includegraphics[height=0.14\linewidth]{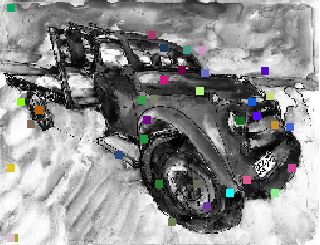} &
	\includegraphics[height=0.14\linewidth]{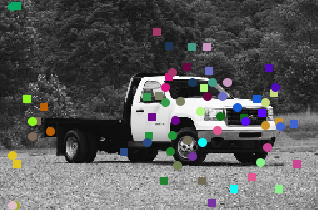} &
     \includegraphics[height=0.14\linewidth]{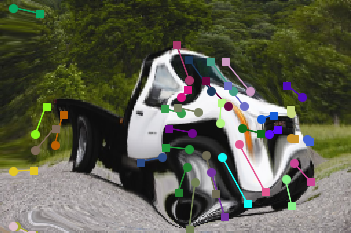} \\
	(e) & (f) & (g) & (h)
\end{tabular}
\end{center}
\vspace{-0.5cm}
\caption{An image can be spatially deformed by moving a set of source points to a set of target points. Matching keypoints are indicated by color. (a) Content image with all correspondences. (b) Style image with all correspondences. (c) Content image with original NBB keypoints. (d) Style image with original NBB keypoints. (e) Content image with our selected keypoints. (f) Style image with our selected keypoints. (g) Content image with keypoints aligned by just matching the centers. (h) Content image warped with keypoints aligned by a similarity transformation. The lines indicate where the circle source points move to (square target points). Figure is best viewed zoomed-in on screen.}
\label{fig:points}
\end{figure}

We only used this post-processing procedure as a modification of NBB. If keypoints were provided by FoA, manual selection, or other non-NBB methods, we skipped the filtering process and  simply superimpose the style keypoints onto the content using the linear alignment of ~\cite{Umeyama1991}. We refer to the keypoints in the content image as the ``source points'' and the corresponding keypoints in the style image mapped onto the content image as the ``target points.'' This process is illustrated in Figure~\ref{fig:points}.

\subsection{Differentiable Image Warping}
We specify an image deformation by a set of source keypoints $P=\{p_1,\ldots,p_k\}$ and the associated 2D displacement vectors $\theta=\{\theta_1,\ldots,\theta_k\}$. For each source keypoint $p_i$ the displacement $\theta_i$ defines the \emph{destination} coordinates $p_i+\theta_i$. 

Following~\cite{Shi_2019_CVPR} we use thin-plate spline
interpolation~\cite{glasbey1998review} to produce a dense flow field from the coordinates of an unwarped image $I$ to a warped image $W(I,\theta)$. Given a 2d coordinate $q = [q_x,q_y]$, the source kepoints $P=\{p_1,\ldots,p_k\}$, kernel function $\phi (r) = \|r\|^2$, and parameters $w,v,b$ consider the flow field $f_\theta(q;w,v,b)$:
\begin{align}
f_\theta(q;w,v,b) = \sum_{i=1}^{k} w_i \phi(||q - p_i-\theta_i||) + v^T q + b
\end{align}

Thin-plate spline interpolation is a closed-form procedure to find the parameters $w,v,b$ which minimize:
\begin{align}
\min_{w,v,b}& \int_{q_x} \int_{q_y}\Big[\Big(\frac{\delta^2 f_\theta(q;w,v,b)}{\delta q_x^2}\Big)^2 + 2\Big(\frac{\delta^2 f_\theta(q;w,v,b)}{\delta q_x \delta q_y}\Big)^2 + \Big(\frac{\delta^2 f_\theta(q;w,v,b)}{\delta q_y^2}\Big)^2\Big] \delta q_y \delta q_x \\
\text{s.t.}&\hspace{0.5cm}\forall i\hspace{0.5cm}f_\theta(p_i;w,v,b) = p_i+\theta_i \hspace{1cm}  
\end{align}

Given $w,v,b$ we can define the inverse coordinate mapping function (i.e. from which pixel coordinates in the unwarped output should we derive the color of pixel $q$ in the warped output). The color of each pixel in the output can then be generated via bilinear sampling on the unwarped output based on the inverse mapping function (as the output coordinate will generally not be an integer). This entire warping module is differentiable with respect to $\theta$, allowing it be used in an end-to-end optimized system.

\section{Spatially Guided Style Transfer}\label{sec:approach}

The input to DST consists of a content image $C$, a style image $S$, and aligned keypoint pairs $P$ (source) and $P'$ (target). Recall that these points don't have to be infused with explicit domain- or category-specific semantics. DST optimizes the stylization parameters (usually the pixels of the output image) $O$ and the deformation parameters $\theta$. The final output is the warped stylized image $W(O, \theta)$. Note that this still falls within the original neural style transfer optimization framework, we optimize an output image to satisfy a content and style loss; however, now the output image is parameterized not only by its pixels, but also by the keypoint offsets $\theta$ which define the warp field.

\subsection{Content and Style Loss Terms} \label{sec:contentstyle}

DST can be used with any one-shot, optimization-based style transfer method with a content loss and a style loss. We evaluate its effects when combined with the original neural style transfer algorithm of Gatys \cite{gatys2016image} and STROTSS. Each method defines a content loss $\lcon(C,O)$ and a style loss $\lsty(S,O)$. When using DST with a base style transfer method, we do not change anything about $\lcon$. The style loss of DST is composed of two terms
\begin{equation}
\lsty(S, O) + \lsty(S, W(O, \theta)).\label{eq:lsty}
\end{equation}
The first loss term is between the style image $S$ and the \emph{unwarped} stylized image $O$. The second term is between $S$ and the spatially deformed stylized image $W(O, \theta)$, with $\theta$ defining the deformation as per Section~\ref{sec:points}. Minimizing Eq.~\eqref{eq:lsty} is aimed at finding a good stylization both with and without spatial deformation. This way we force the stylization parameters $O$ and the spatial deformation parameters $\theta$ to work together to produce a harmoniously stylized and spatially deformed final output $W(O, \theta)$.

\subsection{Deformation Loss Term}
Given a set of $k$ source points $P$ and matching target points $P'$, we define the deformation loss as
\begin{equation}\label{eq:lmatch}
  \lmatch(P,P',\theta) = \frac{1}{k}\sum_{i=1}^k \| p'_i - (p_i+\theta_i)\|_2,
\end{equation}
where $p_i$ and $p_i'$ are the $i$-th source and target point coordinates. Minimizing Eq.~\eqref{eq:lmatch} with respect to $\theta$ seeks a set of displacements that move the source points to the target points. This term encourages the geometric shape of the stylized image to become closer to that of the style.

Aggressively minimizing the deformation loss may lead to significant artifacts, due to errors in keypoint selection/matching or incompatibility between the content and style geometry. To avoid these artifacts, we add a regularization term encouraging smooth deformations. Specifically, we use the (anisotropic) total variation norm of the 2D warp field $f$ normalized by its size
\begin{equation}
  R_\text{TV}(f) = \frac{1}{\text{W}\times \text{H}} \sum_{i=1}^{\text{W}} \sum_{j=1}^{\text{H}} \|f_{i+1, j} - f_{i, j}\|_1 +  \|f_{i, j+1} - f_{i, j}\|_1.
\end{equation}
This regularization term smooths the warp field by encouraging nearby pixels to move in a similar direction. 

\subsection{Joint Optimization}

\begin{figure}[htbp!]
\begin{center}

\includegraphics[width=0.85\linewidth]{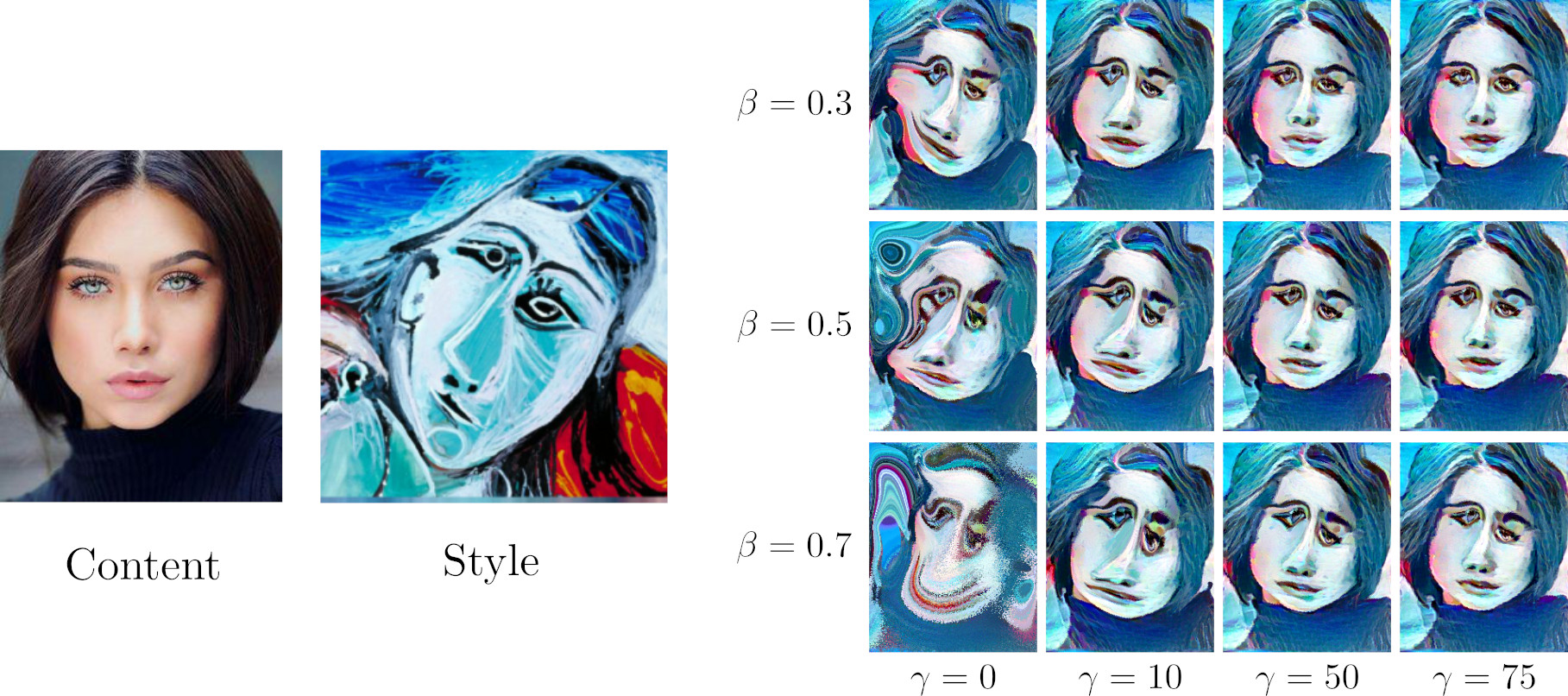}
\caption{DST outputs with varying $\beta$ and $\gamma$ using STROTSS as the base method. Image in the upper right corner (low $\beta$, high $\gamma$) has the least deformation, and the image in the bottom left corner (high $\beta$, low $\gamma$) has the most deformation.} \label{fig:gammabeta}
\end{center}
\end{figure}

Putting everything together, the objective function of DST is
\begin{align}
  L(O, \theta, C, S, P, P') \label{eq:obj}
  &=\,\alpha \lcon(C,O) \\
  &\phantom{=}\,+\, \lsty(S,O)\,+\,\lsty(S,W(O,\theta)) \nonumber \\
  &\phantom{=}\,+\, \beta\lmatch(P, P', \theta) \nonumber \\
  &\phantom{=}\,+\, \gamma R_\text{TV}(f_\theta), \nonumber
\end{align} 
where $X$ is the stylized image and $\theta$ parameterizes the spatial deformation. Hyperparameters $\alpha$ and $\beta$ control the relative importance of content preservation and spatial deformation to stylization. Hyperparameter $\gamma$ controls the amount of regularization on the spatial deformation. The effect of varying $\alpha$ is analyzed in~\cite{gatys2016image, kolkin2019style}. The effect of changing $\beta$ and $\gamma$ is illustrated in Figure~\ref{fig:gammabeta}. We use standard iterative techniques such as stochastic gradient descent or L-BFGS to minimize Eq.~\eqref{eq:obj} with respect to $O$ and $\theta$. Our implementation of DST can be found at \url{https://github.com/sunniesuhyoung/DST}.

\section{Evaluation}\label{sec:results}

\begin{figure*}
\begin{center}
	\includegraphics[width=0.92\textwidth]{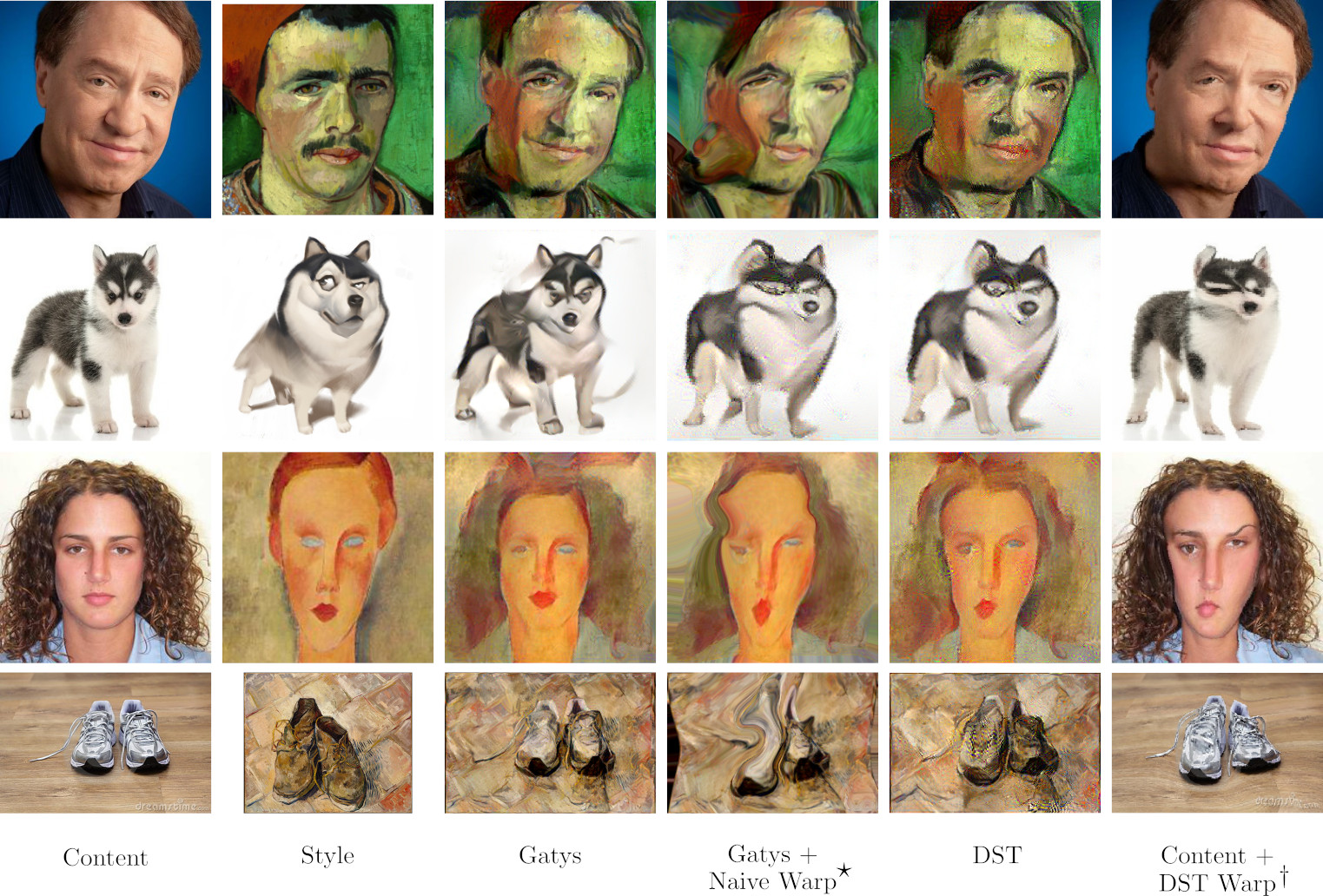}
\end{center}
\caption{DST results with Gatys.$~^{\star}$Naively warped by moving source points to target points.$~^{\dag}$Warp learned by DST applied to the content image.}
\label{fig:gatysresults}
\end{figure*}

\begin{figure*}
\begin{center}
  \includegraphics[width=0.96\textwidth]{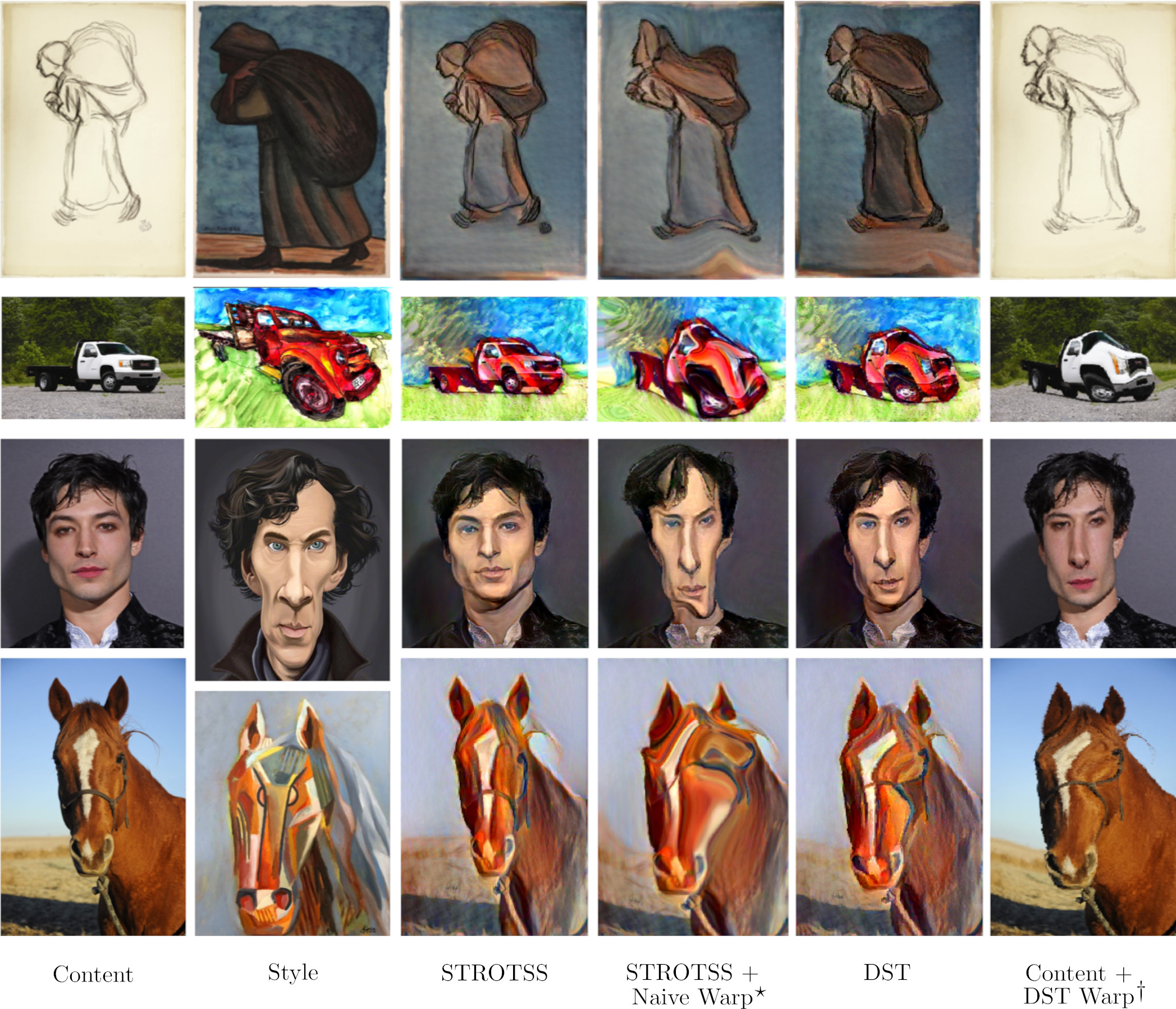}
\end{center}
\caption{DST results with STROTSS.$~^{\star}$Naively warped by moving source points to target points.$~^{\dag}$Warp learned by DST applied to the content image.}
\label{fig:strotssresults}
\end{figure*}

One visually striking effect of DST (when successful) was that the resulting images no longer looked like ``filtered'' versions of the original content, as they often do with standard style transfer methods. We show results of DST combined with Gatys and STROTSS in Figures~\ref{fig:gatysresults} and ~\ref{fig:strotssresults}. For a pair of content and style images, we show the output of DST and the output of unmodified Gatys/STROTSS. To highlight the effect of the DST-learned deformation, we also provide the content image warped by DST and the Gatys/STROTSS output naively warping the source keypoints onto the target ones. While naive warping produces undesirable artifacts, DST finds a warp that harmoniously improves stylization while preserving content. 

As a simple quantitative evaluation, we calculated the (STROTSS) style loss on 185 pairs of DST and STROTSS outputs. Surprisingly, we found that on average this loss was ~7\% higher for DST outputs than STROTSS ones, even for examples we show in Figure~\ref{fig:strotssresults}. While the loss difference is small, this is a mismatch with the human judgment of stylization quality shown in Section~\ref{sec:mturk}.

\subsection{Comparison with FoA and WarpGAN}\label{sec:facecompare}

\begin{figure}[h]
\begin{center}
	\begin{tabular}{cccc}
	\includegraphics[width=0.2\linewidth]{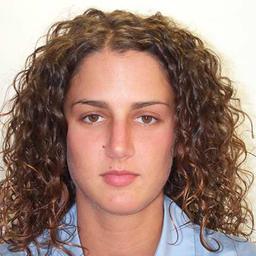} &
	\includegraphics[width=0.2\linewidth]{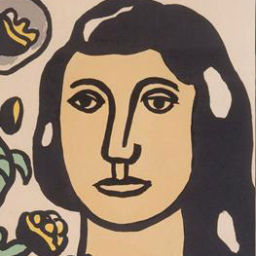} &
	\includegraphics[width=0.2\linewidth]{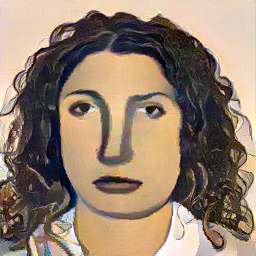} &
	\includegraphics[width=0.2\linewidth]{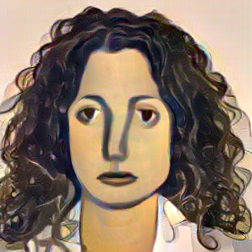} \\
	\includegraphics[width=0.2\linewidth]{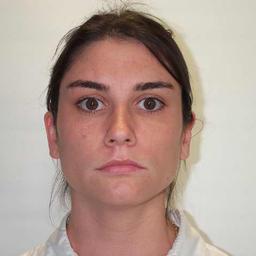} &
	\includegraphics[width=0.2\linewidth]{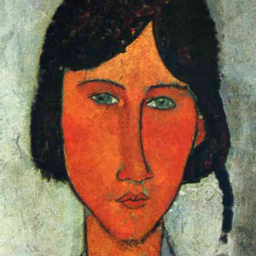} &
	\includegraphics[width=0.2\linewidth]{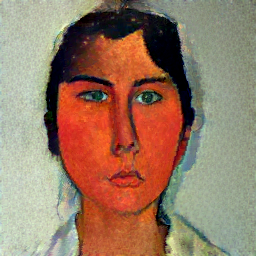} &
	\includegraphics[width=0.2\linewidth]{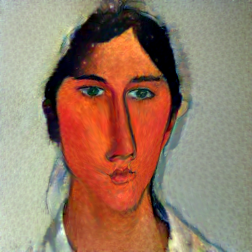} \\
	\includegraphics[width=0.2\linewidth]{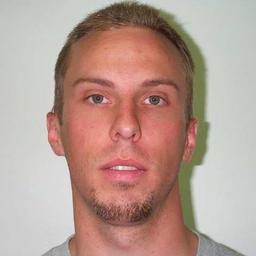} &
	\includegraphics[width=0.2\linewidth]{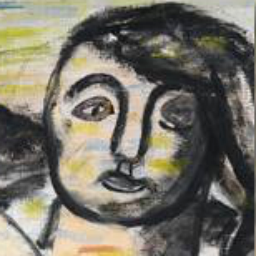} &
	\includegraphics[width=0.2\linewidth]{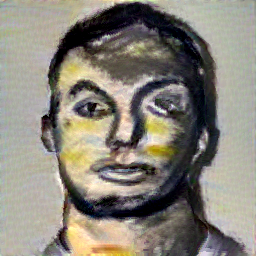} &
	\includegraphics[width=0.2\linewidth]{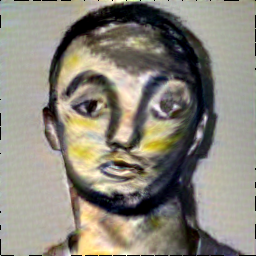}\\
	Content & Style & DST (Ours) & FoA \cite{Yaniv_2019_ACM} \\
    \end{tabular}
\caption{Comparison of DST with Face of Art \cite{Yaniv_2019_ACM} }\label{fig:foacompare}
\end{center}
\end{figure}

\begin{figure}[h]
\begin{center}
	\begin{tabular}{cccc}
	\includegraphics[width=0.2\linewidth]{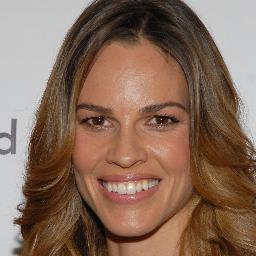} &
	\includegraphics[width=0.2\linewidth]{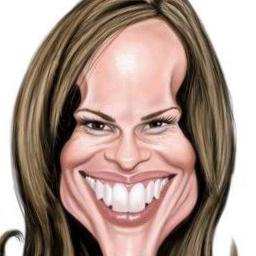} &
	\includegraphics[width=0.2\linewidth]{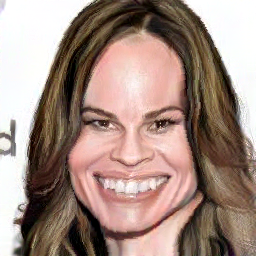} &
	\includegraphics[width=0.2\linewidth]{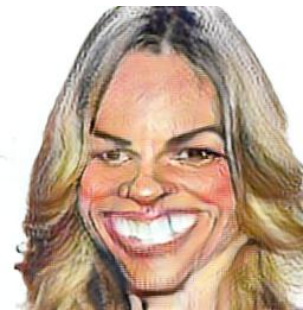} \\
	\includegraphics[width=0.2\linewidth]{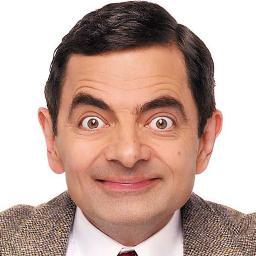}  &
	\includegraphics[width=0.2\linewidth]{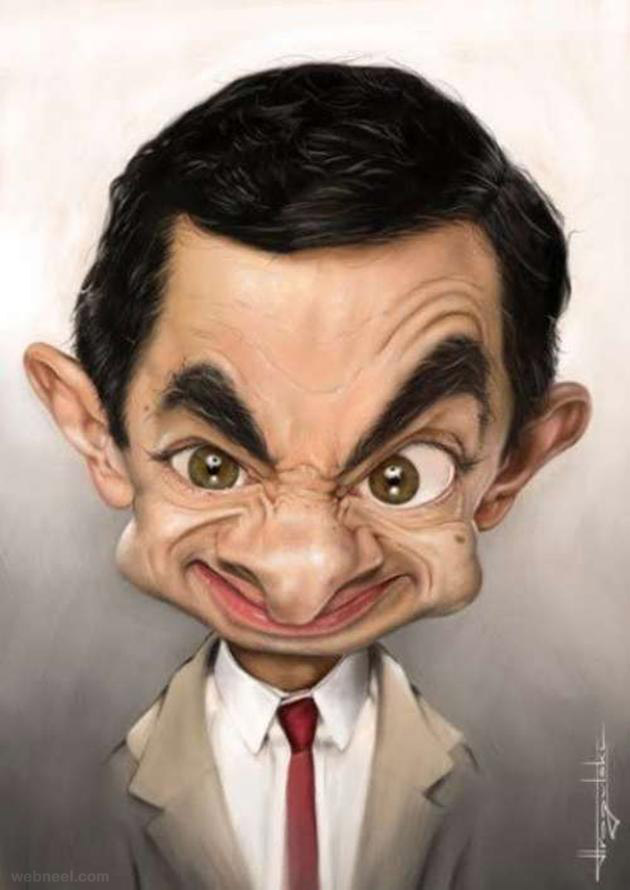} &
	\includegraphics[width=0.2\linewidth]{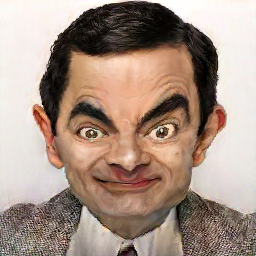} &
	\includegraphics[width=0.2\linewidth]{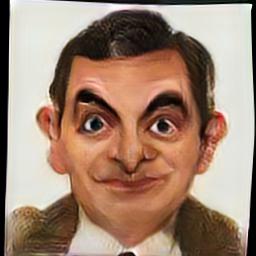}\\
  	\includegraphics[width=0.2\linewidth]{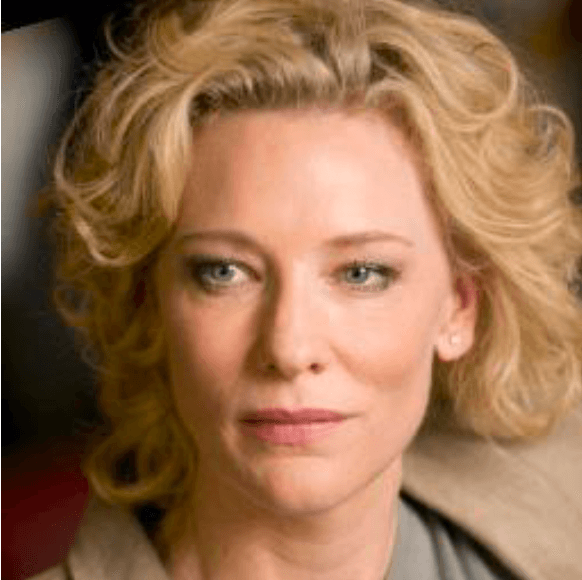}  &
	\includegraphics[width=0.2\linewidth]{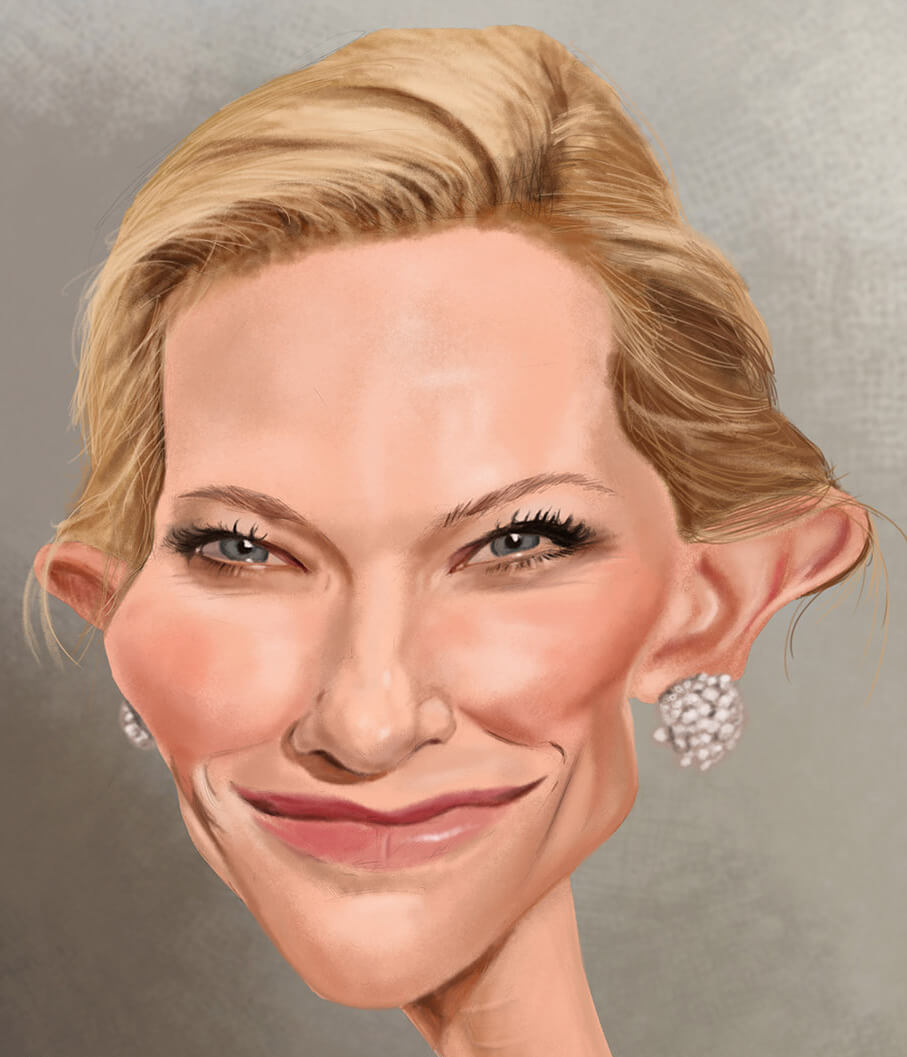} &
	\includegraphics[width=0.2\linewidth]{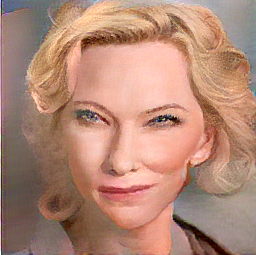} &
	\includegraphics[width=0.2\linewidth]{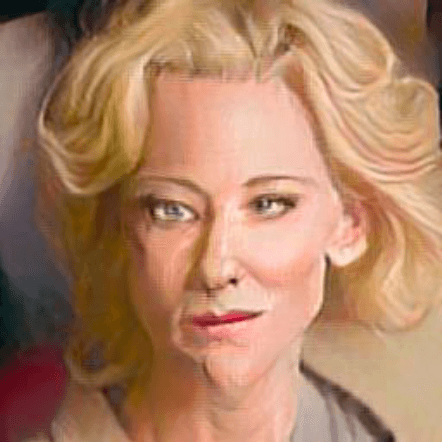}\\
    Content & Style$^\dagger$ & DST & WarpGAN\\
	\end{tabular}\\
\caption{Comparison of DST with WarpGAN. $\dagger$: Note that WarpGAN's does not use a specific style image, so this style image is only used by DST; see text for details.}\label{fig:warpgancompare}
\end{center}
\end{figure}

While DST was the first work to allow open-domain geometry-aware style transfer, we qualitatively compared it with prior work domain-specific methods for human faces. We show results of DST and results of FoA~\cite{Yaniv_2019_ACM} and WarpGAN~\cite{Shi_2019_CVPR} on the same content-style pairs in Figures ~\ref{fig:foacompare} and ~\ref{fig:warpgancompare}. Note that both of these methods require training a model on a dataset of stylized portraits or caricatures, while DST operates with access to only a single content and single style image.

\newcommand{\foaPath}{chapters/DST/Figures/FoA}
\newcommand{\cariPath}{chapters/DST/Figures/CariGan}

DST jointly optimizes the geometric and non-geometric stylization parameters, while FoA transfers geometric style by warping the facial landmarks in the content image to a specific artist's (e.g. Modigliani) canonical facial landmark pattern (with small variations added) learned by training a model on a dataset of stylized portraits. FoA then separately transfers textural style with a standard style transfer method (e.g. Gatys, STROTSS). When we compare DST and FoA in Figure~\ref{fig:foacompare}, we demonstrate ``one-shot FoA" since the style images used to produce the outputs in~\cite{Yaniv_2019_ACM} are unavailable. That is, we assumed that we had access to one content image, one style image, and the trained FoA landmark detector. Using the detector, we found 68 facial landmarks in the content and style images and transform the style image landmarks, as described in Section~\ref{sec:points}, to get the target points. Then we followed FoA's two-step style transfer and transferred the textural style using STROTSS and alter the geometric style by naively the source points to the target points. 

The biggest difference between WarpGAN and DST is that DST is a one-shot style transfer method that works with a single content image and a single style image. WarpGAN, on the other hand, is trained on a dataset of paired pictures and caricatures of the same identities, and generates a caricature for an input content image from its learned deformation model. To compare the performance of WarpGAN and DST, we used content/style image pairs from~\cite{Shi_2019_CVPR} and ran DST. In Figure~\ref{fig:foacompare}, we show the outputs of DST and the outputs of WarpGAN taken from~\cite{Shi_2019_CVPR}. Despite the lack of a \emph{learning} component, DST results are competitive and sometimes more aesthetically pleasing than results of FoA and WarpGAN.

\subsection{Human Evaluation}\label{sec:mturk}
Quantitatively evaluating and comparing style transfer is challenging, in part because of the subjective nature of aesthetic properties defining style and visual quality, and in part due to the inherent tradeoff between content preservation and stylization~\cite{yeh2018quantitative,kolkin2019style}. Following the intuition developed in these papers, we conducted a human evaluation study using Amazon Mechanical Turk on a set of 75 diverse style/content pairs. The goal was to study the effect of DST on the stylization/content preservation tradeoff, in comparison to the base style transfer methods. The evaluation was conducted separately for STROTSS and Gatys-based methods. We considered three DST deformation regimes: low ($\beta=0.3$,$\gamma=75$), medium ($\beta=0.5$,$\gamma=50$), and high ($\beta=0.7$,$\gamma=10$) for STROTSS;  low ($\beta=3$,$\gamma=750$), medium ($\beta=7$,$\gamma=100$), and high ($\beta=15$,$\gamma=100$) for Gatys. So for each base method, we compare four stylized output images. The effect of varying $\beta$ and $\gamma$ is illustrated in Figure~\ref{fig:gammabeta}.

To measure content preservation, we asked MTurk users the question: ``Does image A represent the same scene as image B'', where A referred to the content image and B to the output of style transfer. The users were forced to choose one of four answers: ``Yes'', ``Yes, with minor errors'', ``Yes, with major errors'' and ``No''. Converting these answers to numerical scores (1 for ``No'', 4 for ``Yes'') and averaging across content/style pairs and users, we get a \emph{content score} between 1 and 4 for each of the four methods.

To evaluate the effect of the proposed deformable framework, we presented the users with a pair of outputs, one from the base method (Gatys or STROTSS) and the other from DST, along with the style image. The order of the first two is randomized. We asked the users to choose which of the two output images better matches the style. The fraction of time a method is preferred in all comparisons (across methods compared to, users, content/style pairs) gives a \emph{style score} between 0 and 1. 0.7 means that the method ``wins'' 70\% of all comparisons it was a part of. The evaluation interfaces are provided in the supplementary material.

In total, there were 600 unique content comparisons: 4 questions$\times$75 images for Gatys and an equal number for STROTSS. 123 users participated in the evaluation, and each comparison was evaluated by 9.55 users on average. The standard deviation of the content choice agreement was 0.79 (over a range of 1 to 4). For stylization, there were 450 unique comparisons in total: 3 comparisons between the base method and each of the 3 DST deformation regimes$\times$75 images for Gatys and likewise for STROTSS. 103 users participated in the stylization evaluation, and each comparison was evaluated by 8.76 users on average. For each comparison, 6.47 users agreed in their choice on average.

\begin{figure}
\includegraphics[width=0.95\linewidth]{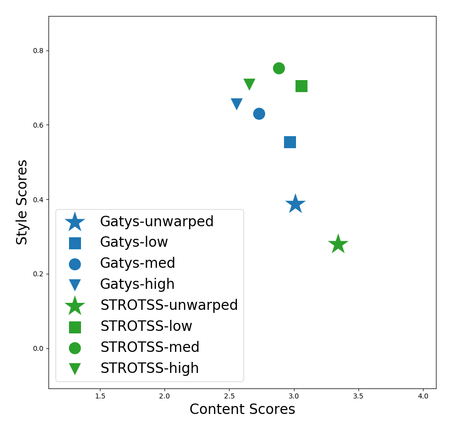}
\caption{Human evaluation results, comparing DST in different deformation regimes with STROTSS (green) and Gatys (blue). DST provides a much higher perceived degree of style capture without a significant sacrifice in content preservation.}\label{fig:mturk}
\end{figure}

Results of this human evaluation are shown in Figure~\ref{fig:mturk}. Across the deformation regimes (low, medium, high), for both STROTSS and Gatys, DST significantly increases the perceived stylization quality, while only minimally reducing the perceived content preservation. Note that some reduction to the content score can be expected since we intentionally alter the content more by deforming it, but our evaluation shows that this drop is small.

\subsection{Limitations}
\begin{figure}
\includegraphics[width=0.95\linewidth]{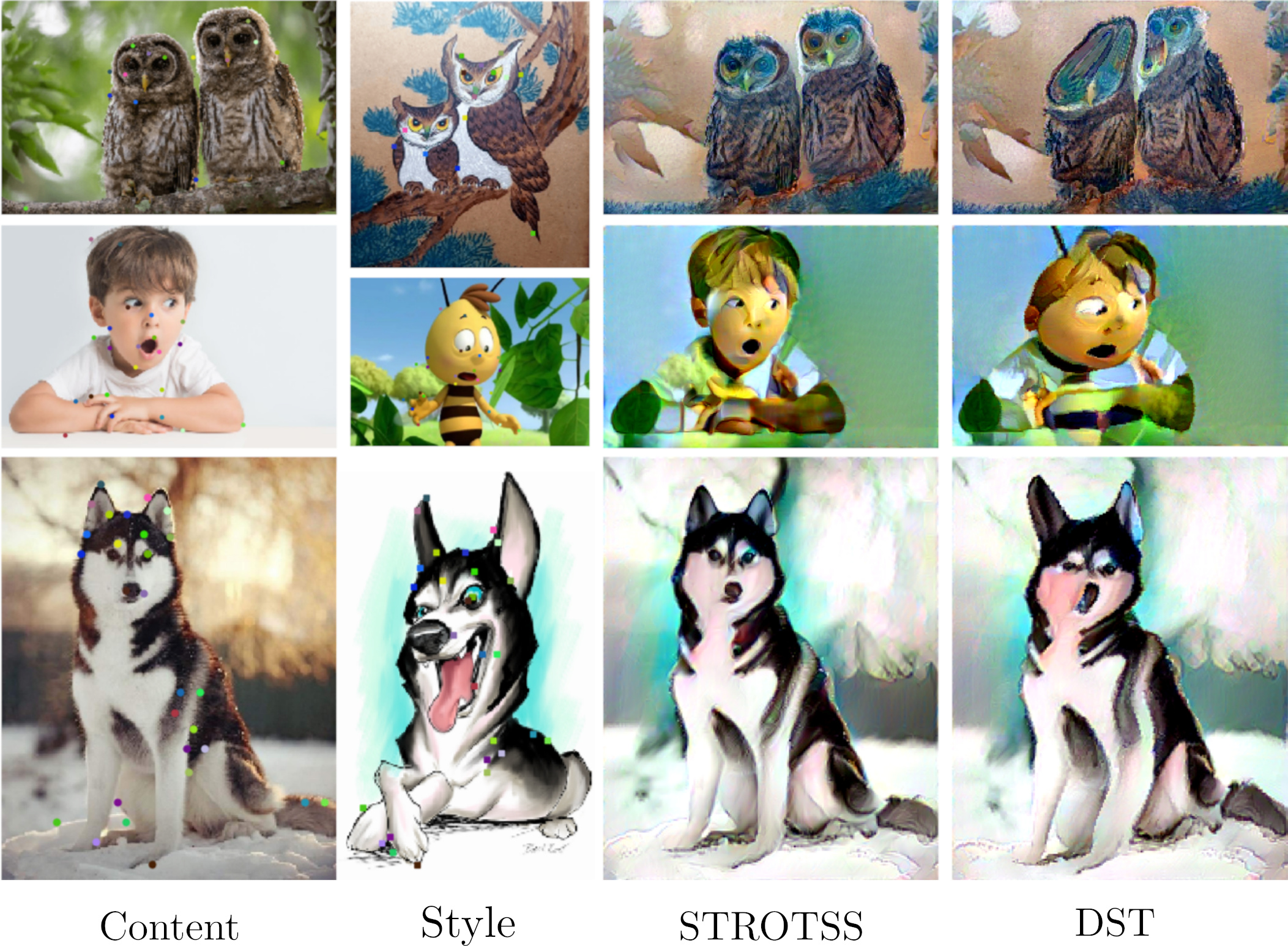}
\caption{Examples of DST failures. We observed that stylization failures are often due to correspondence errors or overly complex scene layout.}\label{fig:failures}
\end{figure}

In Figure~\ref{fig:failures}, we show unsuccessful examples of DST where the output image did not deform towards having a similar shape as the style image or deformed only partially. We observed that bad deformations often stem from poorly matching or too sparse set of keypoints. We expect finding better matching keypoints between images and making the method more robust to poor matches will improve results.

\section{Conclusion}\label{sec:conclusions}

Style transfer research prior to DST largely ignored geometry and shape, despite the important role these play in visual style. This chapter presented deformable style transfer (DST), a novel approach that combines the traditional texture and color transfer with spatial deformations. Our method incorporates deformation targets, derived from domain-agnostic point matching between content and style images, into the objective of an optimization-based style transfer framework. This is to our knowledge the first effort to develop a one-shot, domain-agnostic method for capturing and transferring geometric aspects of style.

While this work represents an early step towards incorporating geometry into style transfer, it is far from a satisfying solution. From the narrow perspective of improving DST a better algorithm might develop more robust keypoint matching algorithms for highly stylized images. However, it is doubtful that modifying geometry using deformation fields defined by paired keypoints is the most effective approach to this problem. Even in two-dimensional art, the scenes represented are often three-dimensional, and an artist's stylization of shape and form often take this into account. It seems likely that style transfer algorithms will ultimately have to explicitly model the 3D geometry of the scene to be stylized (and potentially the scene represented in the style exemplar as well). We hope that future work will continue to explore how to  more accurately encode, extract and stylize artistic shape and form.
\chapter{Neural Neighbor Style Transfer}
\label{chpt:nnst}


\begin{figure}[h]
\centering
\includegraphics[width=0.95\linewidth]{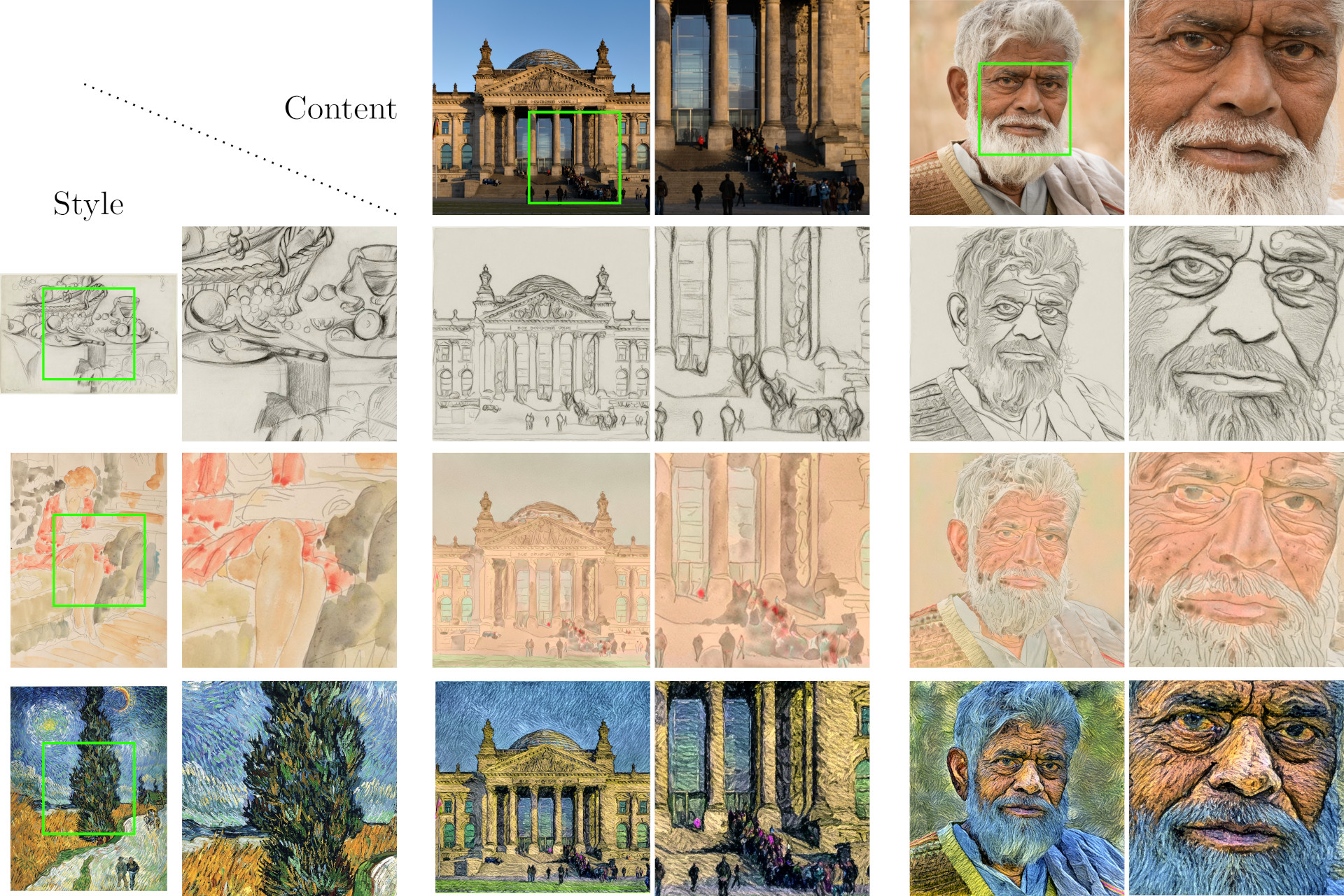}
\caption{Examples produced by Neural Neighbor Style Transfer (NNST) at 1k resolution. NNST synthesizes a stylized output by rearranging features extracted from the target style by a pretrained CNN. Synthesis can be implemented either as direct optimization of output pixels (NNST-Opt, pictured above), or as inference of pixels from features by a learned decoder (NNST-D).}
\end{figure}

\begin{figure*}
    \centering
    \includegraphics[width=\linewidth]{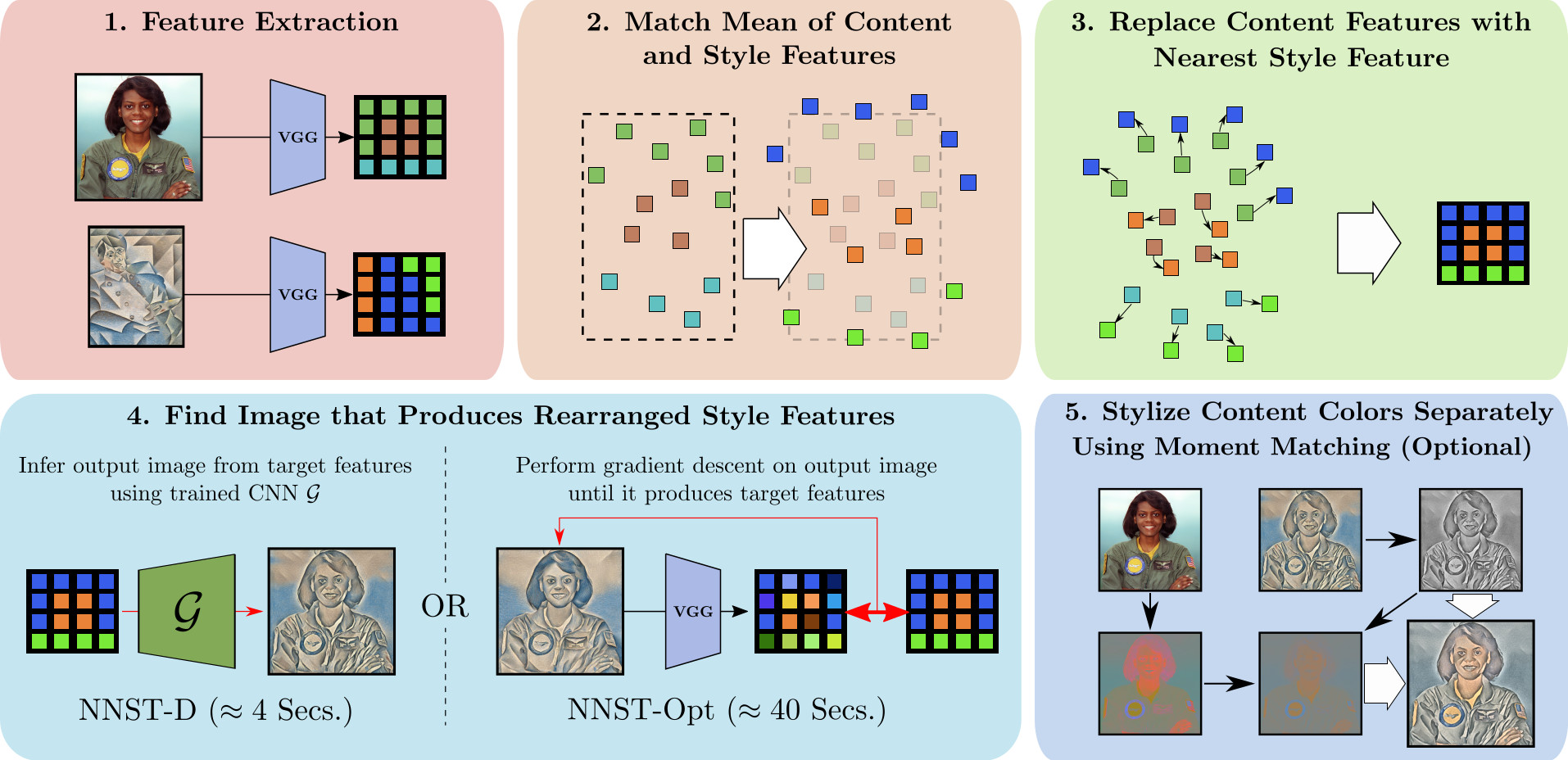}
    \caption{Overview of NNST. The fast and slow variants of out method, NNST-D and NNST-Opt, only differ in step 4; mapping from the target features to image pixels. This simplified diagram omits several details for clarity, namely: we apply steps 1-4 at multiple scales, coarse to fine; we repeat steps 1-4 several times at the finest scale; and we only apply step 5 once (optionally) at the very end.}
    \label{fig:overview}
\end{figure*}

 As detailed in Sections \ref{sec:back_tex} and \ref{sec:back_early}, the earliest style transfer algorithms were based on rearranging and blending patches of pixels taken directly from the style image \cite{efros2001image,hertzmann2001image}. Several recent style transfer algorithms \cite{chen2016fast, li2016combining, gu2018arbitrary} took inspiration from this approach, rearranging neural features extracted from the style image, then recovering an RGB image from the resulting tensor. However, based on our user study when evaluating STROTSS (see Figure \ref{fig:mturk_results_fig_strotss}), these methods do not actually improve over the original style transfer formulation proposed by Gatys et al. \cite{gatys2016image}. NNST is not based on any fundamentally new ideas, at the end of the day it also rearranges neural features and decodes an image from the resulting tensor; instead it outlines a set of important design decisions that dramatically improve outputs' visual quality. The paper this chapter is based on is currently under review, and was joint work with Michal Ku\v{c}era, Daniel S\'{y}kora, Eli Shechtman, Sylvain Paris, and my advisor Greg Shakhnarovich.

\section{Introduction}
The resulting algorithm, Neural Neighbor Style Transfer (NNST), is a straightforward new baseline for style transfer based on replacing content features with their nearest neighbor style feature (in the feature space of a neural network). It offers state-of-the-art visual quality and comes in two variants. The fast variant NNST-D (NNST-Decoder) produces a 512x512 output in $\approx 4.5$ seconds by using a learned decoder to efficiently recover pixels from a rearranged tensor of style features. The slower variant NNST-Opt (NNST-Optimization) trades off speed for quality, producing a 512x512 output in $\approx 40$ seconds, by directly optimizes the pixels of the output image in the same manner as Gatys \cite{gatys2016image}, STROTSS \cite{kolkin2019style}, and others \cite{li2016combining, mechrez2018contextual}. 

Regardless of the mechanism for recovering pixels from features, the details of exactly how the style features are rearranged are of vital importance. Patches of images are in reality tensors, having both spatial extent (the first two dimensions) and being a concatenation of the color channels (a third dimension) in a particular region of the image. Similarly we can think of patches of neural features, where the first two dimensions also correspond to a spatial region, and the third corresponds to the concatenation of a set of activation maps produced by a neural network. The original instinct of neural style transfer algorithms inspired by patch-based synthesis was to follow precisely in their footsteps and match patches of neural features, then average in regions where the patches overlap, as had been done in pixel space. Our first recommendation is that this is inappropriate for neural features, as averaging them destroys distinctive features of the style. We find that a more effective mechanism to recreate stylistic features with large spatial extent is a coarse-to-fine stylization procedure, coupled with using features from multiple neural network layers. Our second recommendation is to pre-process feature maps by zero-centering (over spatial dimensions) before computing matches. While using the cosine distance to match neural features is common, we find zero-centering (which we have not seen proposed elsewhere) dramatically increases the diversity of features used to synthesize the final output. As a fortuitous side-effect, the visual contrast between diverse features actually improves content preservation. Finally, we find that an explicit content loss is unnecessary, the implicit bias provided by initializing with the content image is enough to preserve the perceptual contents of the input.

An additional component of NNST's success is treating the luminance of the final output separately from the hue and chroma. 
While the complex texture synthesis afforded by neural style transfer is important for generating a compelling luminance channel, a  simpler procedure based on moment matching and bilateral filtering 
often generates convincing hue and chroma (parameterized as AB channels in the CIE-Lab colorspace in this work). This procedure is extremely efficient, and due to it's simplicity, is more robust than generating colors using neural style transfer. While this prevents capturing higher order terms of the style's color distribution, and sometimes more compelling results can be achieved by omitting this step, in the majority of cases either higher order distribution terms are not an important part of the style and preventing artifacts is preferable.

Qualitatively our results more accurately capture the texture of the target media than prior work, particularly when seen at high resolution. In addition our color processing dramatically increases the system's reliability, preventing the introduction of many jarring artifacts. 
We evaluate NNST-D and NNST-Opt relative to prior feed-forward and optimization based methods Via a large user study. NNST-D is judged to produce better stylizations than other fast methods and is judged roughly equivalent to the much slower state-of-the-art in optimization-based style transfer, STROTSS. NNST-Opt is judged to offer even higher visual quality, setting a new standard for artistic style transfer.

\section{Neural Neighbor Style Transfer}\label{sec:ot}
\subsection{Feature Extraction}\label{sec:fe}
Like other neural style transfer algorithms, NNST's pipeline relies on a pre-trained feature extractor $\Phi(x)$, where $x$ is an RGB image. Similar to STROTSS, $\Phi(x)$ extracts the hypercolumns \cite{mostajabi2015feedforward,hariharan2015hypercolumns} formed from the activations produced for convolutional layers in the first four blocks of pre-trained VGG16 \cite{simonyan2014very} when $x$ is passed in. We use bilinear interpolation on activations from all layers to give them spatial resolution equal to one quarter of the original image. For an image with height H, and width W, this yields an image representation  $\Phi(x) \in \mathbb{R}^{\frac{H}{4} \times \frac{W}{4} \times 2688}$. Generally we consider style to be rotation invariant, and to reflect this we extract features from the style image rotated at $0^{\circ}, 90^{\circ},180^{\circ}$ and $270^{\circ}$ in all experiments.

\subsection{Feature Matching} \label{sec:HM}
\begin{figure}[t]
    \centering
    \includegraphics[width=\linewidth]{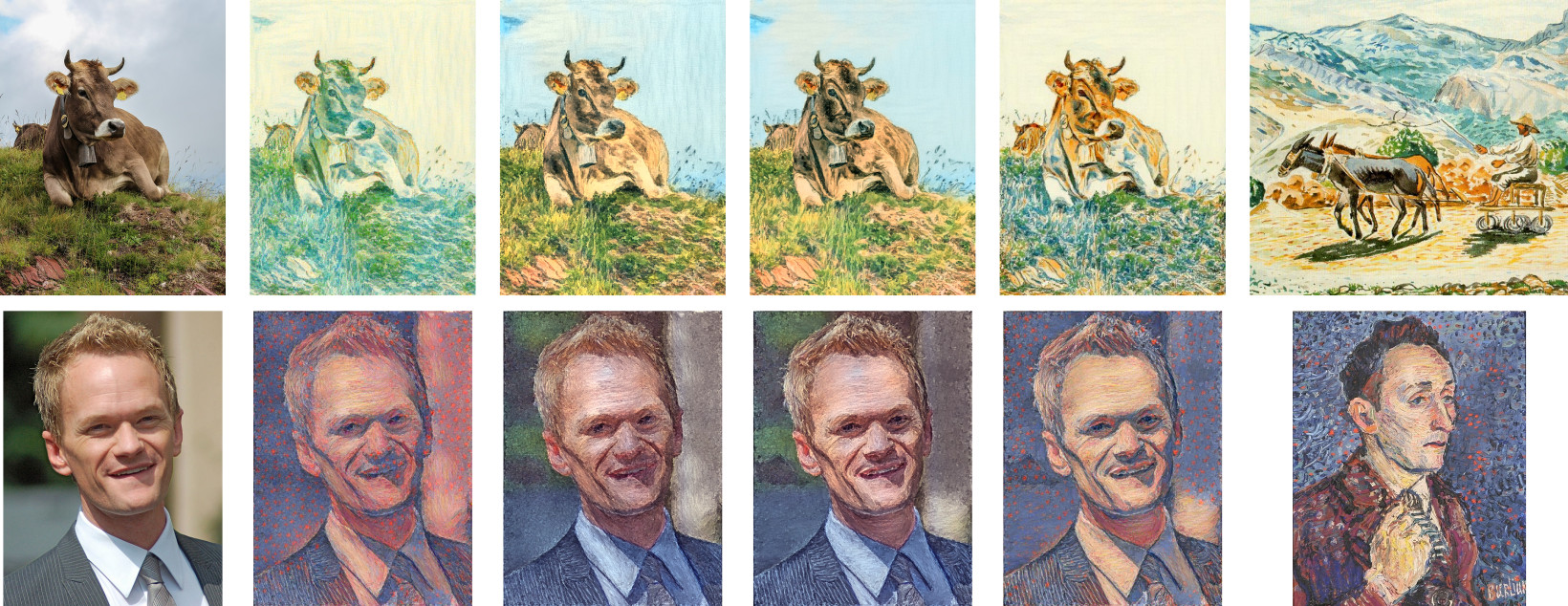}
        \begin{tabularx}{\linewidth}{>{\centering\arraybackslash}X>{\centering\arraybackslash}X>{\centering\arraybackslash}X>{\centering\arraybackslash}X>{\centering\arraybackslash}X>{\centering\arraybackslash}X}
    \hspace{-0.2cm}Content & \hspace{-0.3cm}(a) &  \hspace{-0.5cm}(b) & \hspace{-0.6cm}(c) & \hspace{-0.7cm}(d) & \hspace{-0.3cm}Style
    \end{tabularx}
    \caption{Demonstration of the affect of zero-centering features before nearest-neighbor matching. In (a) there is no zero-centering and no color processing, resulting in lower quality feature pairings that lead to more homogeneous colors and worse content preservation. (b) largely fixes the most egregious errors of (a) by adding color processing, although the features of the face are poorly defined. (c) is the default setting of NNST which uses zero-centering and color-correction, producing nice results in both cases. (d) is the same as (c) but with no color processing, this setting is less reliable and more prone to introducing visual errors than (c), but when colors are mapped correctly (as in the above examples) the results can be stunning. See Figure \ref{fig:chen_metric} for more examples of the effect of zero-centering without color post-processing.}
    \label{fig:mu}
\end{figure}

The core steps of our pipeline are outlined in Figure \ref{fig:overview}. We extract features from the style image and content image (1) zero-center the content features and style features (2). Then use nearest-neighbors matching under cosine distance (3) to replace each content feature (hypercolumn) with the closest style feature. If the content image $C$ is of size $H_c \times W_c$, and style image $S$ is of size $H_s \times W_s$, this yields a new target representation for our stylized output $T \in \mathbb{R}^{\frac{H_c}{4} \times \frac{W_c}{4} \times 2688}$ where the feature vector $T_{i}\in\mathbb{R}^{2688}$ at each spatial location is derived from the original style image, or a rotated copy (recall from Section \ref{sec:fe} that we extract features from rotated copies of the style image as well). For simplicity let $\Phi'(x)$ be the function extracting features from $x$ and its rotations, where an individual feature vector (from any spatial location in any rotation) can be indexed as $\Phi'(x)_j$. Formally:

\begin{align}\label{eq:match}
    T_i =& \Phi'(S)_j \\
    \argmin_j\hspace{0.2cm} &D\Big(\Phi(C)_i - \mu_C, \hspace{0.2cm}\Phi'(S)_j - \mu'_S \Big)
\end{align}

Where $D$ is the cosine distance, $\mu_C$ is the average feature extracted from the content image, and $\mu'_S$ is the average feature extracted from the style image and its rotated copies. While mean subtraction does not have a huge impact when using our color post-processing, it extremely important without it, enabling some stunning results in cases where the content and style are well matched (Figure \ref{fig:mu}) or a high degree of stylization is desired (Figure \ref{fig:chen_metric}).

\begin{figure}[!htp]
    \centering
    \includegraphics[width=\linewidth]{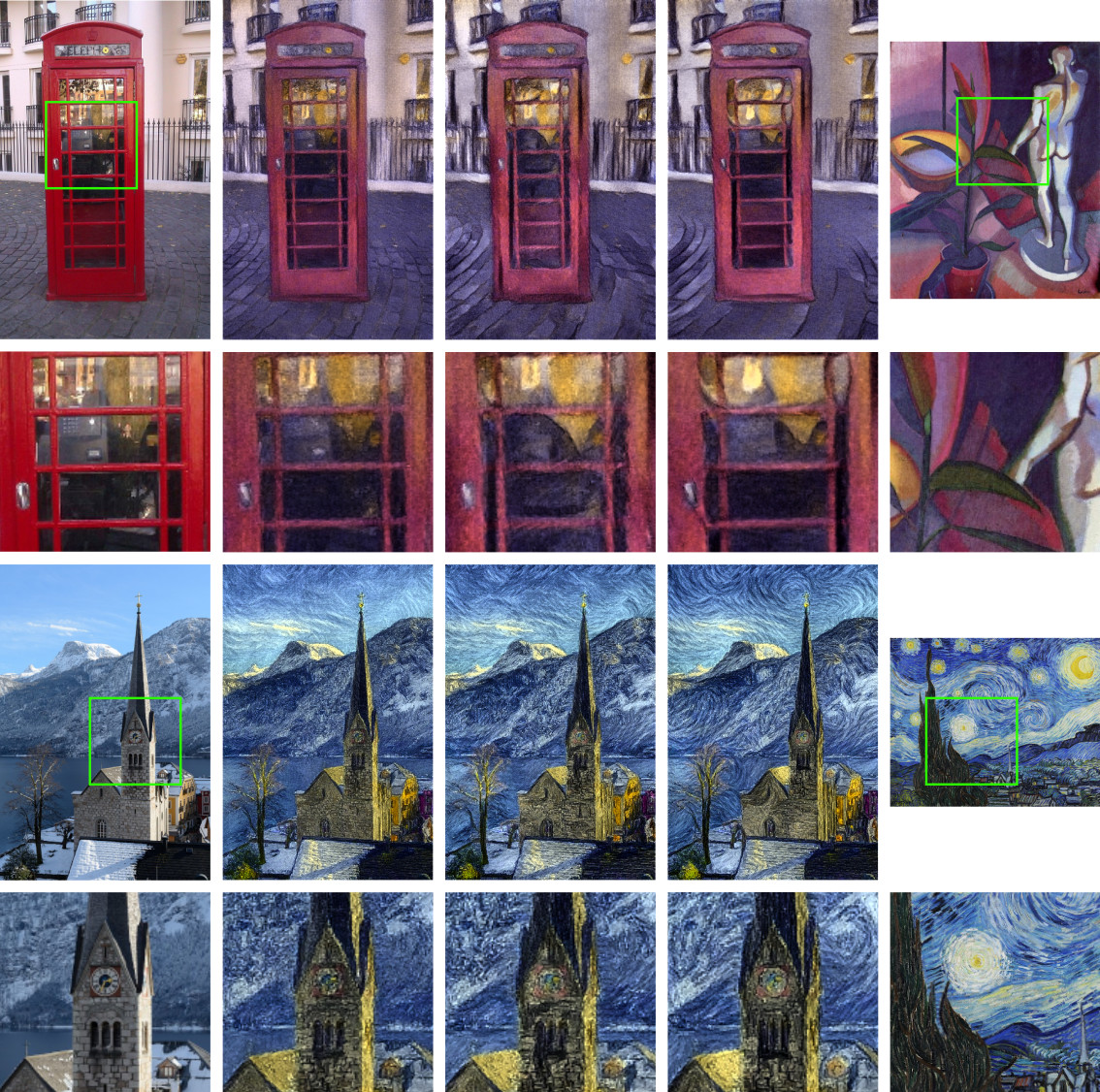}
    \begin{tabularx}{\linewidth}{>{\centering\arraybackslash}X>{\centering\arraybackslash}X>{\centering\arraybackslash}X>{\centering\arraybackslash}X>{\centering\arraybackslash}X}
    Content & (a) &  (b) & (c) & Style
    \end{tabularx}
    \caption{Demonstration of the effect of our final feature splitting phase (c). (a) is our result without feature splitting, content is well preserved, but too many photographic details bleed into the output and the brushstrokes in the 2nd row are poorly defined. (b) mimics our feature splitting phase, recomputing feature matches after each update to the output image, but matches complete hypercolumns instead of computing matches separately for each layer. This leads to unnecessary loss of content details and muddier high frequencies relative to (c).}
    \label{fig:regime}
\end{figure}

In sections \ref{sec:d} and \ref{sec:opt} we describe our feed-forward and optimization based methods for recovering image pixels from $T$ (choosing between these differentiates between NNST-D and NNST-Opt). In the main loop of our pipeline  we produce stylizations at each scale, coarse to fine, and each result is used to initialize the next scale. Throughout this process we match hypercolumns wholesale, and keep the $T$ unchanged throughout the synthesis process at a particular scale. While this is efficient (since $T$ need only be computed once per scale, and computing a single large distance matrix is well suited to GPU parallelism), and the result roughly captures many aspects of the target style, stopping at this point leads to images that fail to capture the high frequencies of the target style (Figure \ref{fig:regime}).

We believe that this effect is due to incompatible hypercolumns, which are not adjacent in the original style, being placed next to each other in $T$. Because these features have overlapping receptive fields, the output is optimized to produce the average of several features (each taken from a different region of the style) at a single output location. This manifests visually as a 'washed out' quality, an issue noted by prior work in style transfer \cite{gu2018arbitrary}, and other patch-based synthesis work ~\cite{kaspar2015self,Jamriska15,fivser2016stylit}.

We find that these issues can be largely resolved by a final phase where the the feature matching process is less constrained, a similar solution to one used by Luan et al. in the image compositing \cite{luan2017deep}. In this final phase, which we call 'Feature Splitting', matches are computed for each layer separately, resulting in $T$ consisting of novel hypercolumns where features at different layers are mixed and matched from different locations/rotations of the style image. Unlike \cite{luan2017deep} we do not compute matches only once, we recompute them after every update to the output image. In this phase features are matched relative to the current output, rather than the initial content. When using our learned decoder $\mathcal{G}$ to synthesize, this amounts to feeding the output back into the same network as 'content' five times (recomputing $T$ each time). When directly optimizing the output image, this amounts to recomputing $T$ using the current output as the 'content' after each Adam update \cite{kingma2019method}. For more examples of the effect of feature splitting and recomputing $T$ after each update, see Figure \ref{fig:chen_freq_split}.

\newpage
\begin{figure}[!htp]
    \centering
    \includegraphics[width=\linewidth]{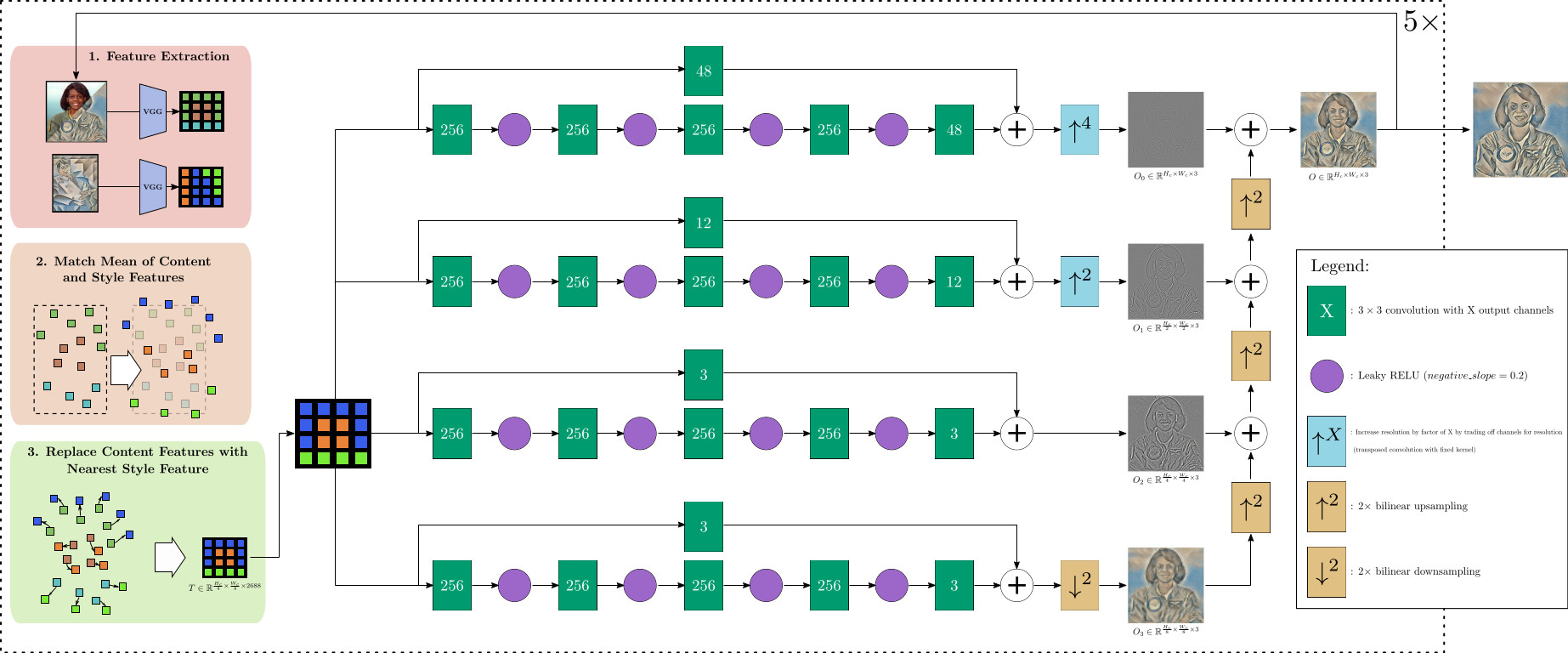}
    \caption{Overview of the architecture of the learned decoder used in NNST-D, and the associated inference procedure at a particular scale (a separate model is trained to produce final outputs that are $64,128,256$ and $512$ pixels on the long side, note that a single model is pictured and each model has 4 branches).}
    \label{fig:nnst_arch}
\end{figure}\vspace{-1cm}
\subsection{Neural Network Decoder (NNST-D)} \label{sec:d}
\subsubsection{Architecture} The decoder from feature tensor to pixels, $\mathcal{G}$, takes as input the target representation $T \in \mathbb{R}^{\frac{H_c}{4} \times \frac{W_c}{4} \times 2688}$. $T$ is then fed into 4 independent branches, each responsible for producing one level of a 4-level laplacian pyramid parameterizing the output image. Each branch has virtually the same architecture (but separate parameters), consisting of five 3x3 convolutional layers with leaky ReLU \cite{maas2013rectifier} activations (except the last layer, which is linear), and a linear residual 3x3 convolution \cite{he2016deep} directly from $T$ to the branch's output. All intermediate hidden states have 256 channels. The four branches differ only in number of output channels, having $48$, $12$, $3$, and $3$ output channels respectively. Transposed convolutions are applied to the first two branches to trade off channel depth for resolution (resulting in one $H_c \times W_c \times 3$ output and one $\frac{H_c}{2} \times \frac{W_c}{2} \times 3$ output). The third branch is not altered (resulting in a $\frac{H_c}{4} \times \frac{W_c}{4} \times 3$ output), and the output of the fourth branch is bilinearly downsampled by a factor of two (resulting in a $\frac{H_c}{8} \times \frac{W_c}{8} \times 3$ output). The final output image is synthesized by treating the output of the four branches as levels of a laplacian pyramid and combining them appropriately. In Figure \ref{fig:nnst_arch} we give an overview of how we employ our architecture at inference.

\subsubsection{Training} We train our model using MS-COCO\cite{lin2014microsoft} as a source of content images, and Wikiart \cite{saleh2015large} as a source of style images, matching the training regime of \cite{zhang2019multimodal, an2021artflow, park2019arbitrary}. These are the standard datasets for this task. MS-COCO is a diverse dataset of roughly 300,000 photographs, and Wikiart is a dataset of roughly 80,000 2D artworks from a variety of movements. Content/Style training pairs are randomly sampled independently from each dataset. For each input pair, two outputs are generated during training, a reconstruction of the style image and a style transfer:
\begin{align}
    \hat{S}  &= \mathcal{G}(\Phi(S)) \\
    O &=\mathcal{G}(T)
\end{align}
Recall that an intermediate output of $\mathcal{G}$, the decoder from feature tensor to pixels, is a laplacian pyramid that is collapsed to form the final output image (see Figure \ref{fig:nnst_arch}). Let the levels of this pyramid for the style reconstruction be $\hat{S}_{0..3}$, let the levels of a 4-level laplacian pyramid constructed from $S$ be denoted as $S_{0..3}$, let $P_i$ be the number of pixels at level $i$. These are used to compute the reconstruction loss:
\begin{equation}
    \mathcal{L}_r = \sum_{i=0}^3 \frac{\|S_i - \hat{S}_i\|_1}{P_i}
\end{equation}
We do not know what the pixels of the style transferred result should be, so we instead optimize this output using a cycle loss between $T$ and features extracted from $O$ \cite{zhu2017unpaired}:
\begin{equation}
    \mathcal{L}_{cycle}=\frac{16}{H_CW_C}\sum_{i=0}^{\frac{H_CW_C}{16}}D \Big(T_i, \hspace{0.2cm}\Phi(O)_i \Big)
\end{equation}
Where $D$ computes the cosine distance, and $i$ indexes over the spatial indexes of $T$ and $\Phi(O)$ (which are both a quarter of the original resolution of $C$). To further improve the 'realism' of our results and encourage better capturing the target style we also employ the adversarial patch co-occurrence loss proposed by Park et al. \cite{park2020swapping}:
\begin{equation}
    \mathcal{L}_{adv}= -\log \mathcal{D}(\Theta^{(4)}(S), \Theta^{(1)}(O))
\end{equation}
Where $\Theta^{(k)}(x)$ is a function that extracts $k$ random patches of size $\frac{\max(H,W)}{8}$ from $x$, and $H,W$ are the height and width of $x$ respectively. $\mathcal{D}$ is a discriminator that evaluates whether a single patch ($\Theta^{(1)}(O)$ or $\Theta^{(1)}(S)$) comes from the same image as 4 patches extracted from the style image ($\Theta^{(4)}(S)$). We use the same discriminator architecture and discriminator training described in \cite{park2020swapping}, where further details can be found. We fit the parameters of our model, $\theta_\mathcal{G}$ to minimize the full objective:

\begin{align}
    \min_{\theta_\mathcal{G}} \mathbb{E}_{C\sim\mathbb{P}_C,\,S\sim\mathbb{P}_S}\Big[\mathcal{L}_r+ \mathcal{L}_{cycle} + \mathcal{L}_{adv} \Big]
\end{align}

Where $\mathbb{P}_C$, $\mathbb{P}_S$ are the distributions of content and style training images respectively. A separate decoder is trained for each output scale (64, 128, 256, 512 pixels on the long side). Training converges fairly quickly, and we use models trained for a three epochs on MS-COCO (Wikiart images are sampled independently with replacement for each MS-COCO example). We train using a batch size of 4, and the Adam optimizer \cite{kingma2019method} with parameters $\eta=2e^{-3}, \beta_1=0.0, \beta_2=0.99$. The same set of decoder models (four total, one for each scale) are used in all experiments.

\subsection{Image Optimization (NNST-Opt)} \label{sec:opt}
While performing style transfer using $\mathcal{G}$ is fast, there are many cases where optimizing the output image directly produces sharper results with fewer artifacts. Given our target features $T$ and feature extractor $\Phi$, we find output image $O$ by minimizing following objective:
\begin{equation}\label{eq:main}
    \min_O \hspace{0.3cm} -\frac{1}{P} \sum_{i=0}^{P-1}\cos\left(\Phi(O)_i,T_i\right)
\end{equation}
where $P=W_cH_c/16$, the number spatial locations in $\Phi(O)$ and $T$. As in other optimization based methods note that we do not update the parameters of $\Phi$, only the output image itself.

Equation \ref{eq:main} is minimized via 200 updates of $O$ using Adam \cite{kingma2019method} with parameters $\eta=2e^{-3}$, $\beta_1=0.9$, and $\beta_2=0.999$. As in STROTSS, we parameterize $O$ as a laplacian pyramid with 8 levels. Doing so allows the average color of large regions to be controlled by a small number of parameters (coefficients in coarse levels of the pyramid).

\begin{figure}[htp]
    \centering
    \includegraphics[width=0.96\linewidth]{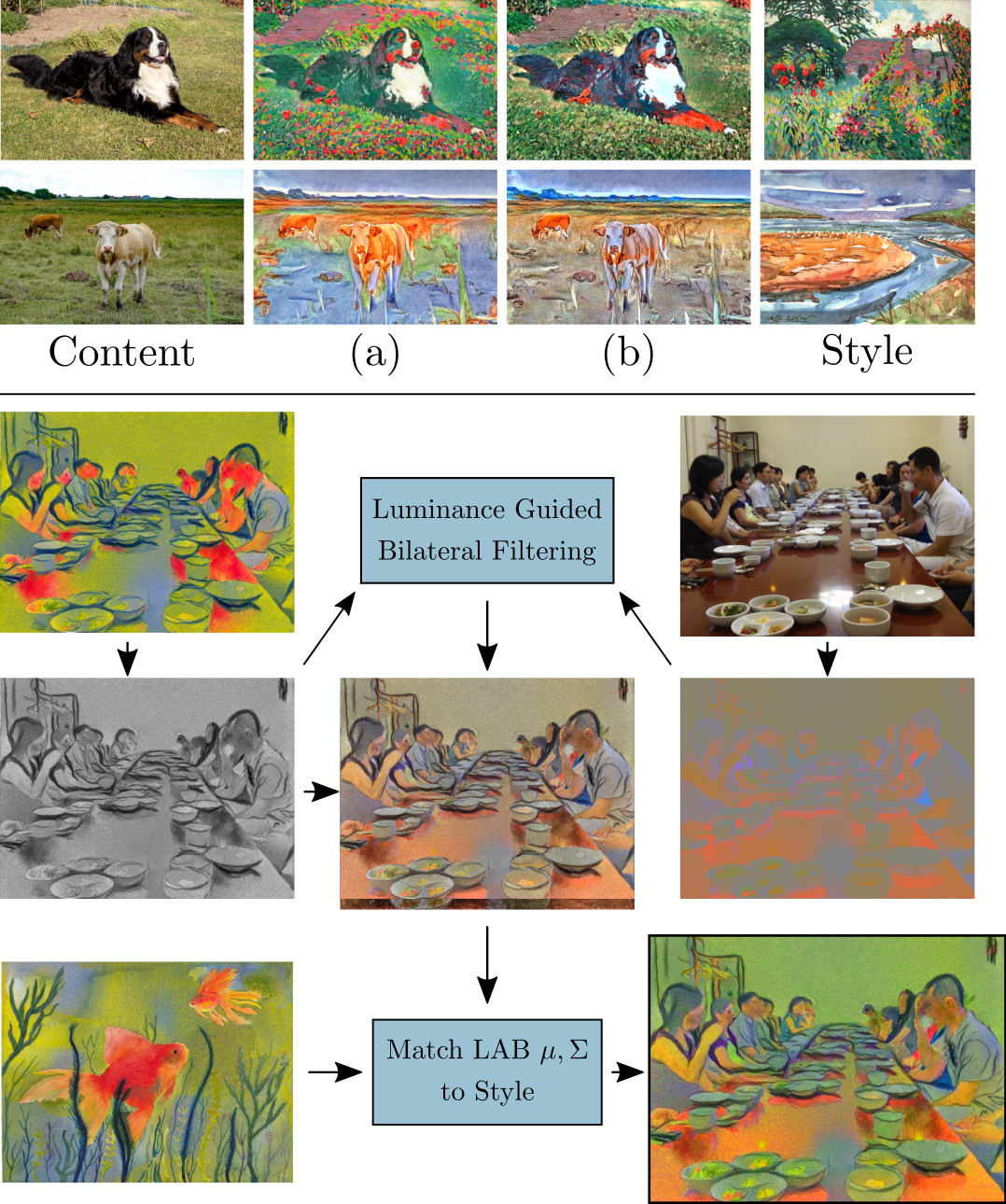}
    \caption{In the first two rows we give examples of NNST-Opt with color post-processing (b) and without (a). Color post-processing helps fix common content preservation errors due to features within a region of homogeneous color in the original content being mapped to features of multiple colors in the output. Below these examples we outline our color processing procedure. First the stylized luminance (produced by NNST-D or NNST-Opt) is extracted. Then it is used as a guide for bilateral filtering on the original content's AB channels, this aligns the boundaries of colored regions to the stylized L channel. After combining the stylized L channel with the filtered AB channels we use simple moment matching to align the output's color distribution with the style.}
    \label{fig:color}
\end{figure}
\subsection{Color Post-Processing}

We observed that a common source of perceptual errors in the outputs produced by NNST-D, NNST-Opt, and other methods, is when a region with a single color in the original content is mapped to multiple colors in the output (see the first two rows of Figure \ref{fig:color}). However, after converting our outputs to Lab colorspace, we observed that the luminance channel generally matches the target style well, and was free of artifacts. This motivated our post-processing step, which is to take the luminance generated by NNST-D or NNST-Opt, but discard the AB channels and replace them with slightly modified AB channels from the original content. In order to match the content's original AB channels to the generated L channel we perform bilateral filtering \cite{Tomasi1998BilateralFF,paris2009bilateral} on the AB channels guided by the L channel. Then we match the mean and covariance of color distribution formed by the output's L channel and filtered AB channels to the style's color distribution. (see the bottom section of Figure \ref{fig:color}). This has some similarity to the color control proposed in \cite{gatys2017controlling} (which suggests disentangling style from color by inheriting hue and chroma from the content image), 
but our addition of bilateral filtering helps outputs' final color to better conform to edges present in the luminance, and moment matching enables stylizations which more closely match the palette of the target style.

In the vast majority of cases this post-processing step improves results, however we have observed a few scenarios where it does not. 

First, for apparently monochrome styles (e.g., pencil or pen drawings), visually imperceptible color variations in the style can distort the second-order statistics used in moment matching, leading to results where desaturated colors are visible. Fortunately in these cases the unprocessed results of NNST are typically monochrome, and we detect this situation by examining the maximimum value of the AB channels' covariance matrix, and not apply the color processing if the value is below a threshold (we find 4e-5 to work well).

\begin{figure}[htp]
    \centering
    \includegraphics[width=\linewidth]{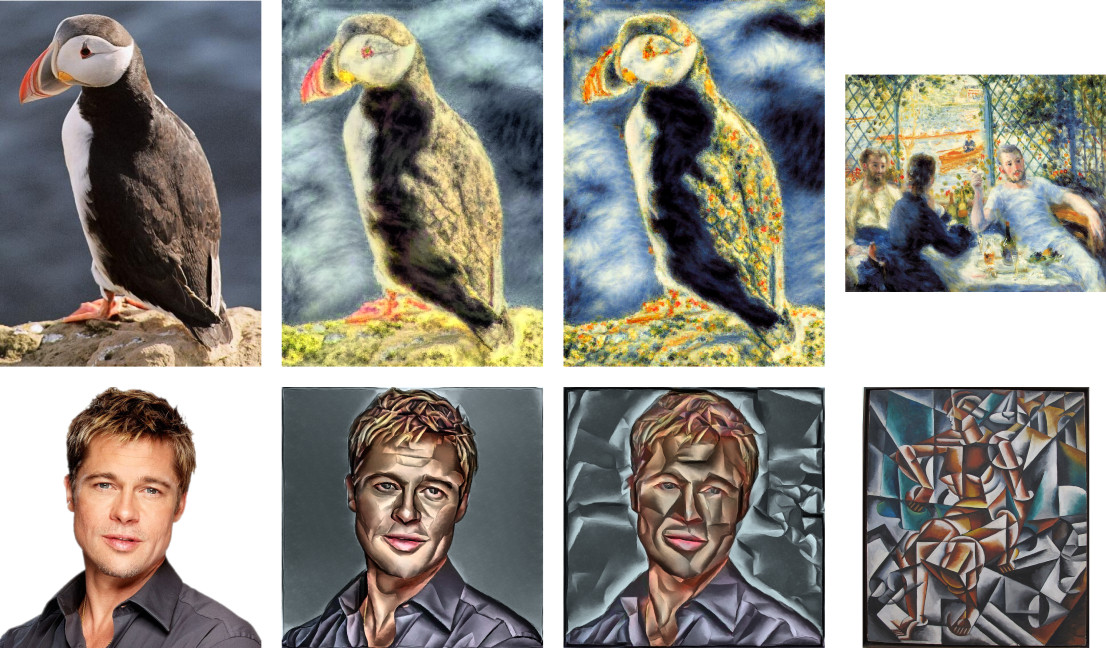}
    \begin{tabularx}{\linewidth}{>{\centering\arraybackslash}X>{\centering\arraybackslash}X>{\centering\arraybackslash}X>{\centering\arraybackslash}X}
    Content & $\downarrow$ Stylization &  $\uparrow$ Stylization & Style
    \end{tabularx}
    \caption{NNST has two mechanisms for controlling the output's level of stylization. In the first row we demonstrate the effect of increasing stylization by ommitting our color post-processing. This increases stylization by allowing greater variation from the hue and chroma of the original content image. In the second row we demonstrate the effect of varying $\alpha$, the parameter controlling the weight of the stylization at the previous scale in the initialization of the next scale. We show results with $\alpha=0.0$ (minimum stylization) and $\alpha=1.0$ (maximum stylization).}
    \label{fig:degree}
\end{figure}

Second, this post-processing step can prevent the output from matching distinctive color features of the style. For example the local color variations that defines the dots of pointillism, the abrupt color shifts within objects of cubism, or the limited multi-modal palettes used in some artwork (see Figures \ref{fig:mu} and \ref{fig:degree}). 

Third, colorful styles with a large white background can cause moment matching to result in over-saturation. In these cases it can be better to not use this post-processing step.

Fortunately, our post-processing step is simple and computationally efficient, taking less than a millisecond in our PyTorch implementation. In a practical setting it is essentially free for users to generate results both with and without the post-processing, then choose the one best suited to their needs.

\subsection{Control of stylization degree}
Including or omitting our color processing is an important mechanism for trading off between content preservation and stylization quality (row 1 of Figure \ref{fig:degree}). However, we can also take advantage of NNST's multi-scale pipeline to control the stylization level of our final output (row 2 of Figure \ref{fig:degree}).

For both NNST-D and NNST-Opt we produce stylizations at eighth, quarter, half, and full resolution. The upsampled output of the previous scale serves as initialization for the next. We initialize the coarsest scale with a downsampled version of the content image. Let $O_s$ be the output of our algorithm at scale $s$. Let $C_{s+1}, S_{s+1}$ be the content and style images at finer scale $s+1$. Let $O_s^\uparrow$ be $O_s$ upsampled to be the same resolution as $C_{s+1}$. Instead of constructing $T$ by finding matches between $\Phi(C_{s+1})$ and $\Phi(S_{s+1})$, we instead find matches between $\Phi(\alpha O_s^\uparrow + (1-\alpha)C_{s+1})$ and $\Phi(S_{s+1})$. The parameter $\alpha$ controls stylization level, with $\alpha=0$ corresponding to the lowest stylization level, and $\alpha=1$ the highest. By default, we set $\alpha=0.25$, as this generally produces a visually pleasing balance between stylization and content preservation.

 \subsection{Implementation and Timing Details}
 We implement our method using the Pytorch framework \cite{NEURIPS2019_9015}. The feed-forward variant of our method, NNST-D takes 4.5 seconds to process a pair of 512x512 content/style images. Our optimization based variant, NNST-Opt, takes 38 seconds to process the same input. Timing results are based on an NVIDIA 2080-TI GPU.

\newpage

\section{Design Decisions}

A pithy description of NNST and non-parametric neural style transfer algorithm proposed by Chen and Schmidt\cite{chen2016fast} much earlier in 2016 would reveal little difference between the two. Both methods explicitly construct a tensor of `target features' by replacing vectors of VGG-derived content features with vectors of VGG-derived style features, then optimize the pixels of the output image to produce the 'target features' (or use a learned decoder). Yet, there is a dramatic difference between the visual quality of the algorithms' outputs. As is often the case, the devil is in the details, and this section explores the important design decisions that can boost a style transfer algorithm's visual quality.

\begin{figure}[htp]
    \centering
    \includegraphics[width=\linewidth]{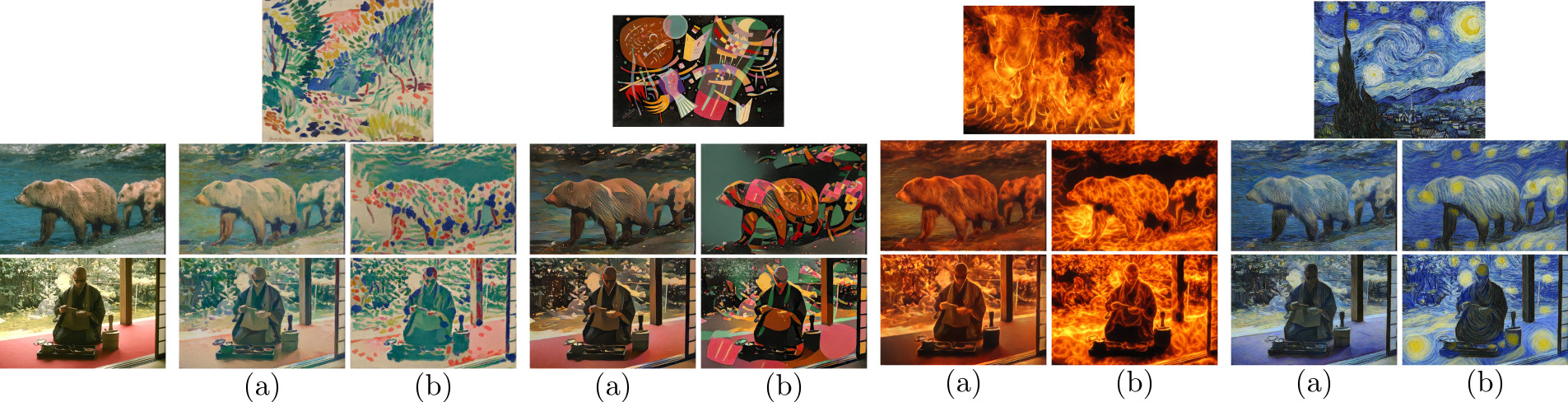}
    \caption{Visual comparison between (a.) the outputs of Chen and Schmidt \cite{chen2016fast} and (b.) a simplified variant of NNST which uses the feature splitting regime across all scales and does not employ color correction. While both algorithms share a similar high level framework, they differ in many details, resulting in NNST much better recreating distinctive visual features of the style image.}
    \label{fig:chen_nnst_comp}
\end{figure}

\begin{figure}[htp]
    \centering
    \includegraphics[width=\linewidth]{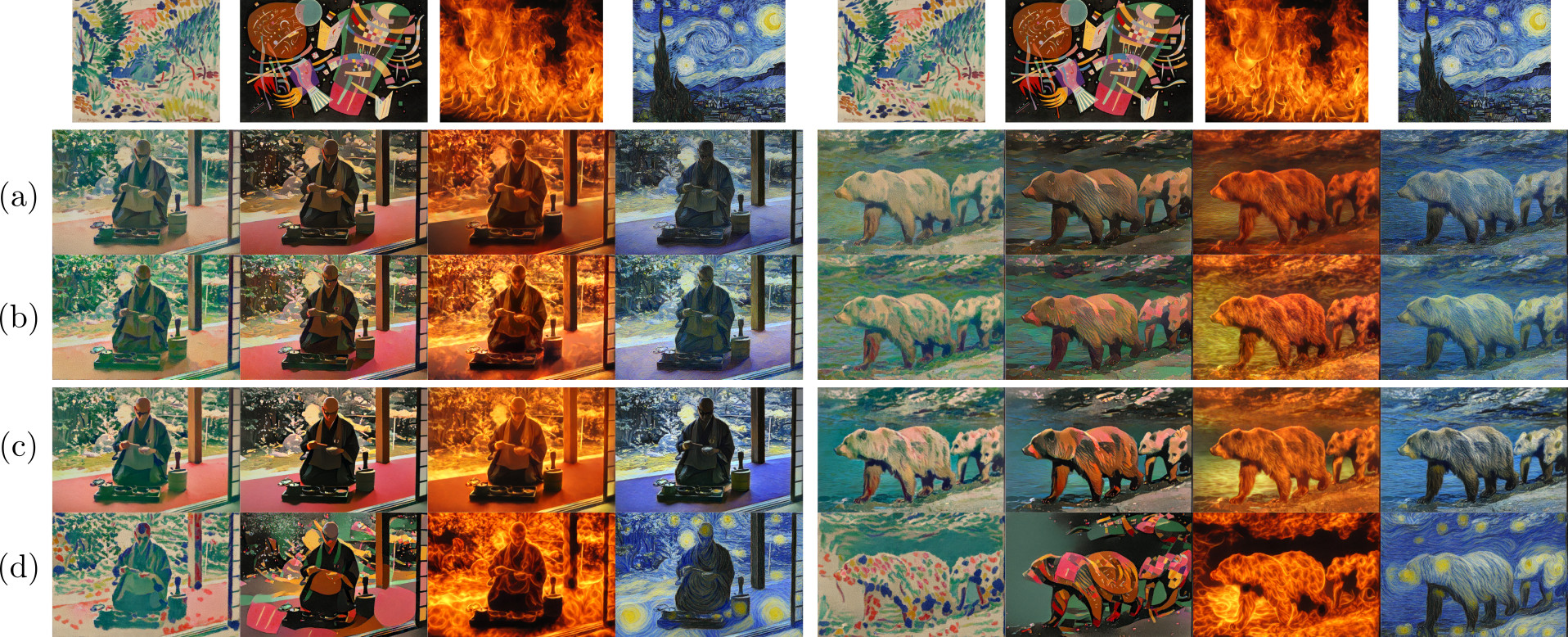} 
    \caption{Visual comparison between matching features separately for each location ($1\times1$ patches) and matching them as $3\times3$ patches: (a.) the outputs of Chen and Schmidt \cite{chen2016fast} (3x3 feature patches matched and overlaps averaged), (b.) a variant of \cite{chen2016fast} where feature patches are matched independently for each spatial location (1x1 patches, no averaging), (c.) a variant of NNST where 3x3 patches are matched and averaged, and (d.) the simplified NNST variant from Figure \ref{fig:chen_nnst_comp}. Note that matching 1x1 rather than 3x3 patches (b. and d. relative to a. and c.) allows more high frequency details of the style to appear in the output.}
    \label{fig:chen_patch}
\end{figure}

\begin{figure}[H]
    \centering
    \includegraphics[width=\linewidth]{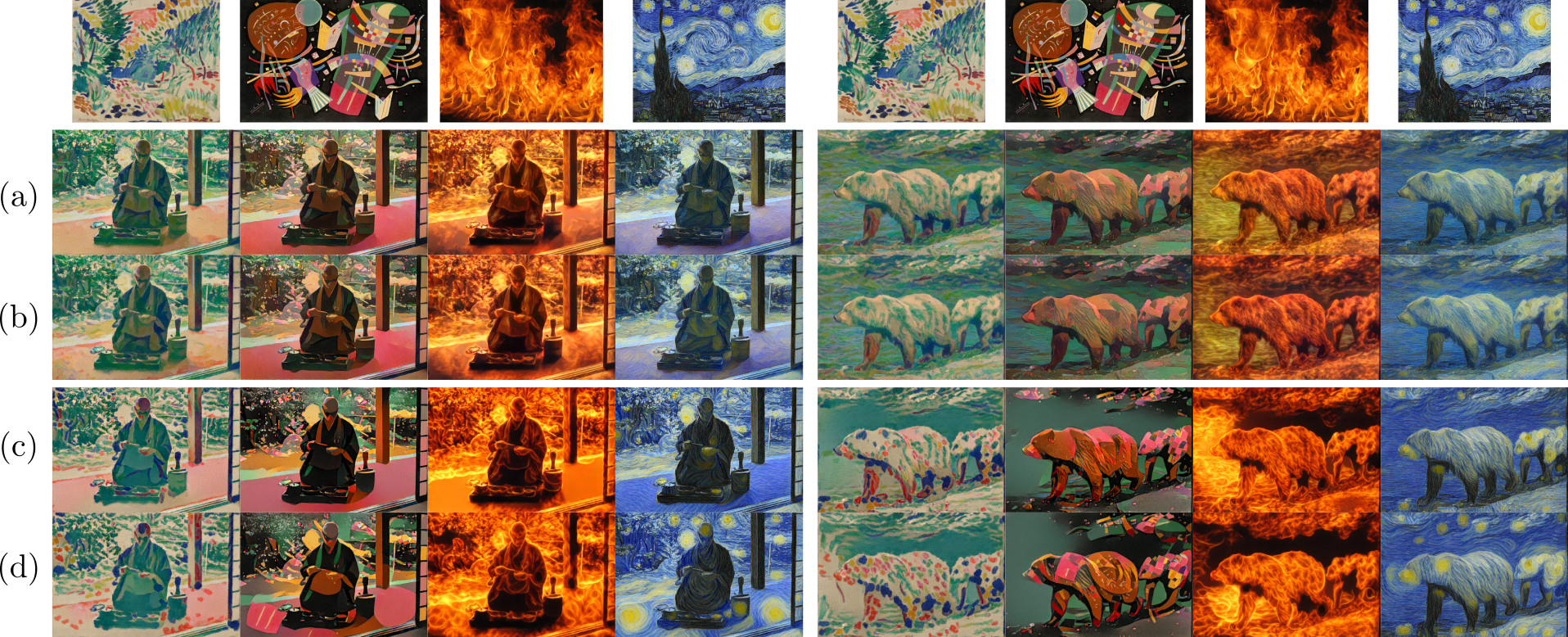} 
    \caption{Visual comparison between single-scale and multi-scale stylization: (a.) \cite{chen2016fast} w/ 1x1 patches (row b of Figure \ref{fig:chen_patch}), (b.) a. applied coarse-to-fine using the same mechanisms as NNST ($\alpha=0.25$), (c) simplified NNST at only the finest scale ($\alpha=1.0$), and (d.) the simplified NNST variant from Figure \ref{fig:chen_nnst_comp}. Stylizing coarse-to-fine increases stylization level and results in visual features of the style with larger spatial extent appearing in the output (and this effect increases with lower $\alpha$, see Figure \ref{fig:degree}). In addition, stylizing coarse-to-fine allows stylistic details to be hallucinated in large flat regions of the content image (compare the floor beneath the monk in c. and d.).}
    \label{fig:chen_scale}
\end{figure}

\begin{figure}[H]
    \centering
    \includegraphics[width=\linewidth]{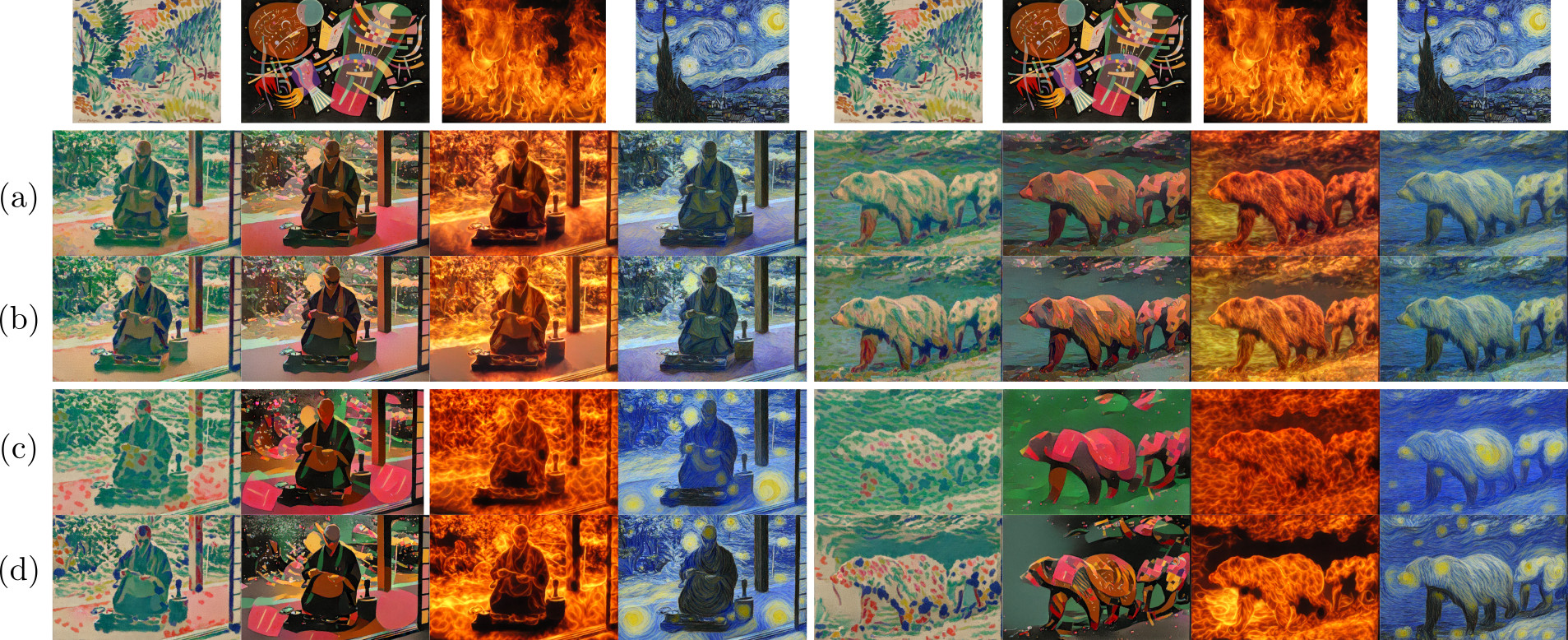} 
    \caption{Visual comparison between using zero-centering or not before matching features using the cosine distance: (a.) multi-scale \cite{chen2016fast} w/ 1x1 patches (row b of Figure \ref{fig:chen_scale}), (b.) a. using the centered cosine distance for feature matching (instead of the standard cosine distance), (c) d. using the standard cosine distance (instead of the centered cosine distance), and (d.) the simplified NNST variant from Figure \ref{fig:chen_nnst_comp}. Using the centered cosine distance not only results in a more diverse set of style features appearing in the output, the contrast between these features helps preserve the perceived contents of the original input.}
    \label{fig:chen_metric}
\end{figure}

\begin{figure}[H]
    \centering
    \includegraphics[width=\linewidth]{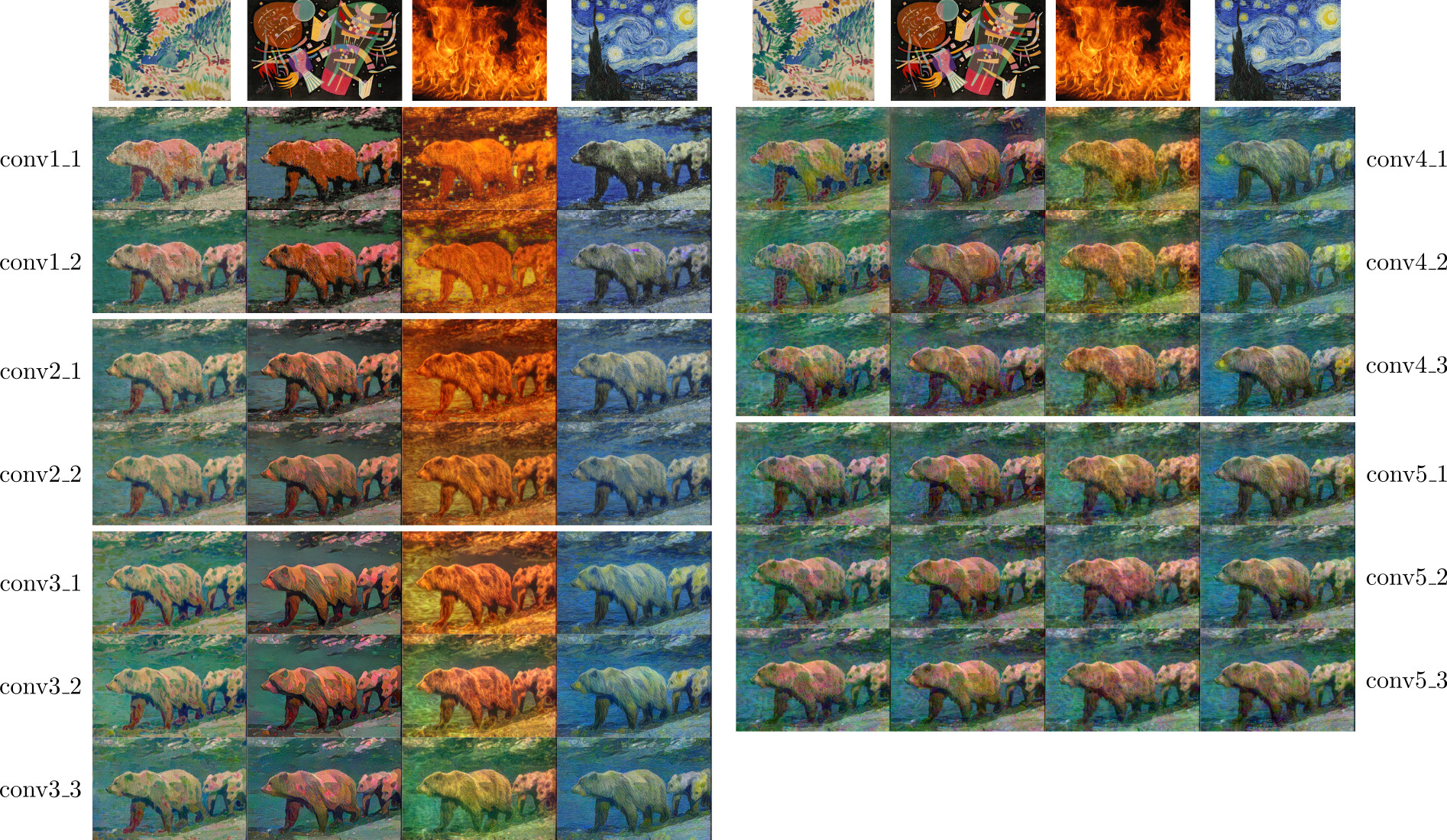} 
    \vspace{0.2cm}
    \includegraphics[width=\linewidth]{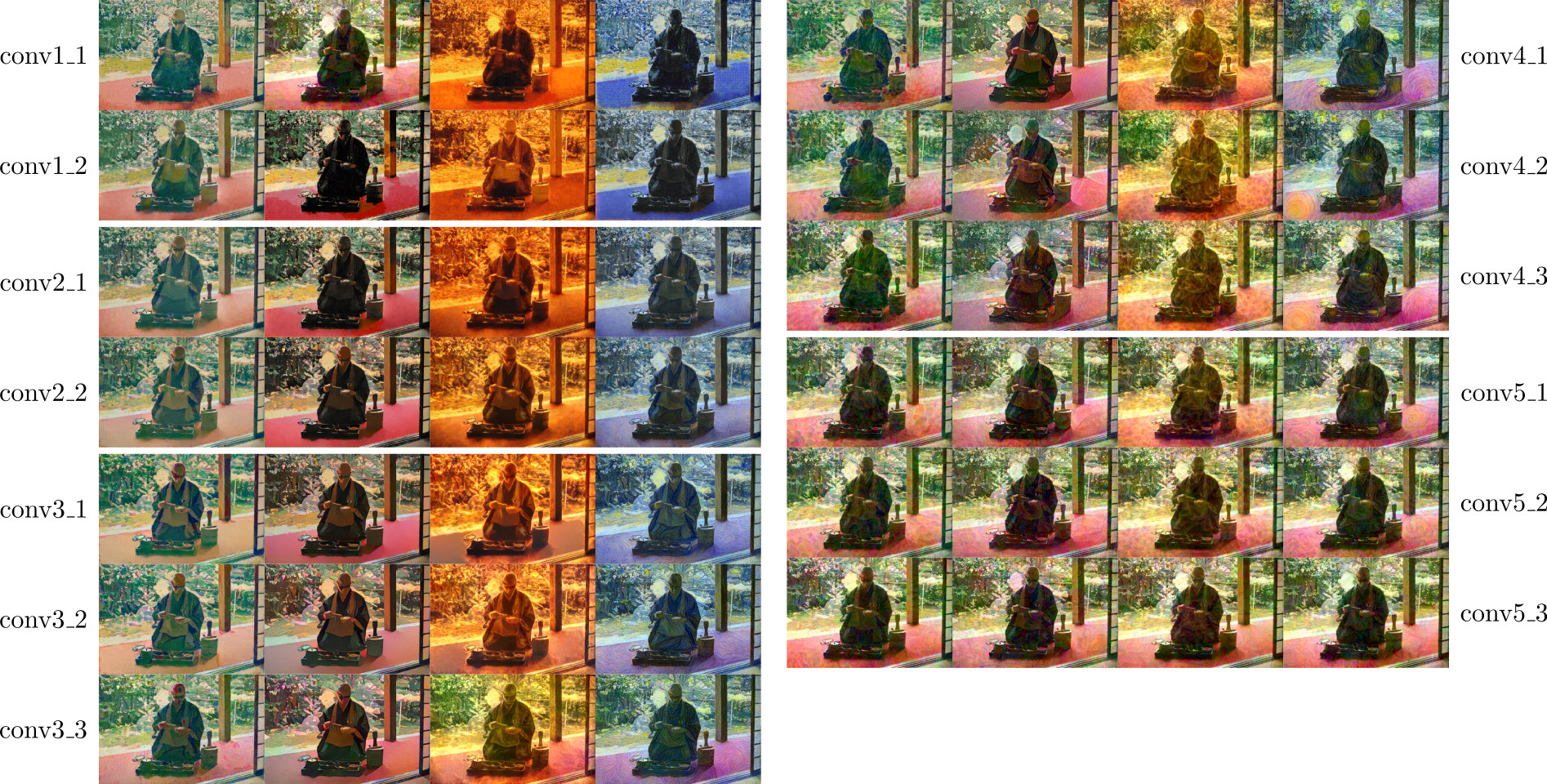}
    \caption{Visual comparison between using different individual convolutional layers of pretrained VGG-16 as a source of features. All images are produced using multi-scale \cite{chen2016fast} w/ 1x1 patches matched with the centered cosine distance (row b. of Figure \ref{fig:chen_metric} corresponds to row conv3\_1 of this figure, the default style features used by \cite{chen2016fast}). Layers in the first two conv. blocks capture color well, but not more complex visual elements. Layers in the third and fourth conv block capture complex visual elements, but not color. Layers in the fifth block do not seem closely tied to stylistic features. No layer alone is sufficient to capture all desired stylistic features.}
    \label{fig:chen_fsingle}
\end{figure}

\begin{figure}[H]
    \centering
    \includegraphics[width=\linewidth]{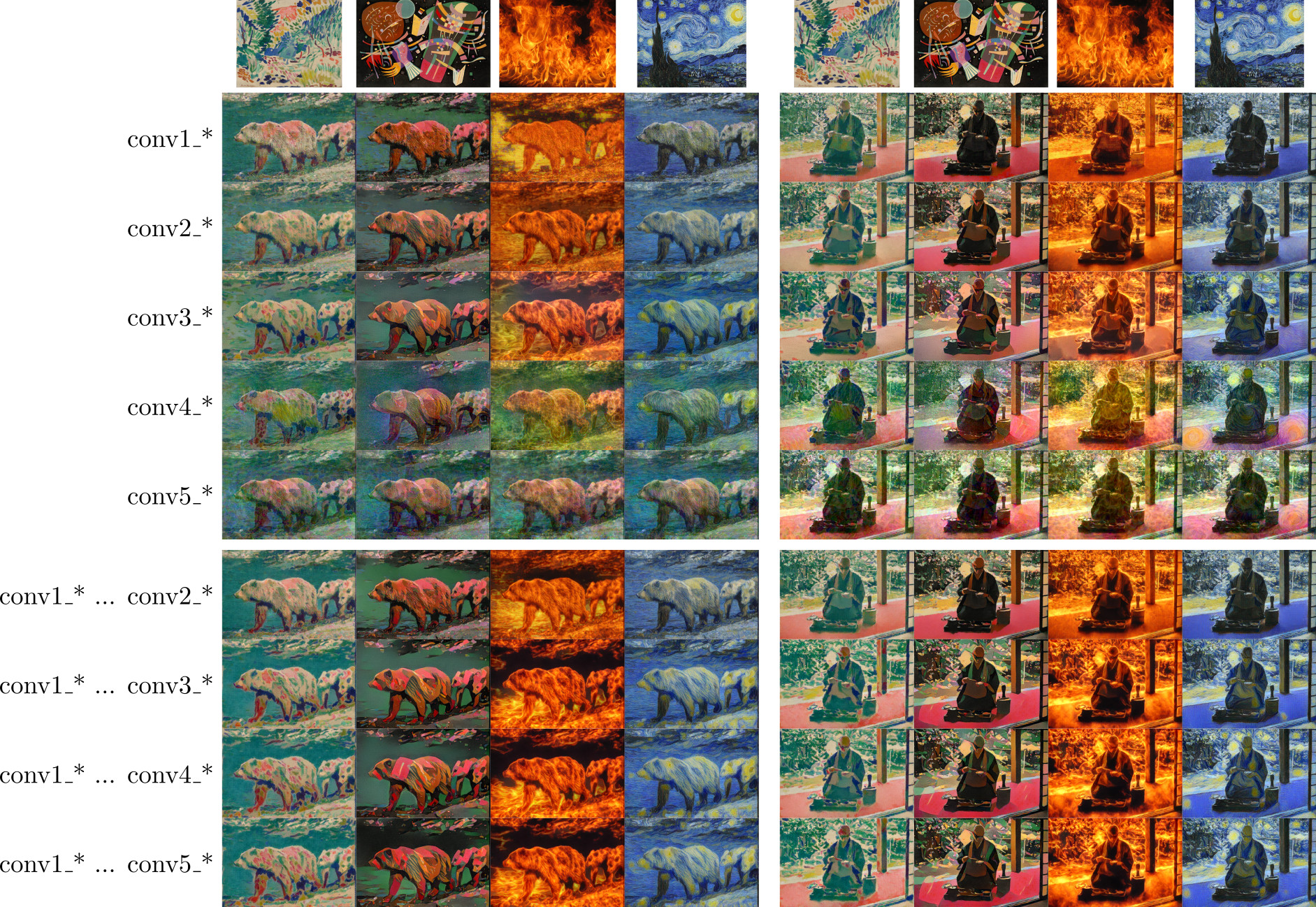} 
    \caption{Visual comparison between using features from multiple layers of pretrained VGG-16. The first 5 rows demonstrate the effect of using all the layers from a given conv. block. Just as no single layer is sufficient, no single conv. block contains a rich enough representation of style to produce satisfactory outputs. The 5th-9th rows demonstrate the effect of using all of the features up to a certain depth in the network. Most important stylistic details can be captured using the first three conv. blocks. Small improvements can be made using the 4th and 5th conv. blocks as well, but it is probably not worth the computational cost (the 4th and 5th blocks each contain 36\% of the total feature channels). While NNST uses all feature through conv. block 4, only using features through block 3 would be an obvious means to increase efficiency.}
    \label{fig:chen_fblock}
\end{figure}

\begin{figure}[H]
    \centering
    \includegraphics[width=\linewidth]{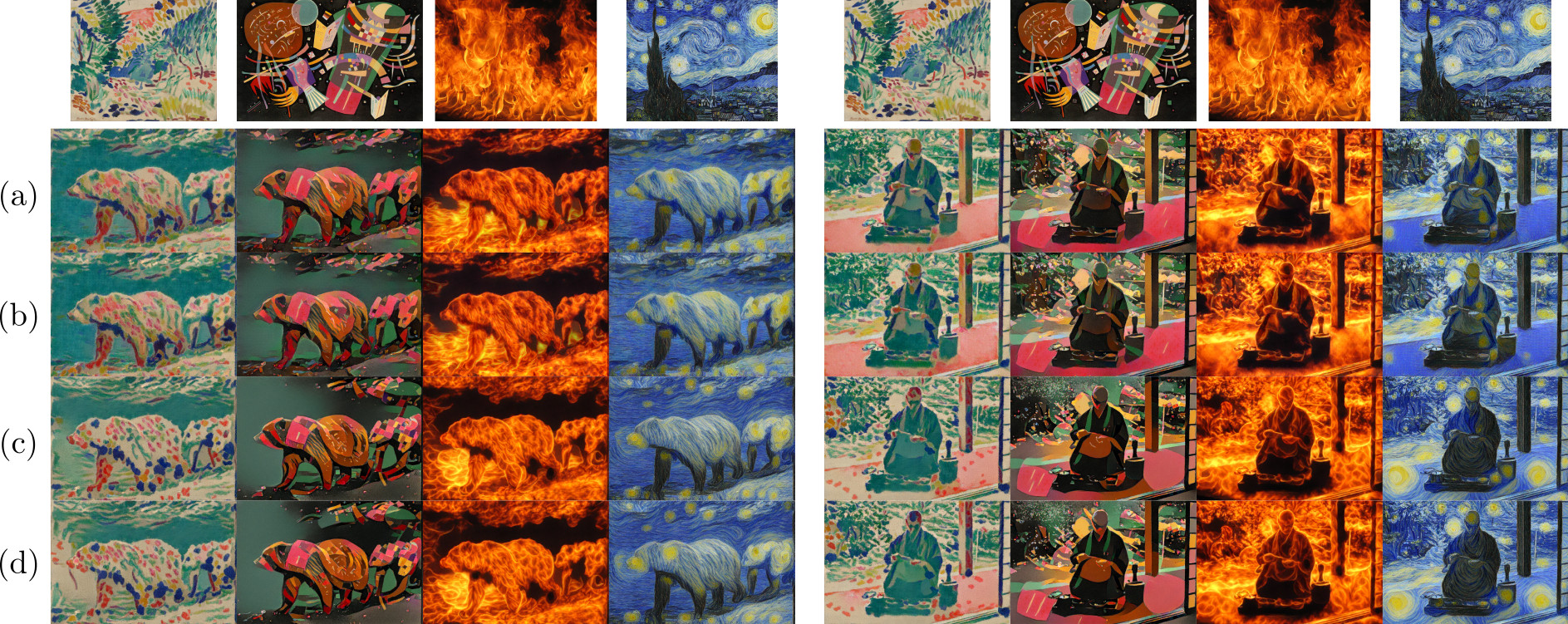} 
    \caption{The design decisions so far lead to an algorithm close to NNST, the only difference that remain to be evaluated are the frequency of computing matches and whether or not to compute matches separately for each layer: (a.) multi-scale \cite{chen2016fast} w/ 1x1 hypercolumns using layers conv1\_1-conv4\_3, matched with the centered cosine distance (row 8 of Figure \ref{fig:chen_fblock}), (b.) a. with nearest neighbors recomputed after each update of the output image, (c.) a. with nearest neighbors computed seperately for each layer, and (d.) the simplified NNST variant from Figure \ref{fig:chen_nnst_comp} (i.e. a. w/ features matched separately for each layer and recomputing matches each update). Nice results can already be achieved without feature splitting or recomputing matches each iteration (a.), however slightly sharper high frequencies can be achieved by recomputing matches (b.), and more diverse stylistic features from each layer are matched seperately (c.). When both of these modifications are applied, we essentially arrive at NNST (d.).}
    \label{fig:chen_freq_split}
\end{figure}

In Figures \ref{fig:chen_nnst_comp}, \ref{fig:chen_patch}, \ref{fig:chen_scale}, \ref{fig:chen_metric}, \ref{fig:chen_fsingle}, \ref{fig:chen_fblock}, \ref{fig:chen_freq_split}, we visually explore the effects of the our design decisions relative to \cite{chen2016fast}, and iteratively modify their method until arriving at NNST. Where appropriate these figures also demonstrate the effect of modifying individual design elements of NNST to match \cite{chen2016fast}.

\newpage

\section{Evaluation}
\label{sec:eval}

\begin{figure}
\centering
\includegraphics[width=\linewidth]{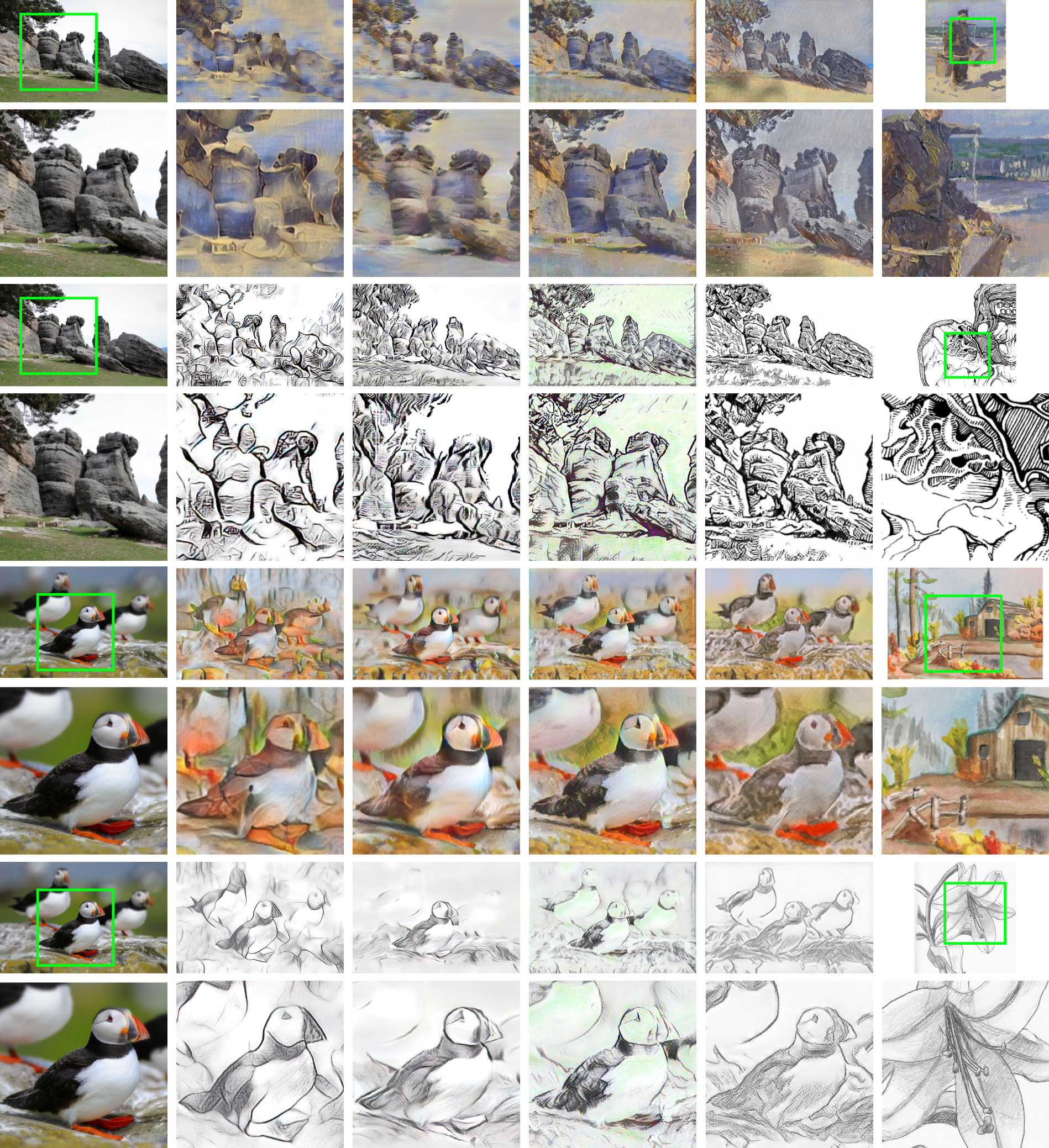}
\begin{tabularx}{\linewidth}{>{\centering\arraybackslash}X>{\centering\arraybackslash}X>{\centering\arraybackslash}X>{\centering\arraybackslash}X>{\centering\arraybackslash}X>{\centering\arraybackslash}X}
Content & \cite{Li17} &  \cite{zhang2019multimodal} & \cite{an2021artflow} & NNST-D (Ours) & Style 
\end{tabularx}
\caption{Qualitative comparision between NNST-D and the top three feed-forward methods from our user study, using oil painting, ink, watercolor, and pencil styles. Below each input and result is a zoomed-in portion of the image. While no method to date reliably reproduces the long range correlations and high frequency details of arbitrary styles, NNST dramatically improves over prior work.}
\label{fig:qual_fast}
\end{figure}

\begin{figure}
\centering
\includegraphics[width=\linewidth]{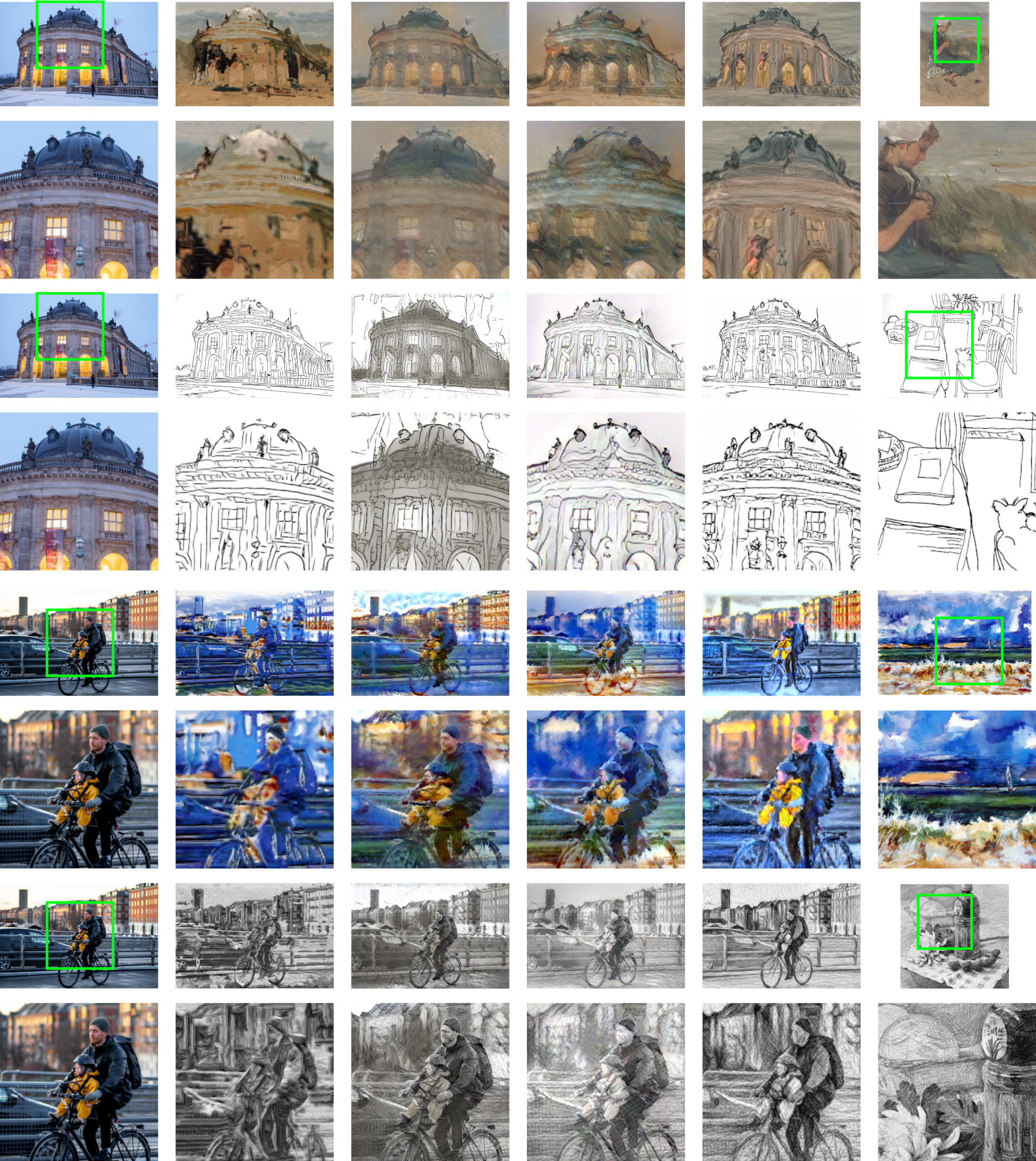}
\begin{tabularx}{\linewidth}{>{\centering\arraybackslash}X>{\centering\arraybackslash}X>{\centering\arraybackslash}X>{\centering\arraybackslash}X>{\centering\arraybackslash}X>{\centering\arraybackslash}X}
Content & \cite{liao2017visual} & \cite{gatys2016image} & \cite{kolkin2019style} & NNST-Opt (Ours) & Style 
\end{tabularx}
\caption{Qualitative comparision between NNST-Opt and the top three optimization based methods from our user study, using oil painting, ink, watercolor, and pencil styles. Below each input and result is a zoomed-in portion of the image. While no method to date reliably reproduces the long range correlations and high frequency details of arbitrary styles, NNST dramatically improves over prior work.}
\label{fig:qual_slow}
\end{figure}

\subsection{Traditional Media Evaluation Set}
In order to benchmark the performance of NNST and prior work we gathered a dataset of 30 high-resolution content photographs from Flickr, chosen for their diversity and under the constraint that they be available under a creative commons license allowing modification and redistribution. We followed the same procedure (also using Flickr) to gather ten ink drawings and ten watercolor paintings. From the Rijksmuseum's open-source collection we take ten impressionist oil paintings created between 1800-1900. We supplement these with ten pencil drawings taken from the dataset used in Im2Pencil \cite{li2019im2pencil}. In total this gives us 40 high-resolution style images. We use this dataset in the following user study, and will make it available to download.

\begin{figure}
    \centering
    \includegraphics[width=\linewidth]{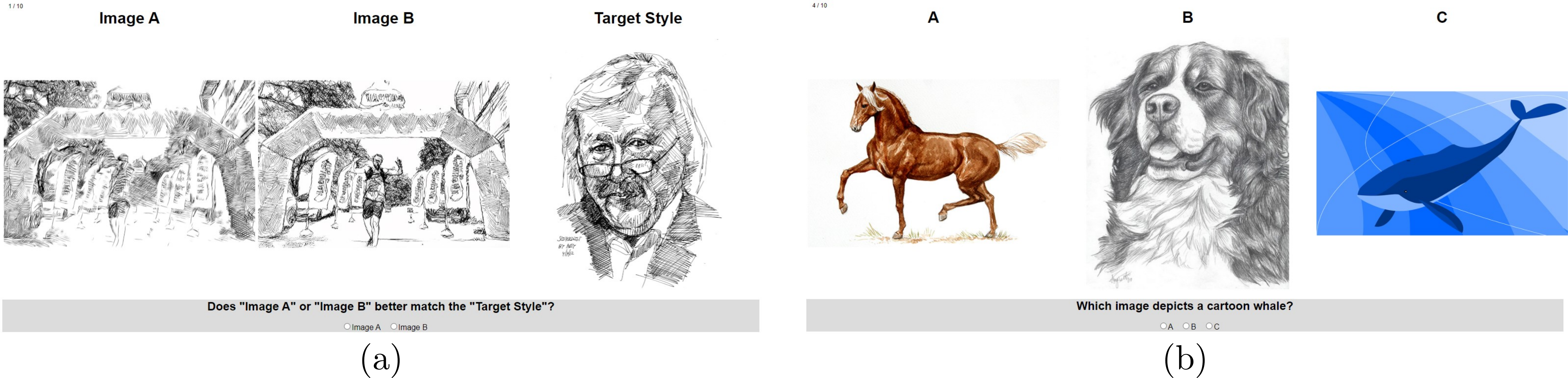}
    \caption{Example of (a) our user study interface and (b) an example attention verification question used in our study.}
    \label{fig:nnst_study}
\end{figure}

\subsection{User Study}
In order to assess the stylization quality of NNST-D and NNST-Opt relative prior work we generated stylizations for all pairwise content/style combinations in the traditional media evaluation set described above (A total of 1200 outputs per method). We conduct a user study using Prolific (\url{https://www.prolific.co/}) where users are shown the output of two algorithms (randomly ordered) for the same content/style pair (randomly selected from the 1200 possible combinations), along with the target style, and asked 'Does "Image A" or "Image B" better match the "Target Style"?'. Users are asked to judge 9 such triplets in sequence, among which is mixed one attention verification question (selecting the image that shows a cartoon whale in a randomly ordered triplet). In total we collected 225 votes per method pair, from a total of 400 unique participants. Examples of the study interface are in Figure \ref{fig:nnst_study}.

Optimization-based methods consistently produce outputs of higher visual quality than fast feed-forward methods, therefore each family of techniques is generally benchmarked separately. However, while we group these methods in Table \ref{tab:study_forced_style}, we compare both variants of our method to both families of technique. For optimization-based methods we benchmark against Gatys \cite{gatys2016image}, CNNMRF \cite{li2016combining}, Deep Image Analogies (DIA) \cite{liao2017visual}, and STROTSS~\cite{kolkin2019style}. We were unable to run the official code for CNNMRF and Deep Image Analogies, and re-implemented their methods. For fast feed-forward methods we benchmark against WCT \cite{Li17}, AvatarNet \cite{sheng2018avatar}, MST \cite{zhang2019multimodal}, and ArtFlow \cite{an2021artflow}.

The results of our study are summarized in Table \ref{tab:study_forced_style}, along with the p-values of rejecting the null hypothesis that the preference rate for NNST-D/Opt is less than 50\%. We calculate these p-values under the assumption that the votes are independent and the sum of votes received by a method is distributed as a binomial. In summary there is a statistically significant preference for our fast variant NNST-D over all benchmarked methods (fast and optimization-based) except STROTSS (STROTSS is preferred but not by a statistically significant margin). There is a statistically significant preference for NNST-Opt over {\em all} benchmarked methods. 

\begin{table}[htp]
    \centering
    \begin{tabular}{|c|c|c|c|c|}
                \hline
                  & \multicolumn{4}{c|}{Feed-Forward} \\ \hline
                  & WCT & Avatar & MST & ArtFlow \\ \hline
         NNST-D   & 71\% \small{(1.0e-10)} & 74\% \small{(1.0e-13)} & 70\% \small{(2.5e-10)} & 60\% \small{(1.6e-3)} \\ \hline
         NNST-Opt & 83\% \small{($<$ 1e-15)} & 85\% \small{($<$ 1e-15)} & 72\% \small{(5.9e-12)} & 69\% \small{(1.4e-9)} \\ \hline \hline
         & \multicolumn{4}{c|}{Optimization-Based} \\ \hline
                  & DIA & CNNMRF & Gatys & STROTSS \\ \hline
         NNST-D   & 61\% \small{(4.1e-4)} & 64\% \small{(1.6e-7)} & 60\% \small{(1.0e-3)} & 49\% \small{(0.66)}\\ \hline
         NNST-Opt & 61\% \small{(2.5e-4)} & 82\% \small{($<$1e-15)} & 65\% \small{(1.3e-6)} & 55\% \small{(1.2e-2)} \\ \hline
    \end{tabular}
    \caption{The percentage of votes received by NNST in our forced choice user study when benchmarked against prior work. In parentheses is the p-value of rejecting the null hypothesis that the preference rate for NNST is less than 50\%.}
    \label{tab:study_forced_style}
\end{table}

\subsection{Limitations} 

Our approach performs well in general but there is nonetheless areas where there remains room for improvement. For instance, physical phenomena like the drips of paint on the bear in Figure~\ref{fig:face} are not reproduced. Also stylistic features based on long range correlations such as the lines and hatching patterns in Figure~\ref{fig:qual_fast} (second and last rows) are not accurately reproduced. These cases are challenging for all methods, and while NNST makes progress relative to prior work, we believe that developing approaches which better mimic the consistency of artwork (both imposed by an artist and by the physical properties of media such as paint) remains an important unsolved problem in artistic style transfer.

We also observe that there is still a marked gap between the visual quality of outputs produced by NNST-D and NNST-Opt. Reducing this disparity will be key to creating a high-quality practical stylization algorithm.

\section{Extensions and Applications}\label{sec:ext_nnst}

\begin{figure}[htp]
\centering
\includegraphics[width=0.85\linewidth]{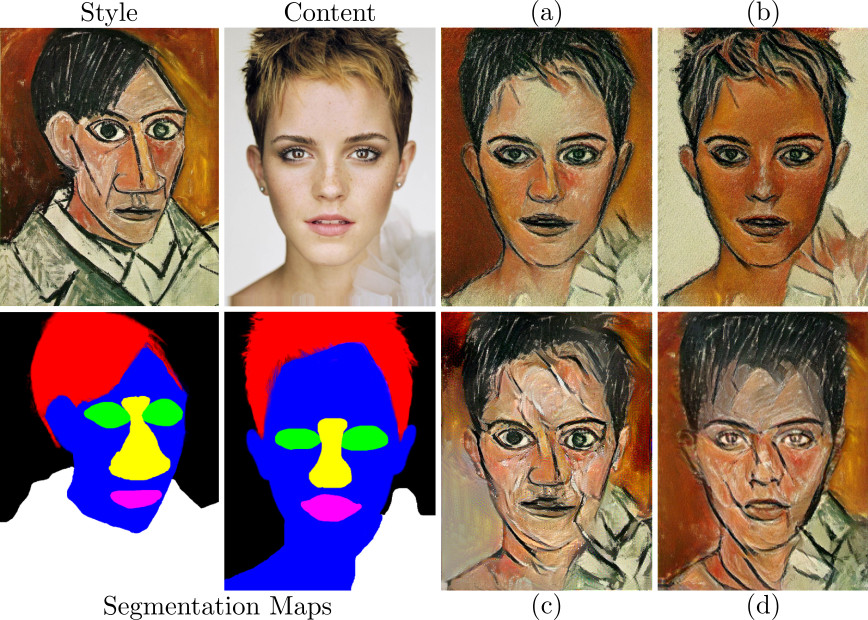}
\vspace{0.25cm}
\hrule
\vspace{0.25cm}
\includegraphics[width=0.85\linewidth]{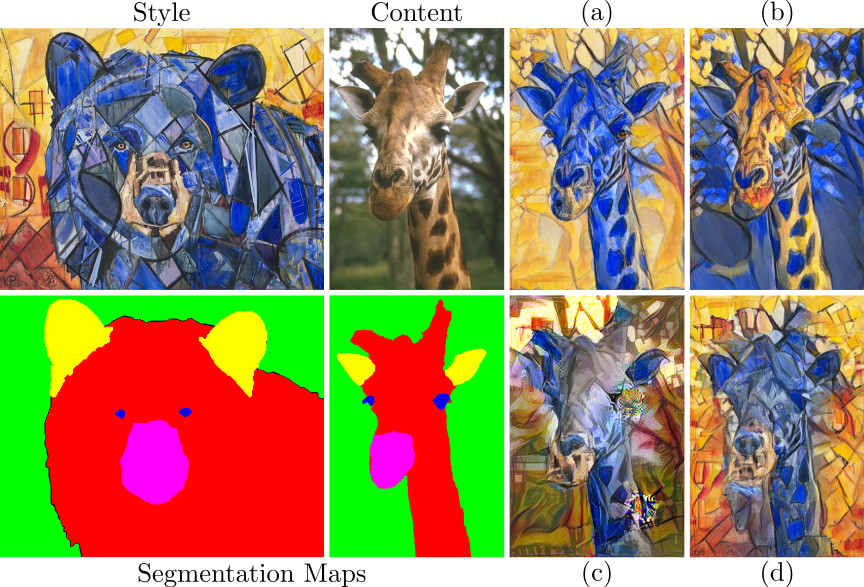}
\caption{Incorporating additional guidance (segmentation masks) into NNST-Opt. In contrast to unguided version the style of the output sementically matches the exemplar, i.e., background texture in the stylized image corresponds to the background in the style exemplar, etc. When compared to the current state-of-the-art in neural style transfer that also support guidance~\cite{gatys2017controlling, kolkin2019style} our approach better preserves stylistic details.}
\label{fig:face}
\end{figure}

Although the primary goal of our approach is to perform a generic style transfer
without requiring additional knowledge about the style and content image, in
the case when such information is available our technique can be easily extended to incorporate it. In this scenario we follow the concept of Image Analogies~\cite{hertzmann2001image} and extend our objective~(\ref{eq:match}) by
adding a term that incorporates further guidance on top of the cosine distance:
\begin{equation*}\label{eq:guided}
    T_i = \arg\min_{\Phi'(S)_j} w_{\cos}D\Big(\Phi(C)_i - \mu_C, \hspace{0.2cm}\Phi'(S)_j - \mu'_S \Big) + w_{guide}D^g(C^g_i, S^g_j).
\end{equation*}
Here~$S^g_i$ and~$C^g_i$ are downsampled versions of style and content guiding
channels (e.g., segmentation masks, see Figure~\ref{fig:face}), $D^g$~is a
metric which evaluates guide similarity at pixels~$i$ and~$j$ (in our
experiments we use sum of squared differences), and~$w_{\cos}$ and~$w_{guide}$
are weights that balance the influence of the cosine and guiding term (in our experiments we set~$w_{\cos}=0.5$ and~$w_{guide}=0.5$). In
Figure~\ref{fig:face} we demonstrate the effect of incorporating additional
segmentation masks as a guiding channels. Doing so forces features from the style to be transferred in a more predictable and semantically meaningful way. In
contrast to previous neural approaches that also support guidance~\cite{gatys2017controlling, kolkin2019style} our technique better preserves visual aspects of the original
style exemplar.

\begin{figure}[htp]
\includegraphics[width=\linewidth]{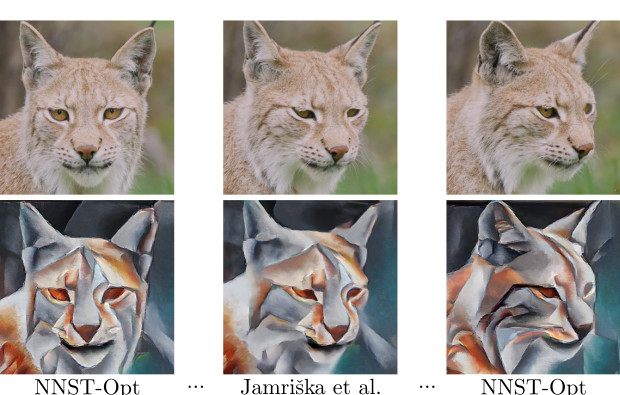}
\caption{Our approach used as a generator of stylized frames
for example-based video stylization---a few selected keyframes are stylized using NNST-Opt and the rest of the sequence is stylized using the method of Jamri\v{s}ka et al.~\cite{Jamriska19}. \href{https://home.ttic.edu/~nickkolkin/nnst_video_supp.mp4}{Video results} are available online.}
\label{fig:lynx}
\end{figure}

Besides single image style transfer our approach is practical also in the context of example-based video stylization~\cite{Jamriska19,Texler20-SIG} where the aim is to propagate the
style from a sparse set of stylized keyframes to the rest of the video sequence. In
the original setting, the stylization of keyframes is tedious as those need to
be created by hand to stay perfectly aligned with the content in the
video. Using our approach, however, one can stylize the entire sequence
fully automatically without the need to preserve alignment. By transferring
the style from an arbitrary exemplar image one can stylize a subset of frames
and then run an existing keyframe-based video stylization technique of
Jamri\v{s}ka et al.~\cite{Jamriska19} or Texler et al.~\cite{Texler20-SIG} to propagate the style to the rest of the sequence while maintaining temporal coherence (see~Figure~\ref{fig:lynx} and online \href{https://home.ttic.edu/~nickkolkin/nnst_video_supp.mp4}{video}).

\begin{figure}[htp]
\includegraphics[width=\linewidth]{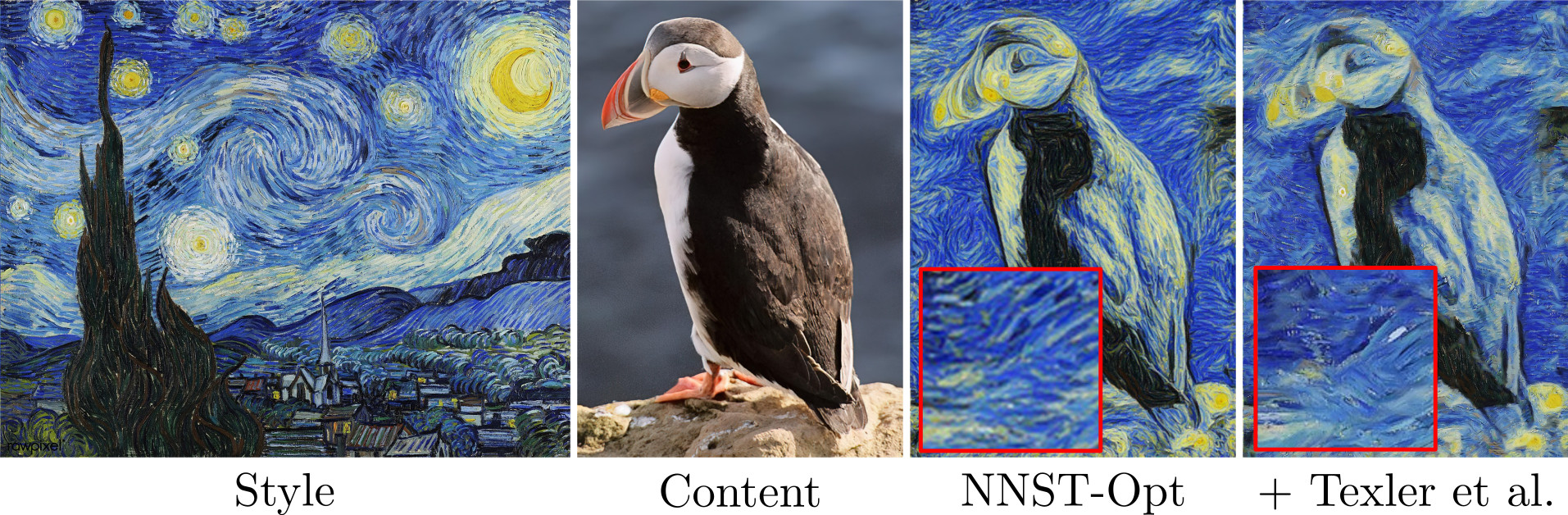}
\caption{NNST-Opt combined with the method of Texler et al.~\cite{Texler20-CAG}. the output of NNST-Opt is used as a
guide for patch-based synthesis algorithm that can operate at far higher
resolution, in this case producing a 4K resolution output. Producing outputs at very high-resolutions allows subtle but important features of physical media (e.g. craquelure) to be reproduced. However, patch synthesis cannot replicate the ability of NNST (and other neural methods) to adapt the style features to salient contours of the content, resulting in muddier middle frequencies (e.g. brushstrokes) in the 4k output.}
\label{fig:puffin}
\end{figure}

One of the limiting factors of NNST is that the memory constraints of currently available
GPUs mean that it can only deliver outputs of moderate resolution (up to 1k). To
obtain higher resolution images NNST can be plugged into the method of
Texler et al.~\cite{Texler20-CAG}. In this approach the result of neural style
transfer is used as a guide to drive patch-based synthesis
algorithm of Fi\v{s}er et al.~\cite{fivser2016stylit}, producing a high-resolution counterpart of the stylized image generated by the neural method (in Figure~\ref{fig:puffin} nearest neighbor field is upsampled to obtain a 4K output). However, a compromise here is that when comparing
middle scale features NNST performs better than patch-based
synthesis of ~\cite{Texler20-CAG} since it can adapt the style features to follow salient structures visible in the content image.

\section{Conclusion}

We have demonstrated a conceptually simple approach to artistic stylization of images, and explored several key design choices to motivate our algorithm. We showed qualitatively and quantitatively that our approach is flexible enough to support various scenarios and produce high-quality results in all these cases. Put together, we believe that these characteristics make our approach suitable for practical applications and a solid basis for future work. 




\chapter{Conclusion}
\label{chpt:conclusion}

This thesis has detailed my efforts to discover better algorithms for artistic style transfer, in particular by using non-parametric tools to better capture important visual details of an artwork's style. Chapter \ref{chpt:STROTSS} describes a non-parametric, but statistically well-motivated, style loss leveraging optimal transport. Chapter \ref{chpt:dst} describes a non-parametric parameterization of geometry and proportion based on keypoints which enabled stylization of these spatial attributes, in contrast to prior work almost exclusively focused on the stylization of texture. Chapter \ref{chpt:nnst} outlines a framework which uses nearest neighbors, one of the simplest non-parametric tools, to produce stylizations with state-of-the-art visual quality, and explores the important design decisions which lead to this success. The algorithms presented represent progress in capturing the local textures of a particular style (STROTSS in Chapter \ref{chpt:STROTSS}, and NNST in Chapter \ref{chpt:nnst}) and a first step towards stylizing the geometry of arbitrary content (DST in Chapter \ref{chpt:dst}); however, I believe the field of style transfer has a long road to travel before it can truly model an artist's style. 

All of the algorithms proposed in chapters \ref{chpt:STROTSS} through \ref{chpt:nnst} focus on the case where only a single content and style example are available at inference. Ultimately defining `style' using a single image is an artificial constraint that does not align well with peoples' typical usage of the word. `Style' is rarely defined by just one artwork, instead it's usually based on similarities within a body of work created by one, or even multiple, artists. Defining style using multiple images can help disentangle style from content \cite{sanakoyeu2018style,kotovenko2019content} by revealing which visual features are consistently present, and which shift based on content. While methods which currently take advantage of this are tuned to a particular artist or a small number of similar artists, they produce compelling results, making a strong empirical case that style should be defined using multiple images. A challenge for future work is discovering methods which can efficiently and effectively make use of a variable number of novel style examples at inference. In addition, I believe that fully taking advantage of the style images available will require methods which explicitly leverage the semantics of artwork.

\begin{figure}[htp]
    \centering
    \includegraphics[width=\linewidth]{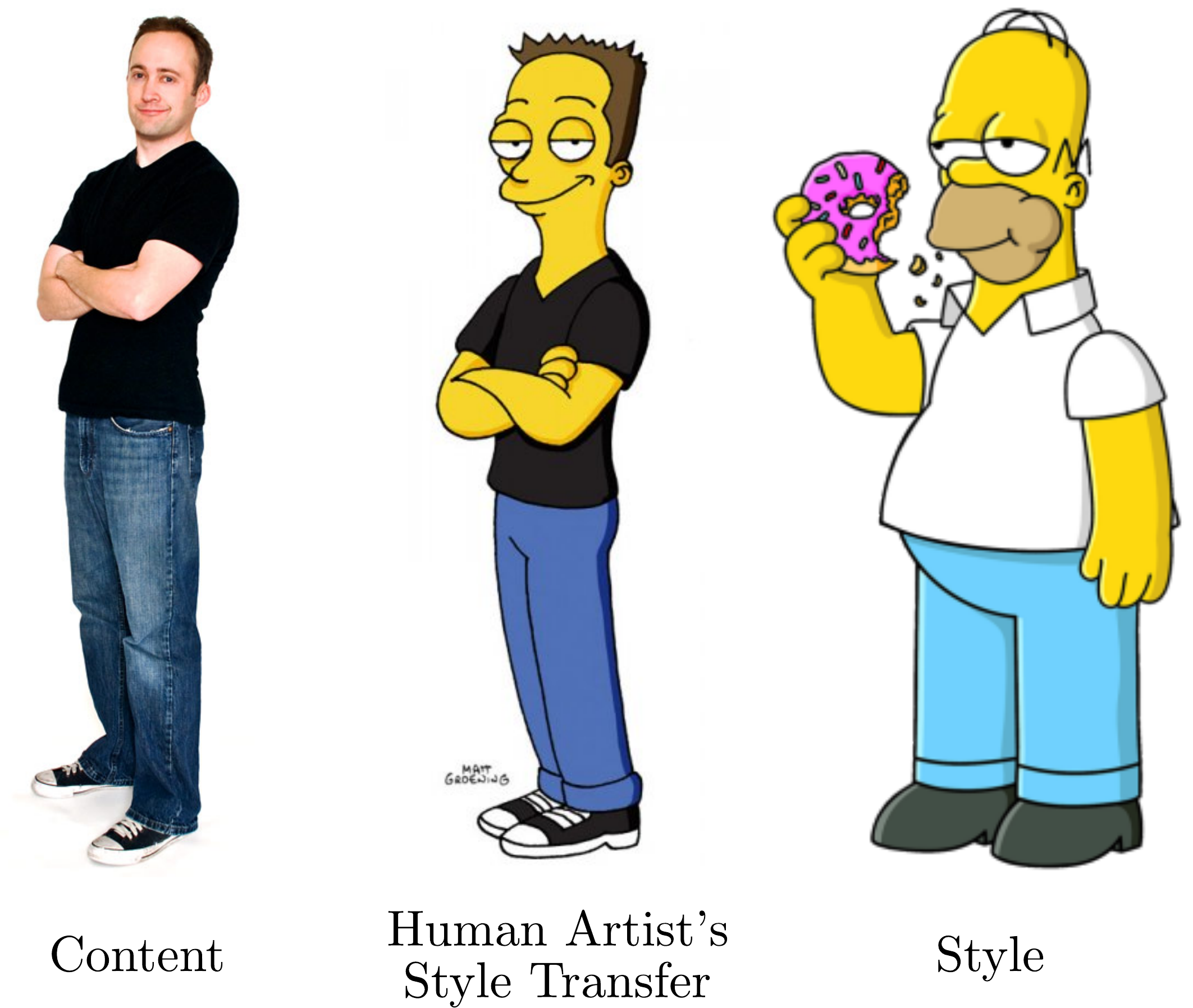}
    \caption{Example of the quality of style transfer created by human artists. Achieving this seems impossible without semantic understanding of \textbf{both} the content and style image. Semantic understanding would help explain how an artist renders the content's hair and shoes in style of the Simpsons despite there being no close analogue in the style exemplar. Image credit \href{http://cartoonkevin.com/}{Kevin Mcshane.}}
    \label{fig:pie}
\end{figure}

One of the largest gaps between style transfer algorithms today and most artist's creative process is that artists will alter or generalize style based on the semantics of the object to be represented. Figure \ref{fig:pie} displays a content image of artist Kevin McShane, a style image of Homer Simpson, and a self-portrait drawn by McShane in the style of the Simpsons. What would be required for an algorithm to produce an output of similar quality to McShane given access to the single content image and style image? This feat is probably super-human, McShane has obviously seen himself in numerous poses/lighting conditions, and has almost certainly seen a large number of images from the Simpsons. This allows him to render his hair and shoes in style of the Simpsons despite there being no close analogue in the style exemplar. It seems almost certain that stylization on the level of a human artist will require semantic understanding of \textbf{both} the content and style images. The semantics of objects in the content image must be understood so they remain recognizable in the final output. The semantics of objects the style image must be understood so that their appearance can be compared with photorealistic analogues, giving some hope of inferring the underlying rules the artist follows when depicting an object, and generalizing these rules to objects not present in the original artwork.

Semantics are not the only aspect of scene understanding important to modeling artistic style. As discussed in Chapter \ref{chpt:dst}, shape and proportion play an important role in art, and many artists' styles are influenced by the fundamentally three-dimensional nature of the scenes represented. This manifests not only through artistic modifications of a scene's underlying geometry, but also through the perspective from which a viewer is shown the rendered scene (close to the main subject, far from it, looking upwards, etc.). Ultimate success in style transfer will likely require explicitly representing the underlying geometry of content images, and the implied geometry of artworks.

This implies that style transfer must become more closely tied to mainstream computer vision tasks such as recognition, localization, and depth prediction. Not only will this benefit style transfer, developing systems which can parse the semantics and geometry of artwork can lead to models which better mimic humans' robust ability to understand scenes. Systems trained using large datasets of photographs improve every year on benchmarks where the test data matches the training distribution, but performance drops catastrophically when the distribution of test data shifts \cite{hendrycks2020many}. This is in stark contrast to humans, who can typically understand semantics, even in artistic styles they have never seen before. 
When we go to the art museum, it is common to encounter works by unfamiliar artists who produce images in a visually distinctive style. Yet we are easily, often automatically, able to recognize the semantic contents of these images. Representational art, unlike natural imagery, is a product of the artist's internal content representations. Not only that, it is specially designed to be recognizable to other humans. These properties make representational art an invaluable window into the invariances of the human visual system. I hope that future work, my own and others', will continue to explore and leverage the connection between our capabilities in recognition and creativity.


\bibliographystyle{plain}

\bibliography{references.bib}

\end{document}